\documentclass{article}

\usepackage{microtype}
\usepackage{graphicx}
\usepackage{subfigure}
\usepackage{booktabs} % for professional tables

\usepackage{hyperref}

\usepackage{url}
\usepackage{wrapfig}
\usepackage{enumitem}
\usepackage[utf8]{inputenc} % allow utf-8 input
\usepackage[T1]{fontenc}    % use 8-bit T1 fonts
\usepackage{booktabs}       % professional-quality tables
\usepackage{amsfonts}       % blackboard math symbols
\usepackage{nicefrac}       % compact symbols for 1/2, etc.
\usepackage{microtype}      % microtypography
\usepackage{xcolor}         % colors
\usepackage{float}
\usepackage{amsmath}
\usepackage{amssymb}
\usepackage{mathtools}
\usepackage{amsthm}
\usepackage{graphicx}
\usepackage{tabularx}
\usepackage[english]{babel}
\usepackage{bm}
\usepackage{balance}
\usepackage{multirow}

\usepackage[font=small,skip=0pt]{caption}
\usepackage[accepted]{icml2024}

\usepackage{amsmath}
\usepackage{amssymb}
\usepackage{mathtools}
\usepackage{amsthm}
\usepackage{enumitem}
\usepackage[capitalize,noabbrev]{cleveref}
\theoremstyle{plain}
\newtheorem{theorem}{Theorem}
\newtheorem{proposition}[theorem]{Proposition}

\theoremstyle{definition}

\theoremstyle{remark}

\theoremstyle{definition}

\newcommand{\ours}{\text{MCD}}
\newcommand{\ourslinear}{\text{MCD-Linear}}
\newcommand{\oursnonlinear}{\text{MCD-Nonlinear}}

\usepackage[textsize=tiny]{todonotes}

\icmltitlerunning{Discovering Mixtures of Structural Causal Models from Time Series Data}

\begin{document}

\twocolumn[
\icmltitle{Discovering Mixtures of Structural Causal Models from Time Series Data}



\icmlsetsymbol{equal}{*}

\begin{icmlauthorlist}
\icmlauthor{Sumanth Varambally}{yyy}
\icmlauthor{Yi-An Ma}{yyy}
\icmlauthor{Rose Yu}{comp,yyy}
\end{icmlauthorlist}

\icmlaffiliation{yyy}{Halıcıo\u{g}lu Data Science Institute, University of California, San Diego, La Jolla, USA}
\icmlaffiliation{comp}{Department of Computer Science and Engineering, University of California, San Diego, La Jolla, USA}

\icmlcorrespondingauthor{Sumanth Varambally}{svarambally@ucsd.edu}

\icmlkeywords{Machine Learning, ICML}

\vskip 0.3in
]



\printAffiliationsAndNotice{}  % leave blank if no need to mention equal contribution

 % can handle non-linear relationships between variables and flexible noise distributions

\begin{abstract}
 Discovering causal relationships from time series data is significant in fields such as finance, climate science, and neuroscience. However,  contemporary techniques rely on the simplifying assumption that data originates from the same causal model, while in practice, data is heterogeneous and can stem from different causal models. In this work, we relax this assumption and perform causal discovery from time series data originating from \textit{a mixture of causal models}. We propose a general variational inference-based framework called \ours{} to infer the underlying causal models as well as the mixing probability of each sample. 
 Our approach employs an end-to-end training process that maximizes an evidence-lower bound for the data likelihood. We present two variants: \ourslinear{} for linear relationships and independent noise, and \oursnonlinear{} for nonlinear causal relationships and history-dependent noise. We demonstrate that our method surpasses state-of-the-art benchmarks in causal discovery tasks through extensive experimentation on synthetic and real-world datasets, particularly when the data emanates from diverse underlying causal graphs. Theoretically, we prove the identifiability of such a model under some mild assumptions. Implementation is available at \url{https://github.com/Rose-STL-Lab/MCD}.
\end{abstract}

% \vspace{-8mm}
\section{Introduction}
% moving to sectiosn
% For instance, causal discovery algorithms can help uncover the relationships between various complex climatic phenomena from sea temperature measurements \citep{runge2019inferring}.  

% \textcolor{blue}{[Perhaps bring the mixture part in machine learning problems upfront. That'll avoid too much nuanced discussions about types of causal problems that are not central to our contributions.]}

% Causal discovery aims to infer the underlying causal structure among observed variables in the data \citep{spirtes2000causation}. It is a powerful tool to improve our understanding of the world. 
% Time series data presents unique challenges to causal discovery as (1) time series data often exhibits complex dependencies among both time steps and variables. (2) The space of all possible directed acyclic graphs (DAGs) increases super-exponentially with the number of time steps and variables \citep{oeis}. As a result, traditional causal discovery algorithms can hardly scale to large datasets. (3) Distinguishing between spurious correlations and true causal relationships is more difficult, especially for high-dimensional time series.

\begin{figure}
    \centering
    \includegraphics[width=0.5\textwidth]{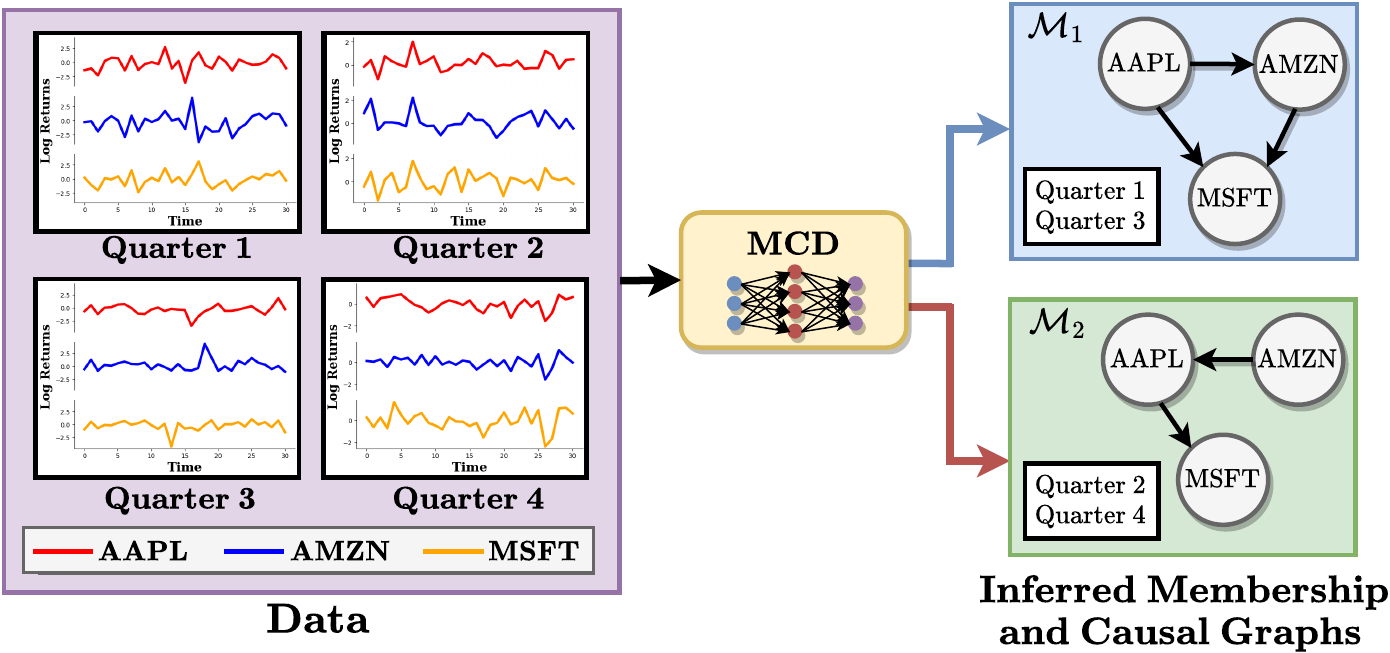}
    \caption{\ours{} discovers multiple causal graphs from time-series data by determining the mixture component membership for each sample and inferring one graph per mixture component. }

    \label{fig:overview}
\end{figure}

Causal discovery extends and complements the scope of prediction-focused machine learning with the notions of controllability and counterfactual reasoning.
It aims to infer the underlying causal structure among observed variables in the data \citep{spirtes2000causation}.
Many methods have been developed for causal discovery from time-series data based on structural causal models (SCMs) \citep{hyvarinen2010estimation, pamfil2020dynotears, yao2021learning, gong2022rhino}, conditional independence tests \citep{malinsky2018causal, runge2019detecting,runge2020discovering}, as well as the weaker notion of Granger causality \citep{granger1969investigating, khanna2019economy, tank2021neural}. 

Unfortunately, the existing methods predominantly assume that a single causal model applies to the entire data set.  In machine learning tasks, data are often multi-modal and highly heterogeneous. For example, gene regulatory networks are particular to different cells at different developmental stages. But during experiments for cell lineage, one can only track the RNA expression levels of different cells with related but distinct gene regulatory networks, since every measurement destroys the cell ~\citep{singlecell}. Similarly, stock market interactions can vary over different time periods. Using a single causal model to explain the data can result in oversimplification and an inability to capture diverse causal mechanisms. 

The task of discovering mixtures of causal graphs from data has received limited attention in the literature. Recent work, such as \citet{thiesson2013learning, saeed2020causal, markham2022distance}, have tackled the challenge of inferring causal models from mixture distributions. However, these approaches primarily focus on independent data and do not specifically address time series data. \citet{lowe2022amortized} touched upon this problem by inferring a per-sample summary graph in an amortized framework, but their approach is limited to inferring Granger causal relationships and does not account for instantaneous effects.

% Thus, it is imperative to account for this heterogeneity through multiple causal models to accurately represent the data distribution. 

In this paper, we investigate a more realistic setting in which time series data is generated from a mixture of unknown structural causal models (SCMs). We assume there are $K$ mixture components. The membership of which time series comes from which SCM component is also unknown. Our goal is to perform causal discovery by learning the complete SCMs as well as the corresponding membership for each time series sample.  A complete SCM includes both the causal graph and its associated functional equations. Figure \ref{fig:overview} summarizes the problem setting that \ours{} tackles.

% We propose a scalable deep learning-based method, Mixture Causal Discovery (\ours{}) for causal discovery from heterogeneous time series data. \ours{} combines deep learning and variational inference to infer the complete SCM and the mixture membership of each sample.   To compute the intractable posterior,  we derive and optimize a novel Evidence Lower Bound (ELBO)  of the data likelihood. We parameterize the structural equations and the noise variables with neural networks.  Theoretically, we characterize a sufficient condition for the identifiability of our mixture models, assuming the existence of `representative' points whose membership to the clusters is known with a high degree of certainty. 

We propose a variational inference-based framework, Mixture Causal Discovery (\ours{}), for causal discovery from heterogeneous time series data. Our approach learns the complete SCM and the mixture membership of each sample. To compute the intractable posterior, we derive and optimize a novel Evidence Lower Bound (ELBO) of the data likelihood. We present two variants: (1) \ourslinear{}, which models linear relationships and independent noise, and (2) \oursnonlinear{}, which uses neural networks to model functional relationships and history-dependent noise. Theoretically, we characterize a necessary and sufficient condition for the identifiability of a mixture of linear Gaussian SCMs and derive a sufficient condition for the identifiability of general SCMs under some mild assumptions. In summary, our contributions are as follows: 
\begin{itemize}
% [noitemsep,topsep=0pt]
    \item We tackle the realistic and challenging setting of discovering mixtures of SCMs for time series data with additive noise. We propose a novel variational inference approach, \ours{}, to simultaneously infer the complete SCM  and the mixture membership of each sample.
    \item Theoretically, we show that under mild assumptions, mixtures of identifiable causal models are identifiable for both linear Gaussian and general SCMs. We also derive the relationship between our proposed ELBO objective and true data likelihood. 
    \item We derive two instances of 
\ours{}: (1) \ourslinear{}, which models linear relationships with independent noise, and (2) \oursnonlinear{}, which models nonlinear relationships with history-dependent noise. 
    
    \item Experimentally, we demonstrate the strong performance of our method on both synthetic and real-world datasets.  Notably, \ours{} can accurately infer the mixture causal graphs and mixture membership information, even when the number of SCMs is misspecified.
\end{itemize}

\section{Related work}
In this section, we provide a focused literature survey on causal discovery for time series and multiple causal models.

\textbf{Causal Discovery for time series data.}
Many works on time series causal discovery use the notion of Granger causality \citep{granger1969investigating}.  \citet{tank2021neural} use component-wise Multi-Layer Perceptrons (cMLP) or Long Short Term Memory Networks (cLSTMs) with sparsity constraints on weight matrices to infer non-linear Granger causal links. \citet{khanna2019economy} use component-wise Statistical Recurrent Units (SRU), which incorporate single and multi-scale summary statistics from multi-variate time series for Granger causal detection. Amortized Causal Discovery. \citep{lowe2022amortized} aims to infer Granger causality from time series data using a variational auto-encoder framework in conjunction with Graph Neural Networks. However, Granger causality is not true causality; it only indicates the presence of a predictive relationship. Granger causality also cannot account for instantaneous effects, latent confounders, or history-dependent noise \citep{peters2017elements}.

In contrast to Granger Causality, the framework of SCMs can theoretically model instantaneous effects, latent confounders, and history-dependent noise. \citet{hyvarinen2010estimation} incorporate vector autoregressive models to the LiNGAM \citep{shimizu2006linear} algorithm to propose the VARLiNGAM algorithm for time series data. DYNOTEARS, proposed in \citet{pamfil2020dynotears}, uses the NOTEARS DAG constraint \citep{zheng2018dags} to learn a dynamic Bayesian network. However, VARLiNGAM and DYNOTEARS only account for \textit{linear} causal relationships and do not account for history-dependent noise. \citet{runge2019detecting} extend the PC algorithm to time series data with the PCMCI method. PCMCI$^+$ \citep{runge2020discovering} can handle instantaneous edges. Rhino \citep{gong2022rhino} learns the temporal adjacency matrix given observational data while modeling the exogenous history-dependent noise distribution. However, these methods assume a single causal graph for the whole data distribution. 

\textbf{Learning mixtures of causal models.}
Several works focus on the problem of causal discovery from heterogeneous independent data, but not time series. \citet{thiesson2013learning} use a heuristic search-and-score method to learn mixture of directed acyclic graph (DAG) models. However, this method only models linear causal relationships and Gaussian noise. \citet{saeed2020causal} use the FCI algorithm \citep{spirtes2001anytime} to recover the maximal ancestral graph (MAG) and use it to detect variables with varying conditional distributions across the mixture components. \citet{strobl2022causal} propose using longitudinal data, i.e., data about the same variables measured across different, potentially irregularly-spaced points of time, to infer a mixture of DAGs. However, they do not infer causal relationships \textit{across} time. \citet{markham2022distance} define a distance covariance-based kernel used to cluster sample points based on the underlying causal model. Any causal discovery algorithm can be used to infer a DAG for each inferred cluster. \citet{huang2019specific} presume `individuals' have multiple associated samples and cluster them into groups. They learn individual-specific and shared causal structures across groups using a linear non-Gaussian mixture model. 

% \citep{zhou2022causal} handle heterogeneous observational data by modeling the causal effects as functions of exogenous covariates. They can identify causal graphs with both hidden confounders and cyclic relationships. \citep{huang2020causal} exploit non-stationarity in time series data to determine causal relationships using conditional independence tests. 

Recent work \citep{huang2020causal, zhou2022causal} tackled causal discovery from data governed by heterogeneous and non-stationary causal mechanisms over time. Unlike our approach, they infer a single graph for all samples. In contrast, we model the heterogeneity of causal models across samples. Our method learns one SCM per inferred mixture component and the mixture membership of each sample. 
 % \citep{hu2018causal} address the challenge of causal inference for bivariate data generated from a mixture of mechanisms. They achieve this by clustering sample-wise Gaussian Process parameters. \sv{this is a causal inference paper, but CI on a mixture of models. Not sure if we should cite.}

% \sv{Note: some of these works have been added based on R4's comments from NeurIPS rebuttal. Pls remove if you think not super relevant}
% There has been some literature on modeling changing dynamical causal structures in time series data in contexts not necessarily related to causal discovery. \citep{fox2008nonparametric} analyze Switching Linear Dynamical Systems which can adapt to structural changes in time series data.  These systems assume the capability to switch between various linear regimes, allowing them to represent complex behaviors by transitioning between multiple linear models. \sv{Reviewer also mentioned Cross Spectral Factor Analysis, but I don't see the link. Feel free to add it here if you think it is relevant.} 
Another line of work deals with causal discovery from non-stationary time-series. Regime-PCMCI \citep{saggioro2020reconstructing} assumes that a time series can be divided into different regimes (albeit with linear causal relationships) with distinct DAGs, and aims to infer the appropriate regime for each time index. PCMCI$_\Omega$ \citep{gao2023causal} uses conditional independence tests for semistationary time series, in which a finite number of causal models occur sequentially and periodically over time. This differs from our setting, in which we assume different causal graphs govern different samples. In certain scenarios, for example, analyzing climate patterns over different locations, our approach can pool information from different samples. On the other hand, Regime PCMCI and PCMCI$_\Omega$ would have to infer causal graphs from different locations separately.

\textbf{Preliminaries.}
A Structural Causal Model \citep{pearl2009causality} (SCM) explicitly encodes the causal relationships between variables. Formally, an SCM over $D$ variables consists of a 5-tuple $\langle \mathcal{X}, \varepsilon, \mathcal{F}, \mathcal{G}, P(\epsilon) \rangle$:
\begin{enumerate}[nolistsep]
    \item A set of endogenous (observed) variables $\displaystyle \mathcal{X} = \left\{ X^1, X^2, \ldots , X^D \right\}$; 
    \item A set of exogenous (noise) variables $\displaystyle \varepsilon = \left\{ \epsilon^1, \epsilon^2, \ldots , \epsilon^m \right\}$ which influence the endogenous variables. In general, $m \geq D$ due to latent confounders; but we assume causal sufficiency, i.e., $m=D$; 
    \item A Directed Acyclic Graph (DAG) $\mathcal{G}$,  denoting the causal links amongst the members of $\mathcal{X}$;
    \item A set of $D$ functions $\displaystyle \mathcal{F} = \left\{f^1, f^2, \ldots , f^D \right\}$ determining $\mathcal{X}$ through the structural equations $X^i = f^i(\text{Pa}^i_\mathcal{G}, \epsilon^i)$, where $\text{Pa}^i_\mathcal{G} \subset \mathcal{X}$ denotes the parents of node $i$ in graph $\mathcal{G}$ and $\epsilon^i \subset \varepsilon$;
    \item $P(\epsilon)$, which describes a distribution over noise $\epsilon$.
\end{enumerate}
% We assume that each sample $X$ generated by the SCM comes from a sample space $\mathbb{X}$. When $X$ consists of independent data, $\mathbb{X} \in \mathbb{R}^D$. For 

Given time series data ${X} \in \mathbb{R}^{D \times T}$, where $T$ is the number of time steps, we can describe the causal relationships as:
 \begin{equation}
     X_t^d = f^{d}_t(\text{Pa}^d_\mathcal{G}(<t), \text{Pa}^d_\mathcal{G}(t), \epsilon_t^d),
 \end{equation}
where $X_t^d$ denotes the value of the $d^\text{th}$ variable of the time-series at time step $t$, $\text{Pa}^d_\mathcal{G}(<t)$ denote the parents of node $d$ from the preceding time-steps (lagged parents) and $\text{Pa}^d_\mathcal{G}(t)$ are the parents at the current time-step (instantaneous parents). We assume that $X_t$ is influenced by at most time-lag $L$ preceding time steps, i.e. $\text{Pa}_\mathcal{G}(<t) \subseteq \{X_{t-1}, ..., X_{t-L}\}$. This is a common assumption, shared with Rhino \citep{gong2022rhino}, VARLiNGaM \citep{hyvarinen2010estimation} and PCMCI \citep{runge2019detecting} amongst others. The causal relationships can be modeled as a temporal adjacency matrix $\mathcal{G}_{0:L}$, where $\mathcal{G}_{1:L}$ represents the lagged relationships, and $\mathcal{G}_0$ represents the instantaneous edges. We set $\mathcal{G}_{\tau}^{i,j} = 1$ if $X_{t-\tau}^{i} \rightarrow X_{t}^j$, and 0 otherwise. 
In practice, we input $L$ as a hyperparameter.
We use the additive noise model due to its identifiability \citep{gong2022rhino}:
\begin{equation}
    X_t = f_t(\text{Pa}_\mathcal{G}(<t), \text{Pa}_\mathcal{G}(t)) + \epsilon_t
    \label{eqn:scm}
\end{equation}
We mute the variable index $d$ for simplicity. $X_t \in \mathbb{R}^D$ represents the values of all variables at time $t$. 
 
 Our model shares similar assumptions to \citet{gong2022rhino}, including causal stationarity, minimality and sufficiency, and some mild conditions on the likelihood function. These assumptions are restated in Appendix \ref{sec:theory_assumptions}.

\section{Mixture Causal Discovery (\ours{})}
 
 \begin{wrapfigure}[13]{r}{0.2\textwidth}
    \centering
    \includegraphics[width=0.2\textwidth]{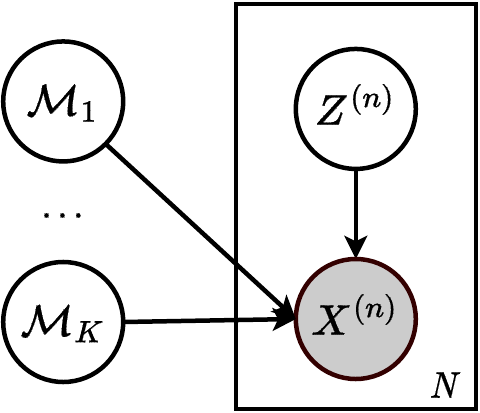}
    \caption{Probabilistic graphical model diagram of a mixture of SCMs. Shaded circles are observed and hollow circles are latent.}
    \label{fig:pgm_assumption}
\end{wrapfigure}

In this section, we detail our approach to learning mixtures of structural causal models from time-series data.
\subsection{Problem setting.}
We are given $N$ samples of multi-variate time series with $D$ variables, each of length $T$, denoted by $\left\{ X^{1:D, (n)}_{1:T}\right\}_{n=1}^N$. We assume that each sample is generated by one of the $K$ unknown SCMs, each consisting of a DAG $\mathcal{G}_k$, represented as a temporal adjacency matrix of size $(L+1) \times D \times D$, and its structural equations.

%
% The problem statement is as follows: \begin{quote}
%  \textit{Given time series samples $\left\{ X^{1:D, (n)}_{1:T}\right\}_{n=1}^N$, infer the $K$ unknown SCMs $\mathcal{M}_{1:K}$ that describe interactions occurring in a time window of length $L$.}
%  \end{quote} 
%
Our goal is to infer the DAG,  the structural equations for all $K$ SCMs, as well as the mixture membership of each sample in an unsupervised fashion. As shown in the graphical model of Figure \ref{fig:pgm_assumption}, we represent the $K$  SCMs as random variables $\mathcal{M}_{1:K}$. For each data sample indexed by $n$, we assign a categorical variable $Z^{(n)} \in \left\{1, \ldots, K \right\}$ to represent the membership of each sample to an SCM component. 

\begin{figure*}
    \centering
    \includegraphics[width=0.45\textwidth]{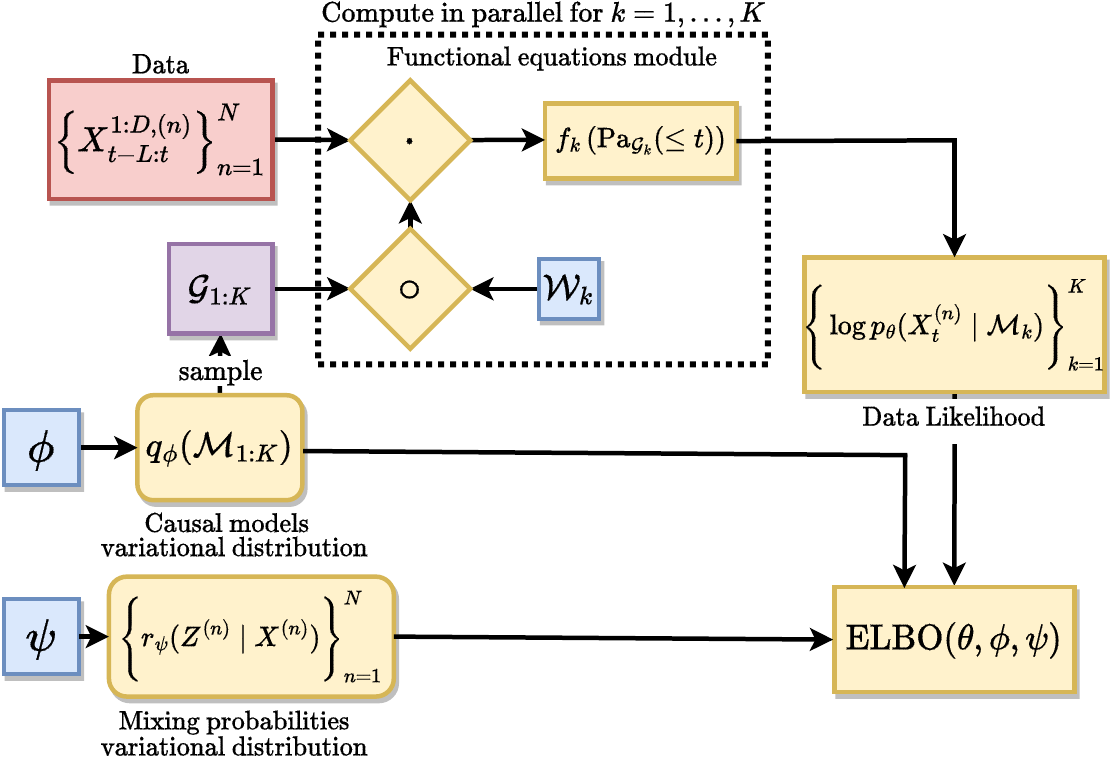}
    \hspace{20pt}
    \includegraphics[width=0.45\textwidth]{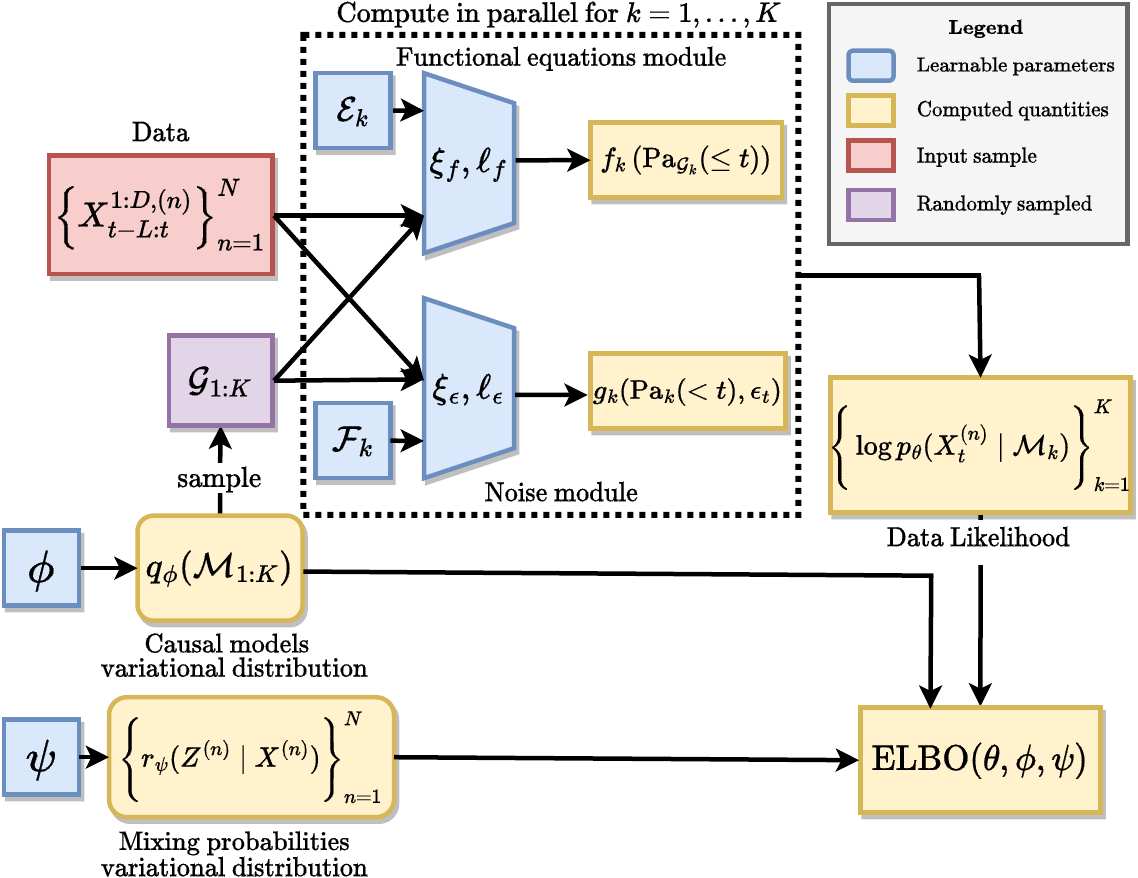}
    \caption{Overview of how the ELBO from Eq. \eqref{eqn:elbo} is calculated for (left) \ourslinear{}, and (right) \oursnonlinear{}. Given time-series data $\left\{ X_{t-L}^{1:D, (n)}\right\}_{n=1}^N$, and a DAG sample $\mathcal{G}_{1:K}$ from the variational distribution $q_\phi(\mathcal{M}_{1:K})$, we calculate the likelihood of the data under all the $K$ causal models. The likelihood is weighted by the mixing probabilities $\left\{ r_\psi \left( Z^{(n)} \mid X^{(n)}\right) \right\}_{n=1}^N$ to calculate the ELBO.}
    \label{fig:summary}
\end{figure*}
\iffalse 
A mixture of SCMs has the following generative process:
\begin{enumerate}[noitemsep]
    \item Choose $\mathcal{M}_{1:K} \sim p \left( \mathcal{M}_{1:K}\right)$.
    \item For each of the $N$ samples $X^{(n)}$:
    \begin{itemize}
        \item Choose a mixture index $Z^{(n)} \sim p \left( Z\right)$.
        \item Draw a time series  $X^{(n)} \sim p \left( X \mid \mathcal{M}_{Z^{(n)}}\right)$ from the marginal distribution of the corresponding causal model.
    \end{itemize}
    \label{eqn:model}
\end{enumerate}
\fi
 
We model each SCM $\mathcal{M}_k$ as a pair $(\mathcal{G}_k, \Theta_k)$, where $\mathcal{G}_k$ is the adjacency matrix, and $\Theta_k$ represents the trainable parameters of the structural equations and noise models. We model the causal relationships of $X_t^{(n)}$ under SCM $k$ as:
\begin{equation}
% \textstyle
    \left. X^{(n)}_t \right\rvert_k  = f_{k}(\text{Pa}_{\mathcal{G}_{k}}(\leq t)) + g_k(\text{Pa}_{\mathcal{G}_{k}}(<t), \epsilon_t),
    \label{eqn:scm-m}
\end{equation}
where the function $f_k$ denotes the structural equation model and $g_k$ denotes the exogenous noise model. 
% \begin{align}
% % \vspace{-15mm}
% \textstyle
%     &\log p_\theta \left( X_{1:T}^{(1:N)} \right)  \nonumber\\
%     & \geq \sum_{n=1}^N \mathbb{E}_{q_\phi(\mathcal{M}_{1:K})}\Bigg[ \mathbb{E}_{r_\psi \left(Z^{(n)}\mid X^{(n)}_{1:T} \right)}\Big[ \log  
%  p_\theta \left(X^{(n)}_{1:T}\mid \mathcal{M}_{Z^{(n)}} \right)  \nonumber\\
%  &+ \log p\left(Z^{(n)}\right)  + H \left( r_\psi \left(Z^{(n)}\mid X^{(n)}_{1:T}\right) \right) \Big] \Bigg] \nonumber \\
% &+ \sum_{k=1}^K \mathbb{E}_{q_\phi(\mathcal{M}_k)} \left[\log p(\mathcal{M}_k) \right] + H \left( {q_\phi(\mathcal{M}_k)} \right) \label{eqn:elbo}  \\
% &\equiv \mathrm{ELBO}(\theta, \phi, \psi). \nonumber 
%     % &\equiv \text{ELBO}(\theta, \phi, \psi) \nonumber
% % \vspace{-20mm}
% \end{align}
\subsection{Variational inference}
Our goal is to infer the true posterior distribution $p \left(\mathcal{M}_{1:K} \mid X^{(1:N)} \right)$. However, it is intractable due to the presence of latent variables $\mathcal{M}_{1:K}$ and $\{Z^{(n)}\}$.  We propose a variational inference framework to infer the parameters of the data generation process. 

\begin{proposition}
    Under the data generation process described in Figure \ref{fig:pgm_assumption}, the data likelihood admits the following evidence lower bound (ELBO):
\begin{align}
% \vspace{-15mm}
    &\log p_\theta \left( X_{1:T}^{(1:N)} \right)  \nonumber\\
    & \geq \sum_{n=1}^N \mathbb{E}_{q_\phi(\mathcal{M}_{1:K})}\Bigg[ \mathbb{E}_{r_\psi \left(Z^{(n)}\mid X^{(n)}_{1:T} \right)}\Big[ \log  
 p_\theta \left(X^{(n)}_{1:T}\mid \mathcal{M}_{Z^{(n)}} \right)  \nonumber\\
 &+ \log p\left(Z^{(n)}\right)  + H \left( r_\psi \left(Z^{(n)}\mid X^{(n)}_{1:T}\right) \right) \Big] \Bigg] \hspace{18mm}\nonumber \\
&+ \sum_{k=1}^K \mathbb{E}_{q_\phi(\mathcal{M}_k)} \left[\log p(\mathcal{M}_k) \right] + H \left( {q_\phi(\mathcal{M}_k)} \right) \label{eqn:elbo}  \\
&\equiv \mathrm{ELBO}(\theta, \phi, \psi). \nonumber 
    % &\equiv \text{ELBO}(\theta, \phi, \psi) \nonumber
% \vspace{-20mm}
\end{align}
\end{proposition}
For derivation details, we refer the reader to Section \ref{sec:ELBO_der}. 
Here, $\log  
 p_\theta \left(X^{(n)}_{1:T}\mid \mathcal{M}_{Z^{(n)}} \right)$ represents the marginal likelihood of $X^{(n)}$ under model $\mathcal{M}_{Z^{(n)}}$, $q_\phi \left( \mathcal{M}_k \right)$ represents the variational distribution of the causal model $\mathcal{M}_k$, and $r_\psi(Z^{(n)}\mid X^{(n)})$ represents the variational posterior distribution of the mixing rate for sample $X^{(n)}$. The number of causal models $K$ is a hyperparameter. $p(Z)$ represents our prior belief about the membership of samples to the causal models, typically considered to be a uniform distribution. 
% Figure \ref{fig:summary} shows a summary of how the likelihood is evaluated for every data sample, given the learned variational distributions and the mixing rates. 

\subsection{Model implementation} \label{sec:model_implementation}

We describe how to parameterize the different loss terms in Eq. \eqref{eqn:elbo}. Figure \ref{fig:summary} shows the ELBO calculation for \ours{}.

\textbf{Causal model.} We parameterize the variational distribution of the causal models as $\displaystyle q_\phi \left( \mathcal{M}_{1:K} \right) = \prod_{k=1}^K q_{\phi_k} \left( \mathcal{G}_k\right) \delta(\Theta_k),$ where $\delta$ represents the Dirac-$\delta$ function, and $\Theta_k$ represents the (learned) parameters of the structural equations and noise models. The distribution of the DAG adjacency matrix $q_{\phi_k} \left( \mathcal{G}_k\right)$ is represented as a product of independent Bernoulli distributions. The expectation over $q_\phi \left( \mathcal{M}_{1:K} \right)$ is computed by sampling once through Monte-Carlo sampling, using the Gumbel-Softmax trick \citep{jang2016categorical}.  

\textbf{Mixing probabilities.} We specify the mixing rates variational distribution $r_\psi \left( Z^{(n)} \mid X^{(n)} \right)$ as a $K$-way categorical distribution, and learn it for each sample. We set
$\displaystyle r_\psi \left( Z^{(n)}\mid X^{(n)}\right) = \text{softmax} \left( \frac{w^{(n)}}{\tau_r} \right)$, where $\displaystyle w^{(n)} = \left[ w^{(n)}_1, \ldots, w^{(n)}_K\right] \in \mathbb{R}^{K}$ are learnable weight parameters for each sample, and $\tau_r$ is a temperature hyperparameter. The number of parameters in the learned membership matrix grows linearly with the number of samples. This dependence on the sample size can be eliminated using a classifier that inputs a $D\times T$ multivariate time series and outputs a categorical distribution over the mixture components. However, in practice, we observe that the number of parameters for the learned membership matrix $W$ is quite small.

% \begin{align*}
%     r_\psi \left( Z^{(n)} = k \mid X^{(n)}\right) = \frac{\exp \left( w^{(n)}_{k}/\tau_r\right) }{\sum_{k'=1}^{K} \exp \left( w^{(n)}_{k'} / \tau_r \right) },
% \end{align*}

We also need an expectation of the likelihood term over $r_\psi \left( Z^{(n)} \mid X^{(n)} \right)$. This involves computing the marginal likelihood of each sample under all $K$ causal models, theoretically requiring $K$ times more operations than using a single model. In practice, we can calculate the marginal likelihoods under all causal models in parallel using PyTorch vectorization. Empirically, the computational complexity increase over using a single model leads to a modest increase in run-time, much less than a factor of $K$ (Appendix \ref{sec:timing_analysis}). 

In theory, any likelihood-based causal structure learning algorithm can be used to implement the marginal likelihood loss term $\log p_\theta \left( X^{(n)}_{1:T} \mid \mathcal{M}_{Z^{(n)}}\right)$ in Eq. \eqref{eqn:elbo}. Appendix \ref{sec:implementation_details} details the computation of $\log p_\theta \left( X^{(n)}_{1:T} \mid \mathcal{M}_{Z^{(n)}}\right)$ based on the model for $f_k$ in Eq \eqref{eqn:scm-m}. 
We implement two variants of \ours{} to show the flexibility of our framework: (1) \ourslinear, which handles linear causal relationships and independent noise, and (2) \oursnonlinear, which handles nonlinear structural equations and history-dependent noise.

\textbf{\ourslinear.}  We implement each of the $K$ models using a linear model:
\begin{equation}
f_{k}^d \left(\text{Pa}_{\mathcal{G}_{k}}(\leq t)\right) = \sum_{\tau=0}^L \sum_{j=1}^D \left(\mathcal{G}_k \circ \mathcal{W}_k\right)_{\tau}^{j, d} \times
      X^{j, (n)}_{t-\tau},  
      \label{eqn:linear_model}
\end{equation}
where $\circ$ denotes the Hadamard product, and $\mathcal{W}_k \in \mathbb{R}^{(L+1)\times D \times D}$ is a learned weight tensor. The parameters of each SCM are given by $\Theta_k= \{\mathcal{W}_k\}$.
Since we do not model the history-dependence of the noise, we set $g_k(\text{Pa}_{\mathcal{G}_{k}}(<t), \epsilon_t) = \epsilon_t$, i.e., the identity function.

\textbf{\oursnonlinear.} We implement each of the $K$ causal models based on Rhino \citep{gong2022rhino} as it can handle instantaneous effects and history-dependent noise. We parameterize the structural equations $f_k$ in \eqref{eqn:scm} with embeddings $\mathcal{E}_k$, which are used in conjunction with neural networks, denoted by $\Xi_f$ and $\ell_f$:
\begin{eqnarray}
    f_{k}^d \left( \text{Pa}_{\mathcal{G}_{k}(\leq t)} \right) = \Xi_f \left( \left[ \sum_{\tau=0}^L \sum_{j=1}^D \left(\mathcal{G}_{k}\right)_{\tau}^{j, d} \times \right. \right. \nonumber \\ 
       \ell_f \left( \left[X^{j, (n)}_{t-\tau}, \left(\mathcal{E}_{k}\right)^{j}_{\tau} \right] \right),  \left(\mathcal{E}_{k}\right)^{d}_0 \Biggr] \Biggr). \label{eqn:mlp}
\end{eqnarray}
$\mathcal{E}_{k} \in \mathbb{R}^{(L+1)\times D \times e}$ are trainable embeddings (with embedding dimension $e$) corresponding to model $\mathcal{M}_k$, and $\Xi_f$ and $\ell_f$ are multi-layer perceptron networks that are shared across all nodes and causal models $\mathcal{M}_{1:K}$. The noise model $g_k(\text{Pa}_{\mathcal{G}_{k}}(<t), \epsilon_t)$ is described using conditional spline flow. The network that predicts parameters for the conditional spline flow model uses a similar architecture, utilizing embeddings $\mathcal{F}_k$ with neural networks $\Xi_\epsilon$ and $\ell_\epsilon$.  Thus, the SCM parameters are $\displaystyle \Theta_k = \left\{ \mathcal{E}_k, \mathcal{F}_k, \Xi_f, \ell_f, \Xi_\epsilon, \ell_\epsilon \right\}$.

\section{Theoretical analysis}
In this section, we examine (1) conditions under which the mixture model is identifiable (2) the relationship between the derived ELBO objective and the true data likelihood.

\textbf{Structural identifiability.}
We examine when the mixtures of SCM models are identifiable. Structural identifiability dictates that two distinct mixtures of SCMs cannot result in the same observational distribution.  Identifiability is an important statistical property to ensure that the causal discovery problem is meaningful \citep{peters2012identifiability}.

We establish a necessary and sufficient condition for the identifiability of mixtures of linear structural vector autoregressive models (SVARs) with Gaussian noise.

\begin{theorem}[Identifiability of linear SVARs with equal-variance additive Gaussian noise]
    Let $\mathcal{F}$ be a family of distributions of $K$ structural vector autoregressive (SVAR) models of lag $L \geq 1$ with zero-mean Gaussian noise of equal variance, i.e. 
\begin{align*}
\textstyle
&\mathcal{F} = \big\{ \mathcal{L}_{\mathcal{M}^{(k)}} :  \mathcal{M}^{(k)} \text{ is specified by the equations } \\
&\mathbf{X}_t = \mathbf{W}^{(k)}\mathbf{X}_t + \sum_{\tau=1}^L \mathbf{A}^{(k)}_\tau \mathbf{X}_{t-\tau} + \varepsilon^{(k)}, \\
&\varepsilon^{(k)} \sim \mathcal{N} \left( 0, \sigma^2 \mathbf{I} \right), 1 \leq k \leq K \big\} 
\end{align*}        
and let $\mathcal{H}_K$ be the family of all $K-$finite mixtures of elements from $\mathcal{F}$. Then the family $\mathcal{H}_K$ is identifiable if and only if the following condition is met: The ordered pairs
% \vspace{-1mm}
\begin{equation}
\textstyle
    \left( \left[\mathbf{B}^{(k)}\right]^{-1}\mathbf{A}_1^{(k)}, ..., \left[\mathbf{B}^{(k)}\right]^{-1}\mathbf{A}_L^{(k)},   \left[\mathbf{B}^{(k)}\right] \left[\mathbf{B}^{(k)}\right]^{T} \right) \\, 
\end{equation}
are distinct over all $k$, where $\displaystyle \mathbf{B}^{(k)} = \mathbf{I} - \mathbf{W}^{(k)}$.
\end{theorem}

To illustrate this condition, we consider the 2D SCM case when $L=0$, i.e. there are no lagged effects. The SCM is specified by $W^{(k)} = \begin{bmatrix} 0 & w_1^{(k)} \\ w_2^{(k)} & 0 \end{bmatrix}$, and the matrix $B^{(k)}$ takes the form $B^{(k)}=\begin{bmatrix} 1 & -w_1^{(k)} \\ -w_2^{(k)} & 1 \end{bmatrix}$, where $w_1^{(k)}w_2^{(k)}=0$ due to acyclicity. Then $\left[B^{(k)}\right]\left[B^{(k)}\right]^T = \begin{bmatrix} 1 + \left(w_1^{(k)}\right)^2 & -(w_1^{(k)}+w_2^{(k)}) \\ -(w_1^{(k)}+w_2^{(k)}) & 1 + \left(w_2^{(k)}\right)^2 \end{bmatrix}$, and the condition is violated iff $W^{(i)}=W^{(j)}$, i.e., they have the same SCM equations. The mixture of linear Gaussian SCMs is identifiable when the structural equation matrices are distinct.

We now examine the identifiability of mixtures of general SCMs. We derive an intuitive sufficient condition for mixture model identifiability in terms of the existence of $K$ representative points from the sample space $\mathbb{X}$. These points exhibit a key characteristic: their association with a particular causal model is unequivocal, as determined by their marginal likelihood functions of the mixture components. 

\begin{theorem}[Identifiability of finite mixture of causal models]
    Let $\mathcal{F}$ be a family of $K$ identifiable causal models, $\displaystyle \mathcal{F} = \left\{ \mathcal{L}_\mathcal{M}^{(k)} 
: \mathcal{M} \text{ is an identifiable causal model }, 1 \leq k \leq K\right\}$ and let $\mathcal{H}_K$ be the family of all $K-$finite mixtures of elements from $\mathcal{F}$, i.e. 
\begin{align*}
\textstyle
\displaystyle \mathcal{H}_K = \Biggl\{ h: h &= \sum_{k=1}^K \pi_k \mathcal{L}_{\mathcal{M}_k}, \mathcal{L}_{\mathcal{M}_k} \in \mathcal{F}, \\
&\pi_k > 0, \sum_{k=1}^K \pi_k = 1 \Biggr\}    
\end{align*}
where $\displaystyle \mathcal{L}_{\mathcal{M}_k}(x) = \sum_{\mathcal{M}} p(x\mid \mathcal{M}) p(\mathcal{M}_k = \mathcal{M})$  denotes the likelihood of $x$ evaluated with causal model $\mathcal{M}_k$. Further, assume that the following condition is met:
% \vspace{-2mm}

For every $k=1,\dots,K$, $\exists a_k \in \mathbb{X}$ such that
% \vspace{-5mm}
\begin{align*}
% \textstyle
\frac{\mathcal{L}_{\mathcal{M}_k}(a_k)}{\sum_{j=1}^K \mathcal{L}_{\mathcal{M}_j}(a_k)} > \frac{1}{2}. \tag{*} \label{eqn:diversity_condition}
\end{align*}
% \begin{align*}
% \text{For every } k, & \, 1 \leq k \leq K, \exists a_k \in \mathbb{X} \text{ such that } \\
% &\frac{\mathcal{L}_{\mathcal{M}_k}(a_k)}{\sum_{j=1}^K \mathcal{L}_{\mathcal{M}_j}(a_k)} > \frac{1}{2}. \tag{*} \label{eqn:diversity_condition}
% \end{align*}
% \vspace{-2mm}
Then the family $\mathcal{H}_K$ is identifiable.
\end{theorem} 
Appendix \ref{sec:identifiablity} contains the relevant definitions and proofs.
To draw a parallel with clustering, this implies that each cluster has at least one point whose membership can be established with a high level of certainty to that specific cluster. Directly verifying the condition \eqref{eqn:diversity_condition} is generally difficult because we rarely know the exact form of the likelihood function. However, this condition can be verified approximately using the estimated likelihood functions for each mixture component, as with our approach, \ours{}. The validity of this verification critically depends on how closely the estimated likelihood function approximates the true likelihood function.

Furthermore, as a direct consequence of the structural identifiability of the Rhino model \cite{gong2022rhino}, a mixture of Rhino models is also identifiable, provided that the assumptions in Section \ref{sec:theory_assumptions} and condition $\left(\ref{eqn:diversity_condition}\right)$ are satisfied.

\textbf{Relationship between ELBO and log-likelihood.} We verify the soundness of our derived ELBO objective in Eq. \eqref{eqn:elbo}. By maximizing the ELBO, we can simultaneously learn the $K$ underlying causal graphs, their associated functional equations, and the membership of each sample to its respective causal model. We show that (Appendix \ref{sec:elbo_loglikelihood}):
% \vspace{-1mm}
 %  \begin{align*}
 %  \textstyle
 %     &\log p_\theta(X) = \text{ELBO}(\theta, \phi, \psi) \\ 
 %     & \textstyle{+ \sum_{n=1}^N \mathbb{E}_{q_\phi(\mathcal{M}_{1:K})} } \Big[\\ 
 %     &\textstyle{\text{KL} \left( r_\psi \left(Z^{(n)} | X^{(n)} \right) \| p(Z^{(n)} | X^{(n)}, \mathcal{M}_{1:K})\right) }\Big] \\
 %     &\textstyle{+ \text{KL}\left(q_\phi\left(\mathcal{M}_{1:K}\right) \|   p\left(\mathcal{M}_{1:K} | X\right)\right).}
 % \end{align*}
   \begin{align*}
     &\log p_\theta(X) = \text{ELBO}(\theta, \phi, \psi) \\ 
     & {+ \sum_{n=1}^N \mathbb{E}_{q_\phi(\mathcal{M}_{1:K})} } \Big[\\ 
     &{\text{KL} \left( r_\psi \left(Z^{(n)} | X^{(n)} \right) \| p(Z^{(n)} | X^{(n)}, \mathcal{M}_{1:K})\right) }\Big] \\
     &{+ \text{KL}\left(q_\phi\left(\mathcal{M}_{1:K}\right) \|   p\left(\mathcal{M}_{1:K} | X\right)\right).}
 \end{align*}
 Maximizing $\text{ELBO}(\theta, \phi, \psi)$ with respect to $(\theta, \phi, \psi)$ is equivalent to jointly (1) maximizing the log-likelihood $\log p_\theta(X)$ (2) minimizing the KL divergence between the variational distribution $q_\phi \left( \mathcal{M}_{1:K} \right)$ and the true posterior $p \left(\mathcal{M}_{1:K} \mid X \right)$, and (3) minimizing the expectation, under the variational distribution $q_\phi(\mathcal{M}_{1:K})$, of the KL divergence between the variational posterior $r_\psi \left(Z^{(n)} \mid X^{(n)} \right)$ for each sample $X^{(n)}$ and the true posterior for mixture component selection $p \left(Z^{(n)} \mid X^{(n)}, \mathcal{M}_{1:K} \right)$.

\section{Experiments}
% We benchmark the causal discovery performance of \ours{} on synthetic and real-world datasets to demonstrate the benefits of learning mixtures of DAGs. \sv{I feel like we should just omit this}

\subsection{Experimental setup}
We train the model on $80\%$ of the data and validate on the remaining $20\%$. We pick the model with the lowest validation likelihood and evaluate the corresponding causal graphs. Details about model validation are in Appendix \ref{sec:validation}.

\textbf{Baselines.}
\label{sec:baselines}
We benchmark against several state-of-the-art temporal causal discovery methods, including Rhino \citep{gong2022rhino}, PCMCI$^+$ \citep{runge2020discovering}, DYNOTEARS \citep{pamfil2020dynotears}, and VARLiNGaM \citep{hyvarinen2010estimation}. PCMCI$^+$ and DYNOTEARS can be used with two options - one where the algorithm predicts one causal graph per sample and one where the algorithm predicts one graph to explain the whole dataset. We denote these options with suffixes -s and -o, respectively. Since these baseline methods cannot discover mixtures of causal graphs, we also report results by grouping samples by their true causal graph. We then predict one causal graph per group. This option is reported for PCMCI$^+$, DYNOTEARS, and Rhino and is denoted by the suffix -g in the results. Appendix \ref{sec:pcmci_process} details the steps for post-processing PCMCI$^+$'s output.

In practice, the number of mixture components, which we treat as a hyperparameter, is often unknown. We use $K^\ast$ to denote the true number of SCMs, and $K$ to represent the input to \ours{}. We report the clustering accuracy for \ours{} in addition to traditional causal discovery metrics like {orientation} F1 score and  AUROC (Area Under the Receiver Operator Curve). We define clustering accuracy as:
\begin{equation*}
    \text{Cluster Acc.} \left( \tilde{Z}, Z\right) = \max_{\pi \in S_{K}} \frac{1}{N} \sum_{n=1}^N \mathrm{1}\left( \pi(\tilde{Z}_n) = Z_n \right),
\end{equation*}
where $\tilde{Z}$ are the assigned mixture labels and $Z$ are the true labels. We refer the reader to Appendix \ref{sec:cluster_acc} for details about the calculation of clustering accuracy.

\subsection{Datasets}
\begin{figure*}
    \centering
    \includegraphics[width=0.8\textwidth]{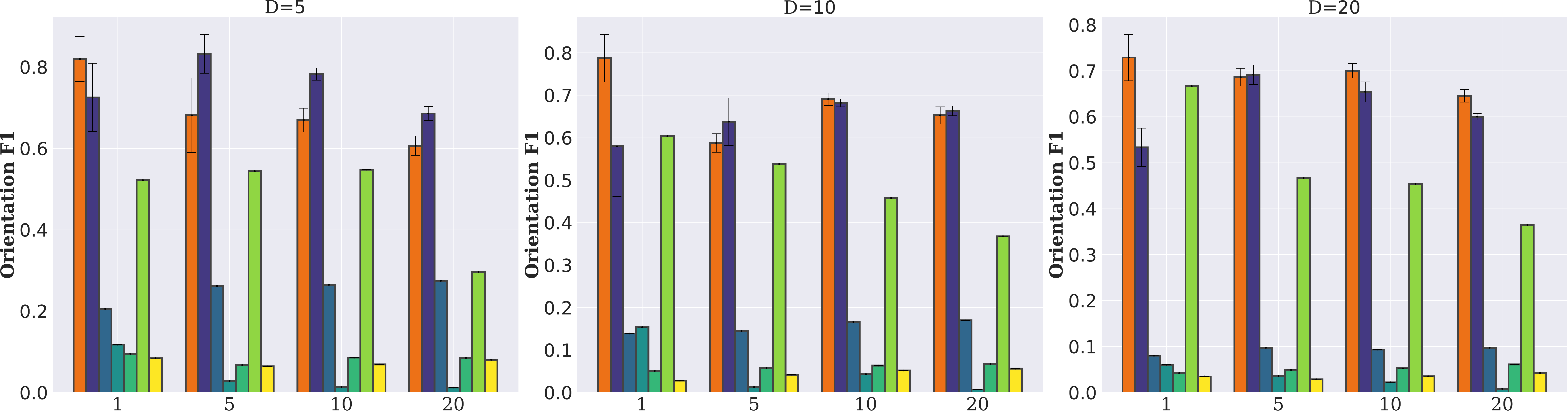}
    
    \includegraphics[width=0.8\textwidth]{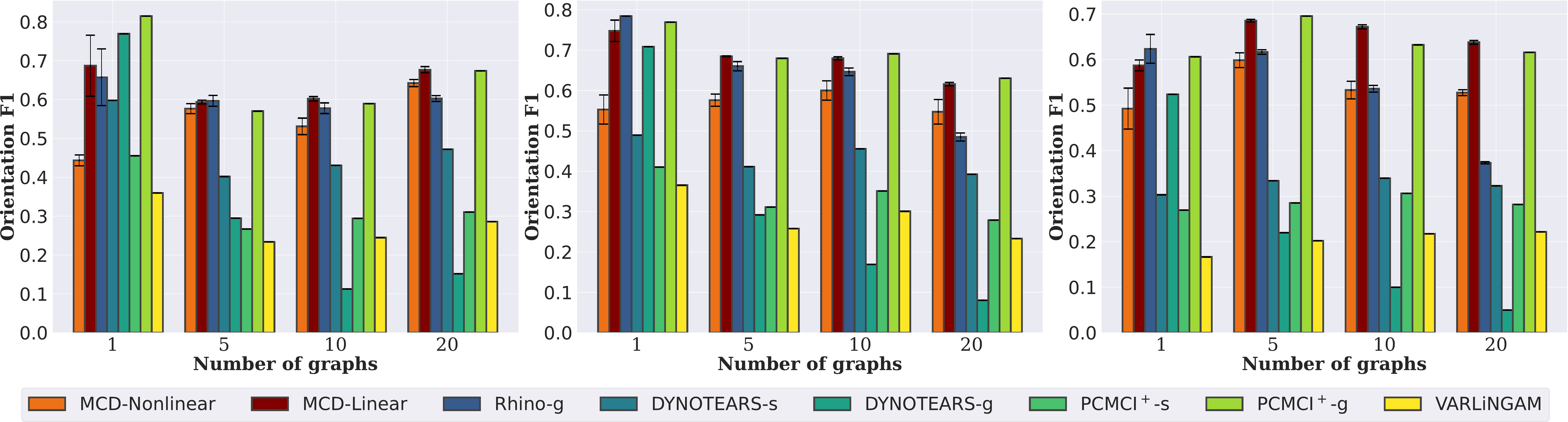}
    
    \caption{Results on the \textbf{nonlinear} (top) and  \textbf{linear} (bottom) synthetic datasets for dimension $D=5, 10, 20$. We report the orientation F1 scores. (-s) indicates that the baseline predicts one graph per sample. (-g) signifies that the baseline was executed on samples grouped according to the ground truth causal graph. These methods use additional information that \ours{} does not.  Average of {5 runs} reported.}
\label{fig:results_syn}

\end{figure*}

\label{sec:datasets}

% \ry{If running out of space, shorten the dataset description and move to appendix}
\textbf{Synthetic datasets.}
% To test the problem setting, we modified the synthetic dataset generation code from Rhino \citep{gong2022rhino}. 
% \sv{TODO: add description of linear datasets}
We generate a pool of $K^\ast$ random graphs (specifically, Erdős-Rényi graphs) and treat them as ground-truth causal graphs. To generate a sample $X^{(n)}$, we first randomly sample a graph $\mathcal{G}_k$ from this pool and use it to model relationships between the variables. We experiment with $D=5, 10, 20$ nodes. For each value of $D$, we generate datasets with $K^\ast=1, 5, 10, 20$ graphs having $N=1000$ samples each. The time series length $T$ is 100, and the time lag $L$ is set to $2$ for all the methods, which is the lag value used to simulate the data. The number of causal graphs $K$ is set to $2K^*$. We experiment with two sets of synthetic datasets:
(1) Linear datasets, in which linear causal relationships are modeled with Gaussian noise;
(2) Nonlinear datasets, in which the functional relationships are modeled as randomly generated multi-layer perceptions with history-dependent noise.
We refer the reader to Appendix \ref{sec:syn_data_setup} for more details about the setup.

\textbf{Netsim Brain Connectivity.}
The Netsim benchmark dataset \citep{smith2011network} consists of simulated blood oxygenation level-dependent (BOLD) imaging data. Each variable represents a region of the brain, with the goal being to infer the interactions between the different regions. The dataset has $28$ different simulations, which differ in the number of variables and time length over which the measurements are recorded. In our experiments, we consider the samples from simulation 3 comprising $N=50$ time series, each with $D=15$ nodes and $T=200$ timepoints. These samples share the same ground-truth causal graph. We introduce heterogeneity by considering a pool of $K^\ast=3$ random permutations and applying a randomly chosen one to the nodes of each sample and its corresponding ground truth causal graph. This setup is denoted as \textbf{Netsim-mixture}. We use a uniform prior for $p(Z)$ and set $L=2$ and $K=5$.

% \begin{wrapfigure}[16]{r}{0.4\textwidth}
%     \centering
%     \includegraphics[width=0.35\textwidth]{images/cluster_acc_syn.pdf}
%     \caption{Clustering accuracy for \ours{} on the synthetic datasets (vs) the number of causal graphs $K^\ast$. The accuracy is averaged across \change{5 runs} and across data dimensionality $D=5, 10, 20$. Hyperparameter $K$ is set to $2K^\ast$ for all settings.}
%     \label{fig:cluster_acc_syn}
% \end{wrapfigure}

\begin{figure}
    \centering
    
    \includegraphics[width=0.22\textwidth]{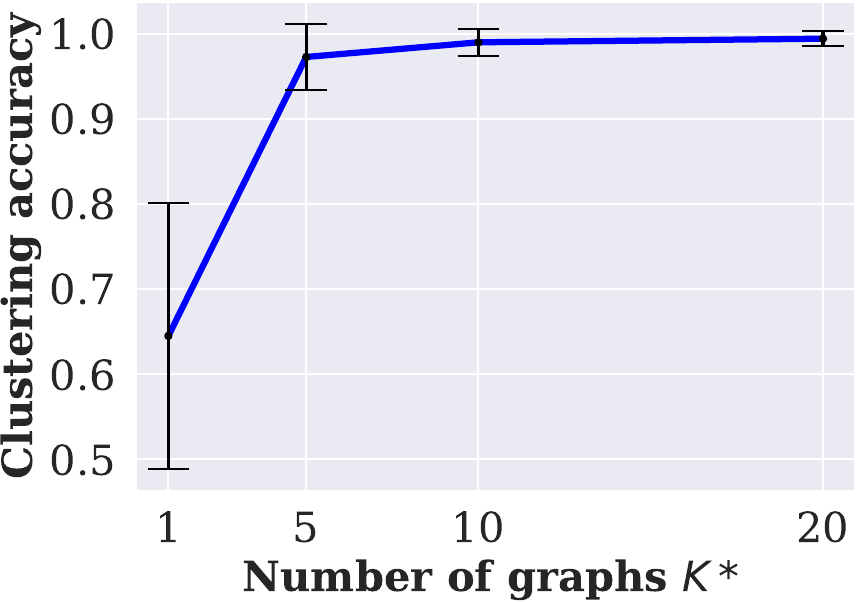}
    \includegraphics[width=0.22\textwidth]{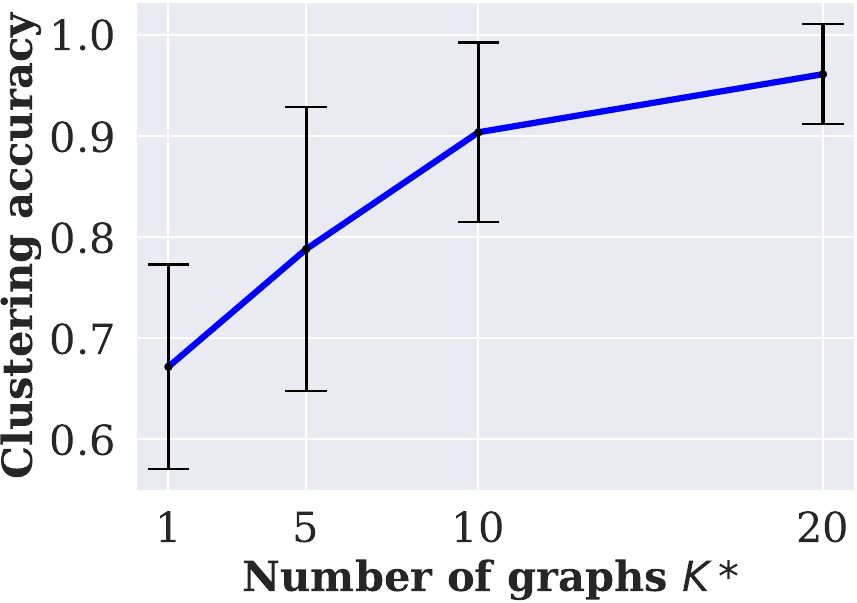}
    \caption{Clustering accuracy of  (left) \oursnonlinear{} on the nonlinear synthetic datasets (right) \ourslinear{} on the linear synthetic datasets, as a function of the true number of causal graphs $K^*$. The accuracy is averaged across 5 runs and data dimensionality $D=5, 10, 20$. Hyperparameter $K$ is set to $2K^\ast$ for all settings.}
    \label{fig:cluster_acc_syn}
\end{figure}

\begin{figure*}
    \centering
    
    \includegraphics[width=0.45\textwidth]{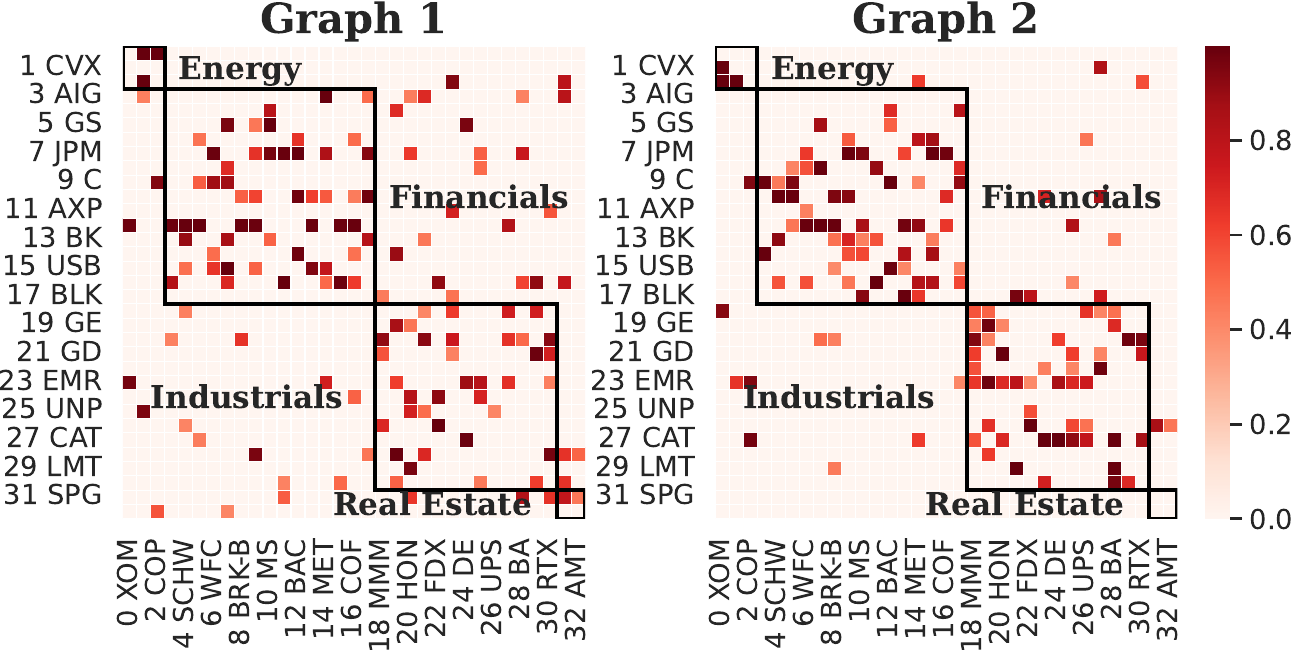}\includegraphics[width=0.45\textwidth]{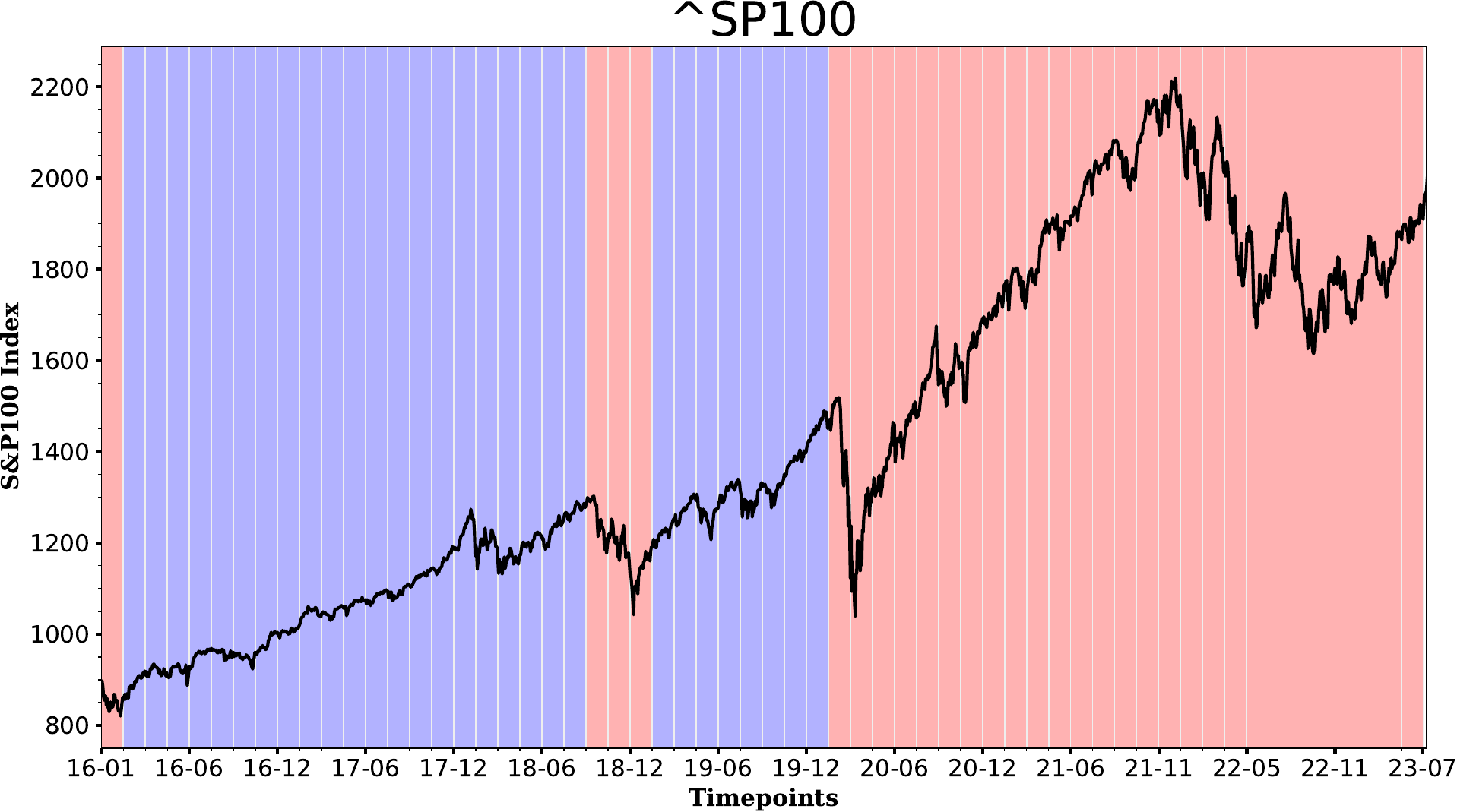}
    \caption{
    (Left) Sub-graphs of the two inferred graphs from \oursnonlinear{}
    (Right) S\&P 100 index overlaid with the mixture membership from \oursnonlinear{}. Red indicates that Graph 1 from Figure \ref{fig:snp100_graphs} is used, while blue indicates that Graph 2 is active.} 
    \label{fig:snp100-results}
\end{figure*}
\textbf{DREAM3 Gene Network.}
The DREAM3 dataset  \citep{prill2010towards} is a real-world biology dataset consisting of measurements of gene expression levels obtained from yeast and E.coli cells. There are 5 distinct ground-truth networks, comprising 2 for E.coli and 3 for Yeast, each with $D=100$ nodes. Each time series consists of $T=21$ timesteps, with $46$ trajectories recorded per graph. Thus, a total of $N=230$ samples are combined across all the networks. We mix samples from all 5 networks to simulate the scenario in which the identity of the cell from which the data is obtained is unknown. This is a challenging dataset due to the high dimensionality of the data and the small number of available samples. We set the time lag $L=2$ and $K=10$. 
We discuss how we post-process the model outputs on the Netsim and DREAM3 datasets in Appendix \ref{sec:netsim_dream3}.

\textbf{S\&P 100.} We also run \ours{} on daily stock returns of companies from the S\&P 100 index. We use the \texttt{yahoofinancials} package to retrieve the daily closing prices of $D=100$ stocks from January 1, 2016 to July 1, 2023. Similar to the setup in \citet{pamfil2020dynotears}, we use log-returns, i.e., differences of the logarithm of the closing prices of successive days. In addition, we normalize the log-returns to have zero mean and unit variance. We chunk the data into segments of length $T=31$ each, resulting in $N=60$ samples. We train our model on the first $48$ samples and validate with the last $12$ samples. Following \citet{pamfil2020dynotears}, we set $L=1$. We set $K=5$ and threshold the edge probabilities at 0.4. We qualitatively analyze the results on this dataset since it lacks ground truth causal graphs.

\subsection{Results on synthetic and real-world datasets}

\begin{table}
 \resizebox{0.5\textwidth}{!}{
\begin{tabular}{|c|cc|cc|}
\hline
                                               & \multicolumn{2}{c|}{\textbf{Netsim-mixture}}                                                             & \multicolumn{2}{c|}{\textbf{DREAM3}}                                                                     \\ \hline
\textbf{Method}               & AUROC$\left(\big \uparrow \right)$ & \textbf{ F1}$\left(\big \uparrow \right)$ & AUROC$\left(\big \uparrow \right)$ & \textbf{F1} $\left(\big \uparrow \right)$ \\ \hline
PCMCI$^+$-s                                    & $0.82$                             & ${0.67}$                                                   & $0.50$                             & $0.01$                                                     \\
PCMCI$^+$-o                                    & $0.71$                             & $0.49$                                                     &                                    0.51&                                                            0.04\\
PCMCI$^+$-g                                    & $0.72$                             & $0.52$                                                     & $0.51$                             & $0.05 $                                                    \\
VARLiNGAM                                      & $0.78$                             & $0.60$                                                     & NA                             & NA                                                     \\
DYNOTEARS-s                                    & $0.85$                             & $0.28$                                                     & $0.50$                             & $0.03$                                                     \\
DYNOTEARS-o                                    & $0.83$                             & $0.45$                                                     & $ 0.50$                            & $0.03 $                                                    \\
DYNOTEARS-g                                    & $0.85$                             & $0.46$                                                     & $0.50$                                   &                       $0.03$                                     \\
Rhino                                          & $0.84 \pm 0.01$                    & $0.62 \pm 0.01$                                            & $0.57 \pm 0.01$                    & $0.08 \pm 0.01$                                            \\
\oursnonlinear{} (this paper) & $\mathbf{0.94 \pm 0.03}$                    & $\mathbf{0.69 \pm 0.08}$                                   & $\mathbf{0.58 \pm 0.01}$           & $\mathbf{0.10 \pm 0.01}$                                   \\
\ourslinear{} (this paper)  & \multicolumn{1}{l}{$0.73 \pm 0.02$}               & \multicolumn{1}{l|}{$0.62 \pm 0.02$}                                      &       $0.51 \pm 0.01$                              &  $0.00 \pm 0.00$                                                          \\ \hline
\end{tabular}
}\caption{Results on Netsim-mixture and DREAM3. -s indicates that the baseline predicts one graph per sample. -o indicates that the baseline predicts one graph for the whole dataset. -g signifies that the baseline is run on samples grouped according to the ground truth causal graph. VARLiNGAM does not run on the DREAM3 dataset. \oursnonlinear{} achieves a clustering accuracy of $86.8 \pm 26.3 \%$ on Netsim-mixture and $95.6 \pm 4.8 \%$ on DREAM3. }
\label{tab:netsim-dream3}
\end{table}

\textbf{Synthetic datasets results.} Results are presented in Figure \ref{fig:results_syn}. On the nonlinear dataset, \oursnonlinear{} handily outperforms all the baselines except Rhino-g. Notably, it performs better than PCMCI$^+$-g, even though PCMCI$^+$ has additional information (i.e., ground truth membership information) that \oursnonlinear{} does not. \oursnonlinear{} achieves comparable, and sometimes better, performance than Rhino-g, especially on the $D=10$ and $D=20$ datasets. The baseline variants that predict one graph per sample perform poorly as expected since one sample does not provide adequate information to infer all the causal relationships. DYNOTEARS and VARLiNGAM, which assume that the causal relationships are linear, perform poorly on these datasets. We also omit \ourslinear{} for this reason.

On the linear dataset, \ourslinear{} achieves a similar or better F1 score than the grouped baseline methods when the number of graphs $K^\ast$ is 5, 10, 20. It achieves a comparable level of performance to the baselines for $K^\ast=1$ despite the misspecification of the number of graphs. \ourslinear{} expectedly outperforms \oursnonlinear{} across all settings since its model matches the parametric form of the data generation process. Nevertheless, \oursnonlinear{} performs better than all the baselines that do not use ground truth membership information. Although VARLiNGAM is a linear model, it performs poorly since it cannot handle linear SCMs with Gaussian noise. 

We report the clustering accuracy for \ours{} in Figure \ref{fig:cluster_acc_syn} for different values of $K^\ast$, averaged over the data dimensionalities $D=5, 10, 20$. \oursnonlinear{} achieves near-perfect clustering for scenarios with multiple underlying graphs. \ourslinear{} also achieves strong clustering performance, although it shows high variability across $D$.  The low clustering accuracy and relatively low F1 and AUROC scores for $K^\ast=1$ are explained by the observation that \ours{} learns two similar mixture components to explain the single underlying mode in the distribution. 

\textbf{Netsim-mixture results.} The results on the Netsim dataset are presented in Table \ref{tab:netsim-dream3}. 
\oursnonlinear{} outperforms all baselines as measured by AUROC and F1 scores. This setting illustrates the benefits of modeling heterogeneity, even when it comes from a simple permutation of nodes. In this setting, \oursnonlinear{} achieves a clustering accuracy of $86.8 \pm 26.3 \%$, highlighting its ability to accurately group samples when the underlying causal models are sufficiently diverse. \ourslinear{} learns a single mode for the dataset and achieves similar results to Rhino.

\textbf{DREAM3 results.}
The results on the DREAM3 dataset are presented in Table \ref{tab:netsim-dream3}. All methods fare poorly at inferring the causal relationships. However, out of all the considered baselines, \oursnonlinear{} achieves relatively better performance in terms of AUROC and F1 score. It is especially encouraging that \oursnonlinear{} can accurately cluster samples by their causal models, with a remarkable clustering accuracy of $95.6 \pm 4.8 \%$. \ourslinear{} infers a single mode from the dataset.

% \begin{figure}
%     \centering
%     \includegraphics[width=0.5\textwidth]{images/subset_graphs.pdf}
%     % \includegraphics[width=0.23\textwidth]{images/subset_graph1.pdf}
    
%     \caption{Sub-graphs of the two inferred causal graphs from \oursnonlinear{}. \sv{refer to figure 6}}
%     \label{fig:snp100-subsets}
% \end{figure}
\textbf{S\&P100 results.}
\oursnonlinear{} infers two distinct causal graphs for the S\&P100 dataset. We aggregate the adjacency matrices across time as described in Appendix \ref{sec:netsim_dream3}. Figure \ref{fig:snp100-results} shows subgraphs of the two discovered causal graphs for stocks in the energy, financials, industrials, and real estate sectors. The model identifies that companies from the same sector interact more than those across sectors, which is evident from the block-diagonal structure of the inferred graphs. Further, the two inferred causal graphs have important differences; e.g., Graph 1 shows more interactions between the financial and industrial sectors than Graph 2, and the direction of the causal influences for \texttt{XOM} in the energy sector is reversed between the two graphs. 

We also overlay the inferred mixture membership for each sample onto the S\&P100 index in Figure \ref{fig:snp100-results} (right). We note that the model automatically identifies that several consecutive time windows are governed by the same causal graph. In addition, the changes in the governing causal graph successfully capture several important events. For example, the stock market crashes in December 2018 and March 2020 (due to COVID-19) are captured by the red (first) causal graph. The `blue' periods, in which Graph 2 is active, exhibit relatively less pronounced trends. Additionally, Graph 2 is much sparser than Graph 1. We show the full causal graphs and interesting patterns that \oursnonlinear{} captures in selected stocks in Appendix \ref{sec:snp100_more_results}. 

\subsection{Ablation Studies}
\begin{figure}
    \centering
    \includegraphics[width=0.155\textwidth]{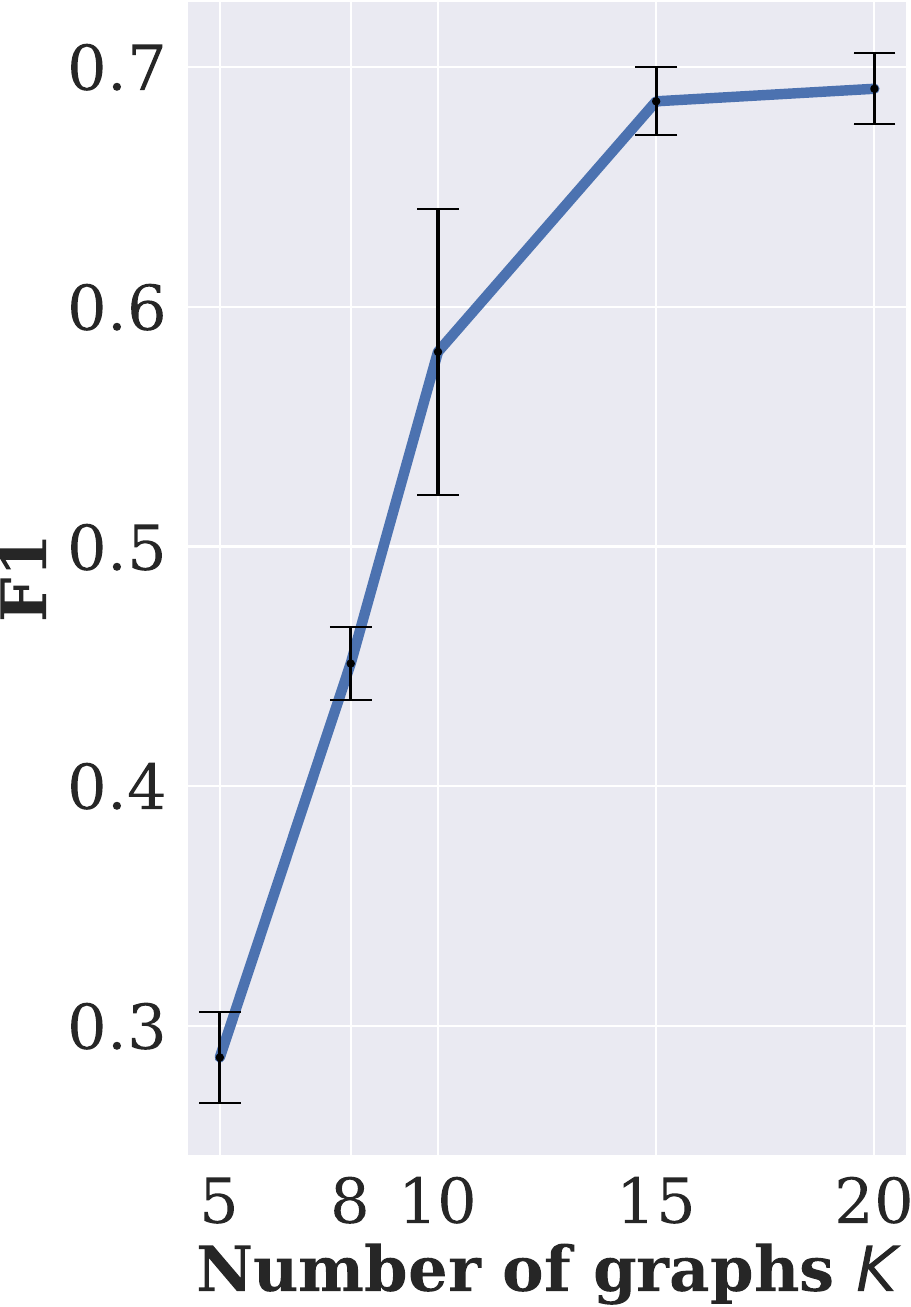}
    \includegraphics[width=0.155\textwidth]{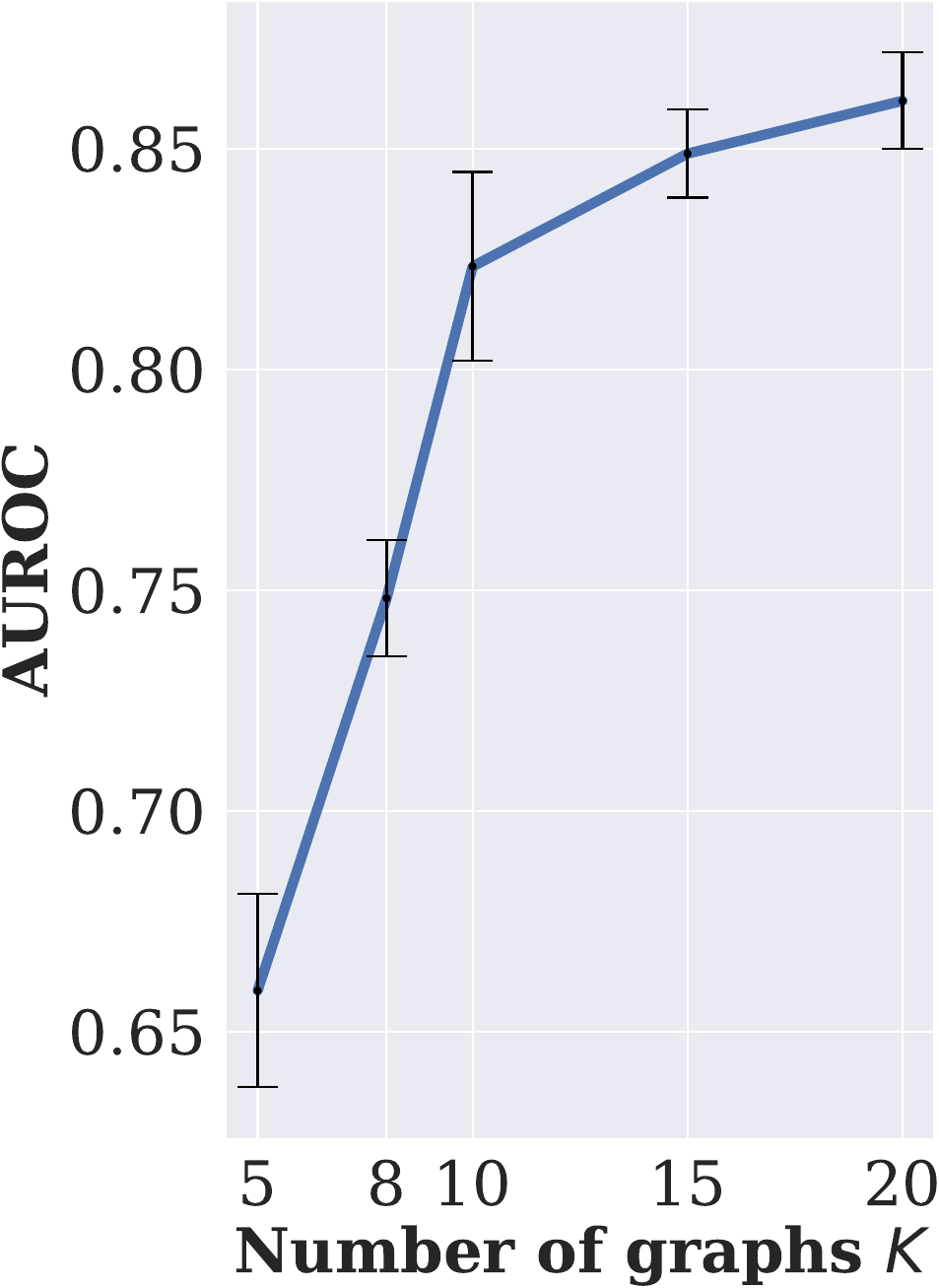}
    \includegraphics[width=0.155\textwidth]{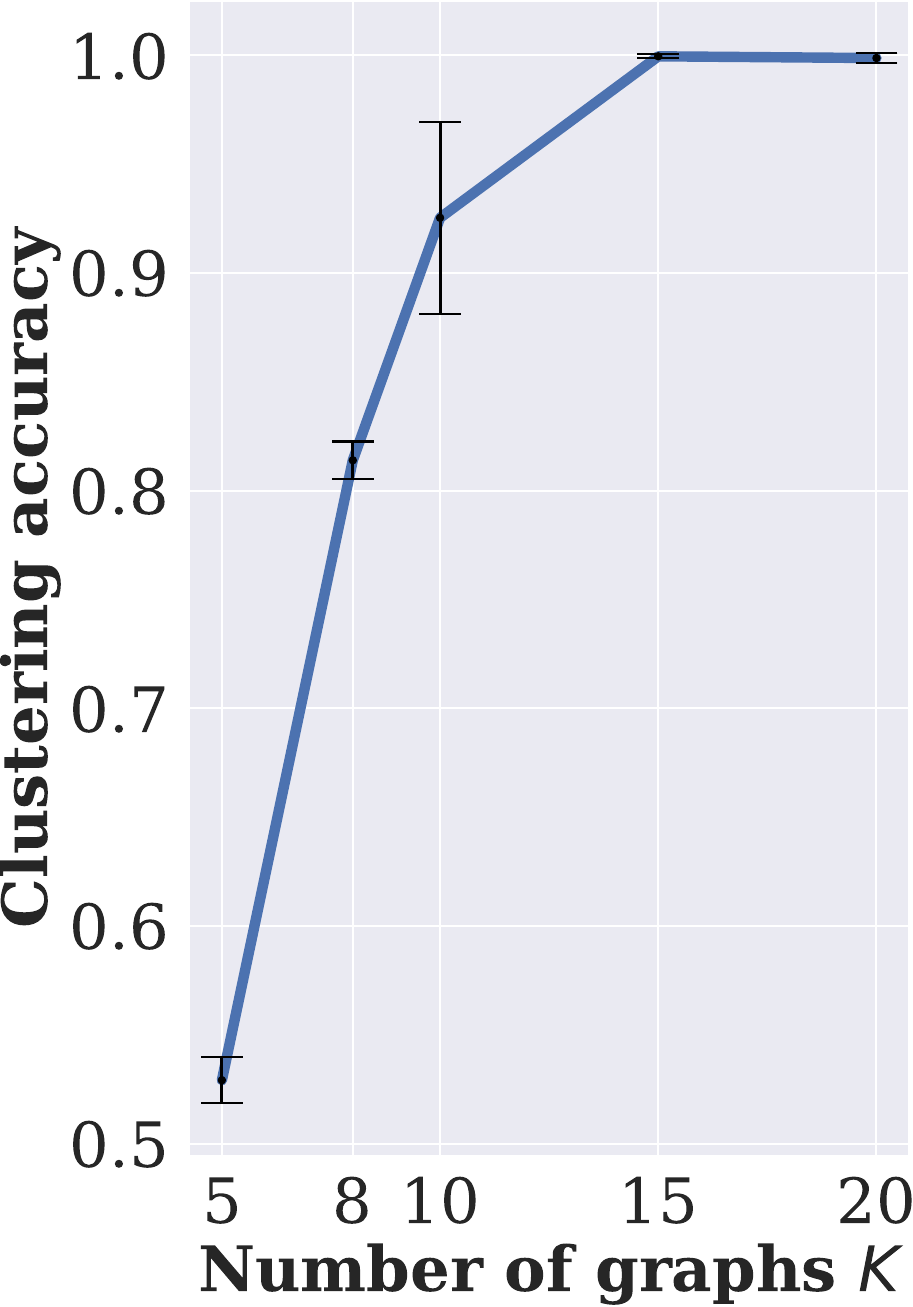}
    
    \caption{Performance of \oursnonlinear{} as a function of hyperparameter input $K$ on synthetic data with $D=10, K^\ast=10$ . Surprisingly, \oursnonlinear{} performs better when the number of graphs is overspecified. }
    \label{fig:ablation_robustness}
    % \vspace{-3mm}
\end{figure}

\textbf{Robustness of \ours{} to the misspecification of number of components.}  We examine the performance of \ours{} when the number of mixture components $K$ is misspecified, and does not equal the true number of underlying components $K^\ast$. Figure \ref{fig:ablation_robustness} shows the performance of \oursnonlinear{} as a function of $K$ on the nonlinear synthetic dataset with dimensionality $D=10$ and ground truth number of graphs $K^\ast=10$. We note that when the number of models is underspecified, our model performs poorly as expected since it cannot fully explain all the modes in the data. Surprisingly, the performance increases with increasing $K$. The clustering accuracy and performance metrics show high standard deviation when $K$ is set to the true number of mixture components $K^\ast=10$. While some random seeds achieve high clustering accuracy, others tend to saturate at a suboptimal grouping when $K=K^\ast$. On the other hand, when $K>K^\ast$, the additional SCMs are used as `buffers,' and the correct grouping is learned during the later epochs as the SCMs are inferred more accurately.   {This phenomenon is further explored in Appendix \ref{sec:cluster_over_steps}}

We perform additional ablation studies on synthetic datasets to investigate the behavior of \ours{} as $K^\ast$ increases, compare causal discovery performance using ground-truth membership assignments, and examine the impact of the similarity of the causal graphs. Details of these studies can be found in Appendix \ref{sec:ablation}.

\section{Conclusion and Discussion}
 In this work, we examine the problem of discovering mixtures of structural causal models from time series data. This problem has far-reaching applications in climate, finance, and healthcare, among other fields, since multimodal and heterogeneous data is ubiquitous in practice. We propose \ours{}, an end-to-end variational inference method, to learn both the underlying SCMs and the mixture component membership of each sample. We demonstrate the empirical efficacy of our method on both synthetic and real-world heterogeneous datasets. We conduct ablation studies on synthetic datasets to investigate \ours{}'s behavior with varying numbers of causal graphs, its robustness to misspecification of the number of graphs, and the impact of similarity among the causal graphs associated with each mixture component. In addition, we discuss the structural identifiability of mixtures of causal models. Future work could tackle data with latent confounders and non-stationarity in time.   

\section*{Impact Statement}
This paper presents work whose goal is to advance the field of Machine Learning. There are many potential societal consequences of our work, none which we feel must be specifically highlighted here.

\section*{Acknowledgement}
This work was supported in part by the U.S. Army Research Office under Army-ECASE award W911NF-07-R-0003-03, the U.S. Department Of Energy, Office of Science, IARPA HAYSTAC Program, NSF Grants SCALE MoDL-2134209, CCF-2112665 (TILOS), \#2205093, \#2146343,  \#2134274,  CDC-RFA-FT-23-0069 and DARPA AIE FoundSci.
% \newpage

\balance
\bibliography{ref}
\bibliographystyle{icml2024}
% \bibliographystyle{iclr2024_conference}
%%%%%%%%%%%%%%%%%%%%%%%%%%%%%%%%%%%%%%%%%%%%%%%%%%%%%%%%%%%%%%%%%%%%%%%%%%%%%%%
%%%%%%%%%%%%%%%%%%%%%%%%%%%%%%%%%%%%%%%%%%%%%%%%%%%%%%%%%%%%%%%%%%%%%%%%%%%%%%%
% APPENDIX
%%%%%%%%%%%%%%%%%%%%%%%%%%%%%%%%%%%%%%%%%%%%%%%%%%%%%%%%%%%%%%%%%%%%%%%%%%%%%%%
%%%%%%%%%%%%%%%%%%%%%%%%%%%%%%%%%%%%%%%%%%%%%%%%%%%%%%%%%%%%%%%%%%%%%%%%%%%%%%%

%%%%%%%%%%%%%%%%%%%%%%%%%%%%%%%%%%%%%%%%%%%%%%%%%%%%%%%%%%%%%%%%%%%%%%%%%%%%%%%
%%%%%%%%%%%%%%%%%%%%%%%%%%%%%%%%%%%%%%%%%%%%%%%%%%%%%%%%%%%%%%%%%%%%%%%%%%%%%%%

\newpage
\onecolumn
\appendix
\section{Theory}
\subsection{ELBO derivation} \label{sec:ELBO_der}

\newtheorem{innercustomthm}{Theorem}
\newenvironment{customthm}[1]
  {\renewcommand\theinnercustomthm{#1}\innercustomthm}
  {\endinnercustomthm}

\newtheorem{innercustomprop}{Proposition}
\newenvironment{customprop}[1]
  {\renewcommand\theinnercustomprop{#1}\innercustomprop}
  {\endinnercustomprop}

\newtheorem{innercustomdefn}{Definition}
\newenvironment{customdefn}[1]
  {\renewcommand\theinnercustomdefn{#1}\innercustomdefn}
  {\endinnercustomdefn}

\begin{customprop}{1}
      Under the data generation process described in Figure \ref{fig:pgm_assumption}, the data likelihood admits the following evidence lower bound (ELBO):
    \begin{align*}
        &\log p_\theta \left( X_{1:T}^{(1:N)} \right) \nonumber \\
        & \geq \sum_{n=1}^N \mathbb{E}_{q_\phi(\mathcal{M}_{1:K})}\Bigg[ \mathbb{E}_{r_\psi \left(Z^{(n)}\mid X^{(n)}_{1:T} \right)}\Big[ \log  
     p_\theta \left(X^{(n)}_{1:T}\mid \mathcal{M}_{Z^{(n)}} \right) + \log p\left(Z^{(n)}\right) \Big] + H \left( r_\psi \left(Z^{(n)}\mid X^{(n)}_{1:T}\right) \right)  \Bigg] \nonumber \\
    &+ \sum_{i=1}^K \mathbb{E}_{q_\phi(\mathcal{M}_i)} \left[\log p(\mathcal{M}_i)\right] + H \left( {q_\phi(\mathcal{M}_i)} \right)  \label{eqn:elbo} 
        % &\equiv \text{ELBO}(\theta, \phi, \psi) \nonumber
    \end{align*}
\end{customprop}

\begin{proof}
    
\begin{figure}
    \centering
    \includegraphics[width=0.3\textwidth]{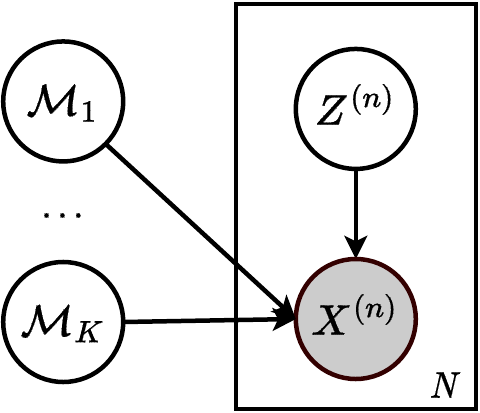}
    \caption{The assumed data generation model. First, the mixture index $Z^{(n)}$ is drawn from a $K$-way categorical distribution ($Z^{(n)} \sim \text{Cat}(K), Z^{(n)} \in \left\{ 1, \ldots, K\right\}$), and a causal model is drawn from the corresponding mixture component distribution $ \mathcal{M} \sim p \left(\mathcal{M}_{Z^{(n)}} \right)$. A sample $X^{(n)}$ is then drawn in according to the chosen causal model $\mathcal{M}$.}
    \label{fig:enter-label}
\end{figure}
Denote the causal models as $\mathcal{M}
_{1:K} = (\mathcal{M}_1,\ldots, \mathcal{M}_K )$ and the sample $X = \left\{ X^{(n)}\right\}_{n=1}^N$.
Then, we can write the log-likelihood under the assumed model as follows:
\begin{align*}
    \log p_\theta(X) &= \log \left[ \sum_{\mathcal{M}_{1:K}} p_\theta \left(X\mid  \mathcal{M}_{1:K}\right) p(\mathcal{M}_{1:K}) \times \frac{q_\phi(\mathcal{M}_{1:K})}{q_\phi(\mathcal{M}_{1:K})} \right] \\
    &= \log \mathbb{E}_{q_\phi(\mathcal{M}_{1:K})} \left[ \frac{p_\theta \left(X\mid  \mathcal{M}_{1:K} \right) p(\mathcal{M}_{1:K})} {q_\phi(\mathcal{M}_{1:K})} \right] \\
    & \geq  \mathbb{E}_{q_\phi(\mathcal{M}_{1:K})}\left[ \log   \frac{p_\theta \left(X\mid  \mathcal{M}_{1:K} \right) p(\mathcal{M}_{1:K})} {q_\phi(\mathcal{M}_{1:K})} \right]  \qquad \text{(using Jensen's inequality)}\\
    & =  \mathbb{E}_{q_\phi(\mathcal{M}_{1:K})}\left[ \log   p_\theta \left(X\mid  \mathcal{M}_{1:K} \right) + \log p(\mathcal{M}_{1:K}) - \log {q_\phi(\mathcal{M}_{1:K})} \right]
\end{align*}
Since the sample points are conditionally independent given the causal models, we can write:
\begin{align*}
    \log p_\theta(X) &\geq \sum_{n=1}^N \mathbb{E}_{q_\phi(\mathcal{M}_{1:K})}\left[ \log   p_\theta \left(X^{(n)}\mid  \mathcal{M}_{1:K} \right) \right] \\
    &+ \mathbb{E}_{q_\phi(\mathcal{M}_{1:K})} \left[\log p(\mathcal{M}_{1:K}) - \log {q_\phi(\mathcal{M}_{1:K}}) \right]
\end{align*}

Further, note that:
\begin{align*}
    \log p_\theta(X^{(n)}\mid \mathcal{M}_{1:K}) &= \log \left[ \sum_{Z^{(n)}} p_\theta(X^{(n)}\mid Z^{(n)}, \mathcal{M}_{1:K}) p(Z^{(n)}\mid \mathcal{M}_{1:K})\right] \\
    &= \log \left[ \sum_{Z^{(n)}} p_\theta(X^{(n)}\mid Z^{(n)}, \mathcal{M}_{1:K}) p(Z^{(n)}) \times \frac{r_\psi \left(Z^{(n)}\mid X^{(n)} \right)} {r_\psi \left(Z^{(n)}\mid X^{(n)} \right)}\right] \\
    &= \log \mathbb{E}_{r_\psi(Z^{(n)}\mid X^{(n)})}\left[ \frac{p_\theta(X^{(n)}\mid Z^{(n)}, \mathcal{M}_{1:K}) p(Z^{(n)})} {r_\psi \left(Z^{(n)}\mid X^{(n)} \right)}\right] \\
    &\geq \mathbb{E}_{r_\psi \left(Z^{(n)}\mid X^{(n)} \right)}\left[ \log 
 \frac{p_\theta(X^{(n)}\mid Z^{(n)}, \mathcal{M}_{1:K}) p(Z^{(n)})} {r_\psi \left(Z^{(n)}\mid X^{(n)} \right)}\right] \qquad \text{(using Jensen's inequality)}\\  
 &= \mathbb{E}_{r_\psi \left(Z^{(n)}\mid X^{(n)} \right)}\left[ \log 
 p_\theta(X^{(n)}\mid Z^{(n)}, \mathcal{M}_{1:K}) + \log p(Z^{(n)}) - \log {r_\psi \left(Z^{(n)}\mid X^{(n)} \right)}\right].
\end{align*}
We use the fact that $p_\theta(X^{(n)}\mid Z^{(n)}, \mathcal{M}_{1:K}) = p_\theta \left(X^{(n)}\mid \mathcal{M}_{Z^{(n)}} \right)$.
Putting it all together, and using the independence of the causal models, we obtain:
\begin{align*}
    \log p_\theta(X) &\geq \sum_{n=1}^N \mathbb{E}_{q_\phi(\mathcal{M}_{1:K})}\left[ \mathbb{E}_{r_\psi \left(Z^{(n)}\mid X^{(n)} \right)}\left[ \log 
 p_\theta(X^{(n)}\mid \mathcal{M}_{Z^{(n)}}) + \log p(Z^{(n)}) - \log {r_\psi \left(Z^{(n)}\mid X^{(n)} \right)}\right] \right] \\
    &+ \sum_{i=1}^K \mathbb{E}_{q_\phi(\mathcal{M}_i)} \left[\log p(\mathcal{M}_i) - \log {q_\phi(\mathcal{M}_i)} \right] \\
    &\equiv \text{ELBO}(\theta, \phi, \psi)
\end{align*}
\end{proof}

\subsection{Theoretical assumptions}
\label{sec:theory_assumptions}
In this section, we list out the theoretical assumptions used in Rhino \citep{gong2022rhino}. Our model also operates under similar assumptions, since we implement the component SCMs as Rhino models.

\textbf{Assumption 1} (Causal Stationarity). \citep{runge2018causal} The time series $X$ with a graph $G$ is called causally stationary over a time index set $\mathcal{T}$ if and only if for all links $X^i_{t-\tau} \rightarrow X_t^j$ in the graph
\begin{equation*}
    X_{t-\tau}^i \not\!\perp\!\!\!\perp X_t^j \mid X_t \backslash \left\{ X_{t-\tau}^i \right\} \qquad \text{holds for all } t \in \mathcal{T}.
\end{equation*}
Informally, this assumption states that the causal graph does not change over time, i.e., the resulting time series is stationary.

\textbf{Assumption 2} (Causal Markov Property). \citep{peters2017elements} Given a DAG $G$ and a probability distribution $p$, $p$ is said to satisfy the causal Markov property, if it factorizes according to $G$, i.e. $\displaystyle p(x) = \prod_{i=1}^D p \left( x_i \mid \text{Pa}^i_G(x_i) \right)$. In other words, each variable is independent of its non-descendent given its parents.

\textbf{Assumption 3} (Causal Minimality). Given a DAG $G$ and a probability distribution $p$, $p$ is said to satisfy the causal minimality with respect to $G$, if $p$ is Markovian with respect to $G$ but not to any proper subgraph of $G$. 

\textbf{Assumption 4} (Causal Sufficiency). A set of observed variables $V$ is said to be causally sufficient for a process $X_t$ if, in the process, every common cause of two or more variables in $V$ is also in $V$, or is constant for all units in the population. In other words, causal sufficiency implies the absence of latent confounders in the data.

\textbf{Assumption 5} (Well-defined Density). The likelihood of each mixture component (i.e. the likelihood function of each Rhino model) is absolutely continuous
with respect to a Lebesgue or counting measure and $\left| \log p\left( X_{0:T}; G\right)\right| < \infty$ for all possible $G$.

\subsection{Identifiability of the mixture of causal models}
\label{sec:identifiablity}

\begin{customdefn}{1}[Identifiability]
    Let $\displaystyle P = \left\{ p_\theta: \theta \in \mathcal{T} \right\}$ be a family of distributions, each member of which is parameterized by the parameter $\theta$ from a parameter space $\mathcal{T}$. Then $P$ is said to be identifiable if 
    \begin{equation*}
        p_{\theta_1} = p_{\theta_2} \implies \theta_1 = \theta_2 \quad \forall \theta_1, \theta_2 \in \mathcal{T}.
    \end{equation*}
\end{customdefn}

\begin{customdefn}{2}[Identifiability of finite mixtures]
    Let $\displaystyle \mathcal{F}$ be a family of distributions. The family of $\displaystyle K-$mixture distributions on $\displaystyle \mathcal{F}$, defined as $\displaystyle \mathcal{H}_K = \left\{ h: h = \sum_{k=1}^K \pi_k f_k, f_k \in \mathcal{F}, \pi_k > 0, \sum_{k=1}^K \pi_k = 1\right\}$, is said to be identifiable if
    \begin{equation*}
        \sum_{k=1}^K \pi_k f_k = \sum_{j=1}^K \pi_j' f_j' \implies \forall k \ \exists j \text{ such that } \pi_k = \pi_j' \text{ and } f_k = f_j'.
    \end{equation*}
\end{customdefn}

Here, we quote a result from \citet{yakowitz1968identifiability} that established a necessary and sufficient condition for the identifiability of finite mixtures of multivariate distributions.

\begin{customthm}{A}[Identifiability of finite mixtures of distributions \citep{yakowitz1968identifiability}] 
    Let $\displaystyle \mathcal{F} = \left \{ F(x; \alpha), \alpha \in \mathbb{R}^m, x \in \mathbb{R}^n \right\}$ be a finite mixture of distributions. Then $\displaystyle \mathcal{F}$ is identifiable if and only if $\mathcal{F}$ is a linearly independent set over the field of real numbers. \label{thm:identifiability}
\end{customthm}

In other words, this theorem states that a mixture of distributions is identifiable if and only if none of the individual mixture components can be expressed as a mixture of distributions from the same family. In general, it can be difficult to comment on the identifiability of a mixture of arbitrary random distributions. However, it is known that the mixture of multivariate Gaussian distributions is identifiable. We use this result to prove the identifiability of a mixture of linear SCMs with Gaussian noise. 

\begin{customprop}{A}[Identifiability of mixture of multivariate Gaussian distributions \citep{yakowitz1968identifiability}] \label{prop:n_gauss_ident}
    The family of $n$-dimensional Gaussian distributions generates identifiable finite mixtures.
\end{customprop}

\begin{customthm}{B}[Identifiability of linear SCMs with equal-variance additive Gaussian noise]
    Let $\mathcal{F}$ be a family of distributions of $K$ linear causal models with Gaussian noise of equal variance, i.e. 
\begin{equation*}
\mathcal{F} = \left\{ \mathcal{L}_{\mathcal{M}^{(k)}} :  \mathcal{M}^{(k)} \text{ is specified by the equations } \mathbf{X} = \mathbf{W}^{(k)}\mathbf{X} + \varepsilon^{(k)}, \varepsilon^{(k)} \sim \mathcal{N} \left( \mu^{(k)}, \sigma^2 \mathbf{I} \right), 1 \leq k \leq K\right\} 
\end{equation*}    
    
and let $\mathcal{H}_K$ be the family of all $K-$finite mixtures of elements from $\mathcal{F}$, i.e. 
\begin{equation*}
\displaystyle \mathcal{H}_K = \left\{ h: h = \sum_{k=1}^K \pi_k \mathcal{L}_{\mathcal{M}^{(k)}}, \mathcal{L}_{\mathcal{M}^{(k)}} \in \mathcal{F}, \pi_k > 0, \sum_{k=1}^K \pi_k = 1 \right\}    
\end{equation*}
where $\displaystyle \mathcal{L}_{\mathcal{M}^{(k)}}(x) =  p \left( x\mid \mathcal{M}^{(k)} \right)$  denotes the likelihood of $x$ evaluated with causal model $\mathcal{M}^{(k)}$.

Then the family $\mathcal{H}_K$ is identifiable if and only if the following condition is met:
\begin{equation}
    \text{The ordered pairs } \left( \left[ \mathbf{B}^{(k)} \right]^{-1} \mu^{(k)},   \left[\mathbf{B}^{(k)}\right] \left[\mathbf{B}^{(k)}\right]^{T} \right) \text{ are distinct over all $k$, $1 \leq k \leq K$}, \label{eqn:gauss_ident_cond}
\end{equation}
where $\displaystyle \mathbf{B}^{(k)} = \mathbf{I} - \mathbf{W}^{(k)}$.
\label{thm:linear_scm}

\end{customthm}

\begin{proof}
Note that the equations for a linear SCM can equivalently be written as:
\begin{equation*}
    \mathbf{X}^{(k)} = \left[\mathbf{B}^{(k)}\right]^{-1} \varepsilon^{(k)} \sim \mathcal{N} \left(\left[\mathbf{B}^{(k)}\right]^{-1}\mu^{(k)}, \sigma^2 \left[\mathbf{B}^{(k)}\right]^{-1} \left[\mathbf{B}^{(k)}\right]^{-T} \right)
\end{equation*}
where $\mathbf{B}^{(k)} = \left( \mathbf{I} - \mathbf{W}^{(k)}\right)$. 

A linear SCM with equal variance Gaussian additive noise is known to be identifiable \citep{peters2014identifiability}. Thus, from Proposition \ref{prop:n_gauss_ident}, we have that the finite mixture is identifiable, as long as the parameters of the resultant Gaussian distributions are distinct, as required by the condition in \eqref{eqn:gauss_ident_cond}. 

Conversely, if condition \eqref{eqn:gauss_ident_cond} does not hold, then $\exists k \neq j$ such that 
\begin{equation*}
p(x \mid \mathcal{M}^{(k)}) = p(x \mid \mathcal{M}^{(j)}),    
\end{equation*}
i.e. there are two mixture components with identical distributions. This family cannot be identifiable, since any mixture of the form $h = \alpha p \left(x \mid \mathcal{M}^{(k)} \right) + (1-\alpha) p \left(x \mid \mathcal{M}^{(j)} \right) = p(x \mid \mathcal{M}^{(k)})$ for any $\alpha \in \left[0, 1 \right]$
\end{proof}

\begin{customthm}{2}[Identifiability of linear SVARs with equal-variance additive Gaussian noise]
    Let $\mathcal{F}$ be a family of distributions of $K$ structural vector autoregressive (SVAR) models of lag $L \geq 1$ with zero-mean Gaussian noise of equal variance, i.e. 
\begin{align*}
\mathcal{F} = \Biggl\{ \mathcal{L}_{\mathcal{M}^{(k)}} :  \mathcal{M}^{(k)} \text{ is specified by the equations } \mathbf{X}_t = \mathbf{W}^{(k)}\mathbf{X}_t &+ \sum_{\tau=1}^L \mathbf{A}^{(k)}_\tau \mathbf{X}_{t-\tau} + \varepsilon^{(k)}, \\
&\varepsilon^{(k)} \sim \mathcal{N} \left( 0, \sigma^2 \mathbf{I} \right), 1 \leq k \leq K\Biggr\} 
\end{align*}    
    
and let $\mathcal{H}_K$ be the family of all $K-$finite mixtures of elements from $\mathcal{F}$, i.e. 
\begin{equation*}
\displaystyle \mathcal{H}_K = \left\{ h: h = \sum_{k=1}^K \pi_k \mathcal{L}_{\mathcal{M}^{(k)}}, \mathcal{L}_{\mathcal{M}^{(k)}} \in \mathcal{F}, \pi_k > 0, \sum_{k=1}^K \pi_k = 1 \right\}    
\end{equation*}
where $\displaystyle \mathcal{L}_{\mathcal{M}^{(k)}}(x) =  p \left( x\mid \mathcal{M}^{(k)} \right)$  denotes the likelihood of $x$ evaluated with causal model $\mathcal{M}^{(k)}$.

Then the family $\mathcal{H}_K$ is identifiable if and only if the following condition is met:
\begin{equation}
    \text{The ordered pairs } \left( \left[\mathbf{B}^{(k)}\right]^{-1}\mathbf{A}_1^{(k)}, ..., \left[\mathbf{B}^{(k)}\right]^{-1}\mathbf{A}_L^{(k)},   \left[\mathbf{B}^{(k)}\right] \left[\mathbf{B}^{(k)}\right]^{T} \right) \text{ are distinct over all $k$}, \label{eqn:svar_gauss_ident_cond}
\end{equation}
where $\displaystyle \mathbf{B}^{(k)} = \mathbf{I} - \mathbf{W}^{(k)}$.

\end{customthm}

\begin{proof}
Note that the SVAR equations can equivalently be written as:
\begin{equation*}
    \mathbf{X}_t = \left[\mathbf{B}^{(k)}\right]^{-1} \sum_{\tau=1}^L \mathbf{A}^{(k)}_\tau \mathbf{X}_{t-\tau} + \left[\mathbf{B}^{(k)}\right]^{-1}\varepsilon^{(k)}
\end{equation*}
where $\mathbf{B}^{(k)} = \left( \mathbf{I} - \mathbf{W}^{(k)}\right)$. 

This implies that 
\begin{equation*}
    p \left( \mathbf{X}_t \mid \mathbf{X}_{t-1}, ..., \mathbf{X}_{t-L}, \mathcal{M}^{(k)} \right) \sim \mathcal{N} \left(\left[\mathbf{B}^{(k)}\right]^{-1} \sum_{\tau=1}^L \mathbf{A}^{(k)}_\tau \mathbf{X}_{t-\tau}, \sigma^2\left[\mathbf{B}^{(k)}\right]^{-1}\left[\mathbf{B}^{(k)}\right]^{-T} \right).
\end{equation*}

Following a similar argument as in the proof of Theorem \ref{thm:linear_scm}, the finite mixture is identifiable if and only if the parameters of the resultant Gaussian distributions are distinct as a function of $\left\{\mathbf{X}_{t-\tau}\right\}_{\tau=1}^L$. Hence, the condition.
\end{proof}

It can be difficult to reason about the identifiability of a mixture of SCMs whose structural equations come from a general class of functions, or whose noise distribution is non-Gaussian.  However, the likelihood can be evaluated quite easily on a finite number of points, at least approximately if not exactly. 

Here, we describe a sufficient condition for the identifiability of finite mixtures of \textit{identifiable} causal models. 

\begin{customthm}{3}[Identifiability of finite mixture of causal models]
    Let $\mathcal{F}$ be a family of $K$ identifiable causal models, i.e. $\displaystyle \mathcal{F} = \left\{ \mathcal{L}_\mathcal{M}^{(k)} 
: \mathcal{M} \text{ is an identifiable causal model }, 1 \leq k \leq K\right\}$ and let $\mathcal{H}_K$ be the family of all $K-$finite mixtures of elements from $\mathcal{F}$, i.e. 
\begin{equation*}
\displaystyle \mathcal{H}_K = \left\{ h: h = \sum_{k=1}^K \pi_k \mathcal{L}_{\mathcal{M}_k}, \mathcal{L}_{\mathcal{M}_k} \in \mathcal{F}, \pi_k > 0, \sum_{k=1}^K \pi_k = 1 \right\}    
\end{equation*}
where $\displaystyle \mathcal{L}_{\mathcal{M}_k}(x) = \sum_{\mathcal{M}} p(x\mid \mathcal{M}) p(\mathcal{M}_k = \mathcal{M})$  denotes the likelihood of $x$ evaluated with causal model $\mathcal{M}_k$. Further, assume that the following condition is met:

\begin{align}
\text{For every } k, 1 \leq k \leq K, \exists a_k \in \mathbb{X} \text{ such that } \frac{\mathcal{L}_{\mathcal{M}_k}(a_k)}{\sum_{j=1}^K \mathcal{L}_{\mathcal{M}_j}(a_k)} > \frac{1}{2}. \tag{*}
\end{align}

Then the family $\mathcal{H}_K$ is identifiable, i.e., if $\displaystyle h_1=\sum_{k=1}^K \pi_k \mathcal{L}_{\mathcal{M}_k} \text{ and } h_2 = \sum_{j=1}^K \pi_j' \mathcal{L}_{\mathcal{M}'_j} \in \mathcal{H}_K$ then:
\begin{equation*}
     {h_1} =  {h_2} \implies \forall k \in \left\{1, \ldots , K \right\} \; \exists j \in \left\{1, \ldots , K \right\} \text{ such that } \pi_k = \pi_j' \text{ and } \mathcal{M}_k = \mathcal{M}'_j.
\end{equation*}
\end{customthm}

\begin{proof}
    From Theorem \ref{thm:identifiability}, we have that $\mathcal{H}_K$ is identifiable if and only if for any $\alpha_1, \ldots , \alpha_K \in \mathbb{R}$,
    \begin{equation*}
        \sum_{j=1}^K \alpha_j \mathcal{L}_{\mathcal{M}_j} = 0 \implies \alpha_j =0 \; \; \forall j \in \left\{1, \ldots , K \right\}
    \end{equation*}
    Note that $\displaystyle \sum_{j=1}^K \alpha_j \mathcal{L}_{\mathcal{M}_j}=0 \implies \sum_{j=1}^K \alpha_j \mathcal{L}_{\mathcal{M}_j}(x)=0 \quad \forall x \in \mathbb{X}$.
    In particular, 
    \begin{equation}
        \sum_{j=1}^K \alpha_j \mathcal{L}_{\mathcal{M}_j}(a_k) = 0 \quad \forall k \in \left\{1, \ldots , K \right\}, \label{eqn:li_condition}
    \end{equation}
    where $a_k$ is as defined in Condition $\left(\eqref{eqn:diversity_condition}\right)$. Denote $\mathcal{L}_{\mathcal{M}_j}(a_k) = \beta_{kj}$. Then Equation \eqref{eqn:li_condition} can be written as:

    \begin{equation}
        \begin{bmatrix}
            \beta_{11} & \ldots  & \beta_{1K} \\
            \vdots  &       & \vdots \\
            \beta_{K1} & \ldots  & \beta_{KK}
        \end{bmatrix} \begin{bmatrix}
            \alpha_1 \\
            \vdots \\
            \alpha_K
        \end{bmatrix} = \bm{0}.
    \end{equation}
    Or equivalently
    \begin{equation}
        \bm{\beta}\bm{\alpha} = \bm{0}.
    \end{equation}

    Note that $\bm{\alpha} = 0$ if and only if $\bm{\beta}$ is full rank. We now show that Condition $\left(\eqref{eqn:diversity_condition}\right)$ implies that $\bm{\beta}$ is strictly diagonally dominant and hence full rank. Note that Condition $\left(\eqref{eqn:diversity_condition}\right)$ can be equivalently written as:
    \begin{align*}
        \frac{\beta_{kk}}{\sum_{j=1}^K \beta_{kj}} > \frac{1}{2} &\implies 2 \beta_{kk} > \sum_{j=1}^K \beta_{kj} \\
        &\implies \beta_{kk} > \sum_{j=1, j \neq k}^K \beta_{kj}
    \end{align*}
    which implies strict diagonal dominance since $\beta_{kj} \geq 0 \quad \forall k, j$. Hence $\bm{\alpha} = 0$ thus implying linear independence.
 \end{proof}

Note that $a_k$ refers to any point in the support of the mixture distribution such that the condition \eqref{eqn:diversity_condition} is satisfied. It does not constitute a 'sample' from the $k^\text{th}$ SCM in the conventional sense of being randomly drawn from the SCM. Instead, it can be intentionally chosen to meet the specified condition.

 \subsection{Relationship between ELBO and log-likelihood} \label{sec:elbo_loglikelihood}
 In this section, we derive an exact relationship between the derived evidence lower bound $\text{ELBO}(\theta, \phi, \psi)$ and the log-likelihood $\log p_\theta(X)$.

 First, note that:
 \begin{equation*}
     p_\theta(X) p\left(\mathcal{M}_{1:K} \mid X\right)=p_\theta\left(X \mid \mathcal{M}_{1:K}\right) p\left(\mathcal{M}_{1:K}\right)
\end{equation*}
and hence:
\begin{equation*}
p_\theta(X)=\frac{p_\theta\left(X \mid \mathcal{M}_{1:K}\right) p\left(\mathcal{M}_{1:K}\right)}{p\left(\mathcal{M}_{1:K} \mid X \right)}.
 \end{equation*}
 The log-likelihood can be written as:
 
 \begin{align*}
\log p_\theta(X)&=\mathbb{E}_{q_\phi (\mathcal{M}_{1:K} )}\left[\log p_\theta(X)\right] \\
&= \mathbb{E}_{q_\phi (\mathcal{M}_{1:K} )}\left[\log \frac{p_\theta\left(X \mid \mathcal{M}_{1:K}\right) p\left(\mathcal{M}_{1:K}\right)}{p\left(\mathcal{M}_{1:K} \mid X \right)} \times \frac{q_\phi(\mathcal{M}_{1:K})}{q_\phi(\mathcal{M}_{1:K})} \right] \\
& =\mathbb{E}_{q_\phi (\mathcal{M}_{1:K} )}\left[\log p_\theta \left(X \mid \mathcal{M}_{1:K}\right)+\log p\left(\mathcal{M}_{1:K}\right)\right]\\
&+\sum_{i=1}^K\text{H}\left(q_\phi(\mathcal{M}_i)\right)
 + \text{KL}\left(q_\phi\left(\mathcal{M}_{1:K}\right) \mid \mid   p\left(\mathcal{M}_{1:K} \mid X\right)\right) \\
 &=\mathbb{E}_{q_\phi (\mathcal{M}_{1:K} )}\left[\sum_{n=1}^N \log p_\theta \left(X^{(n)} \mid \mathcal{M}_{1:K}\right)+\sum_{i=1}^K \log p\left(\mathcal{M}_i\right)\right]\\
&+\sum_{i=1}^K\text{H}\left(q_\phi(\mathcal{M}_i)\right)
 + \text{KL}\left(q_\phi\left(\mathcal{M}_{1:K}\right) \mid \mid   p\left(\mathcal{M}_{1:K} \mid X\right)\right)
\end{align*}
Also note that, using the rules of conditional probability:
\begin{align*}
    \frac{p_\theta(X^{(n)} \mid 
    \mathcal{M}_{1:K} )}{p_\theta(X^{(n)} \mid 
    \mathcal{M}_{1:K}, Z^{(n)})} &= \frac{p_\theta(X^{(n)}, \mathcal{M}_{1:K})}{p(\mathcal{M}_{1:K})} \times \frac{p(Z^{(n)}, \mathcal{M}_{1:K})}{p_\theta(X^{(n)}, Z^{(n)}, \mathcal{M}_{1:K})} \\
    &= \frac{p(Z^{(n)} \mid \mathcal{M}_{1:K})}{p(Z^{(n)}\mid X^{(n)}, \mathcal{M}_{1:K})} \\
    &= \frac{p(Z^{(n)})}{p(Z^{(n)}\mid X^{(n)}, \mathcal{M}_{1:K})}
\end{align*}
where the last step follows from the fact that $Z^{(n)}$ and $\mathcal{M}_i$ are independent.

% Further note that:
% \begin{align*}
%     p(Z^{(n)}\mid X^{(n)}, \mathcal{M}_{1:K}) &= \frac{p(X^{(n)}, \mathcal{M}_{1:K}, Z^{(n)})}{p(X^{(n)}, \mathcal{M}_{1:K})} = \frac{p(X^{(n)} \mid \mathcal{M}_{1:K}, Z^{(n)}) p(Z^{(n)}) p(\mathcal{M}_{1:K})}{\sum_{Z^{(n)}} p(X^{(n)} \mid \mathcal{M}_{1:K}, Z^{(n)}) p(Z^{(n)}) p(\mathcal{M}_{1:K})} \\  
%     &= \frac{p(X^{(n)} \mid \mathcal{M}_{Z^{(n)}})p(Z^{(n)})}{\sum_{Z^{(n)}} p(X^{(n)} \mid \mathcal{M}_{Z^{(n)}})p(Z^{(n)})} \\
%     &= \frac{p(X^{(n)}, Z^{(n)})}{p(X^{(n)})} = p(Z^{(n)} \mid X^{(n)})
% \end{align*}
% where we used the fact that
% \begin{align*}
%     p(X^{(n)}, Z^{(n)}) &= \sum_{\mathcal{M}_{1:K}} p(X^{(n)}, Z^{(n)}, \mathcal{M}_{1:K}) \\
%     &= \sum_{\mathcal{M}_{1:K}} p(X^{(n)}, Z^{(n)}, \mathcal{M}_{1:K}) \\
%     &= \sum_{\mathcal{M}_{1:K}} p(X^{(n)} \mid Z^{(n)}, \mathcal{M}_{1:K}) p(Z^{(n)}) p(\mathcal{M}_{1:K})\\
%     &= \sum_{\mathcal{M}_{1:K}} p(X^{(n)} \mid \mathcal{M}_{Z^{(n)}}) p(Z^{(n)}) p(\mathcal{M}_{1:K}) = p(X^{(n)} \mid \mathcal{M}_{Z^{(n)}}) p(Z^{(n)}).
% \end{align*}

Thus, we can write:
\begin{align*}
    p_\theta(X^{(n)} \mid \mathcal{M}_{1:K}) &= \mathbb{E}_{r_\psi \left(Z^{(n)} \mid X^{(n)} \right)}\left[ 
    p_\theta(X^{(n)} \mid \mathcal{M}_{1:K})\right] \\
    &= \mathbb{E}_{r_\psi \left(Z^{(n)} \mid X^{(n)} \right)} \left[\frac{p_\theta(X^{(n)} \mid \mathcal{M}_{1:K}, Z^{(n)}) p (Z^{(n)})}{p(Z^{(n)}\mid X^{(n)}, \mathcal{M}_{1:K})} \right] \\
    &= \mathbb{E}_{r_\psi \left(Z^{(n)} \mid X^{(n)} \right)} \left[ \frac{p_\theta(X^{(n)} \mid \mathcal{M}_{Z^{(n)}}) p (Z^{(n)})}{p(Z^{(n)}\mid X^{(n)}, \mathcal{M}_{1:K})} \times \frac{r_\psi \left(Z^{(n)} \mid X^{(n)} \right)}{r_\psi \left(Z^{(n)} \mid X^{(n)} \right)} \right].
\end{align*}

Thus,
\begin{align*}
    \log p_\theta(X)&= \mathbb{E}_{q_\phi (\mathcal{M}_{1:K} )}\Bigg[ \sum_{n=1}^N\mathbb{E}_{ r_\psi \left(Z^{(n)} \mid X^{(n)} \right)} \left[  \log p_\theta \left(X^{(n)} \mid \mathcal{M}_{Z^{(n)}} \right)+ \log p(Z^{(n)}) \right] + \text{H}\left(r_\psi \left(Z^{(n)} \mid X^{(n)} \right)\right) \\
    &+ \text{KL} \left(r_\psi(Z^{(n)}\mid X^{(n)}) \mid \mid  p(Z^{(n)}\mid X^{(n)}, \mathcal{M}_{1:K}) \right)    
     +\sum_{i=1}^K \log p\left(\mathcal{M}_i\right)\Bigg] +\sum_{i=1}^K\text{H}\left(q_\phi(\mathcal{M}_i)\right) \\
    &+ \text{KL}\left(q_\phi\left(\mathcal{M}_{1:K}\right) \mid \mid   p\left(\mathcal{M}_{1:K} \mid X\right)\right).
\end{align*}
Noting that
\begin{align*}
    \text{ELBO}(\theta, \phi, \psi) &\equiv \sum_{n=1}^N \mathbb{E}_{q_\phi(\mathcal{M}_{1:K})}\Bigg[ \mathbb{E}_{r_\psi \left( Z^{(n)}\mid X^{(n)} \right)}\Big[ \log 
 p_\theta(X^{(n)}\mid \mathcal{M}_{Z^{(n)}}) + \log p(Z^{(n)}) - \log {r_\psi \left(Z^{(n)}\mid X^{(n)} \right)}\Big] \Bigg] \\
    &+ \sum_{i=1}^K \mathbb{E}_{q_\phi(\mathcal{M}_i)} \left[\log p(\mathcal{M}_i) - \log {q_\phi(\mathcal{M}_i)} \right]
\end{align*}
 we obtain that:
 \begin{align*}
     \log p_\theta(X) &= \text{ELBO}(\theta, \phi, \psi) + \sum_{n=1}^N \mathbb{E}_{q_\phi(\mathcal{M}_{1:K})}\left[\text{KL} \left( r_\psi \left(Z^{(n)} \mid X^{(n)} \right) \mid \mid p(Z^{(n)} \mid X^{(n)}, \mathcal{M}_{1:K})\right)\right] \\
     &+ \text{KL}\left(q_\phi\left(\mathcal{M}_{1:K}\right) \mid \mid   p\left(\mathcal{M}_{1:K} \mid X\right)\right).
 \end{align*}

\section{Additional experiments}

\subsection{More results on the synthetic datasets}
\begin{figure}
    \centering
    \includegraphics[width=\textwidth]{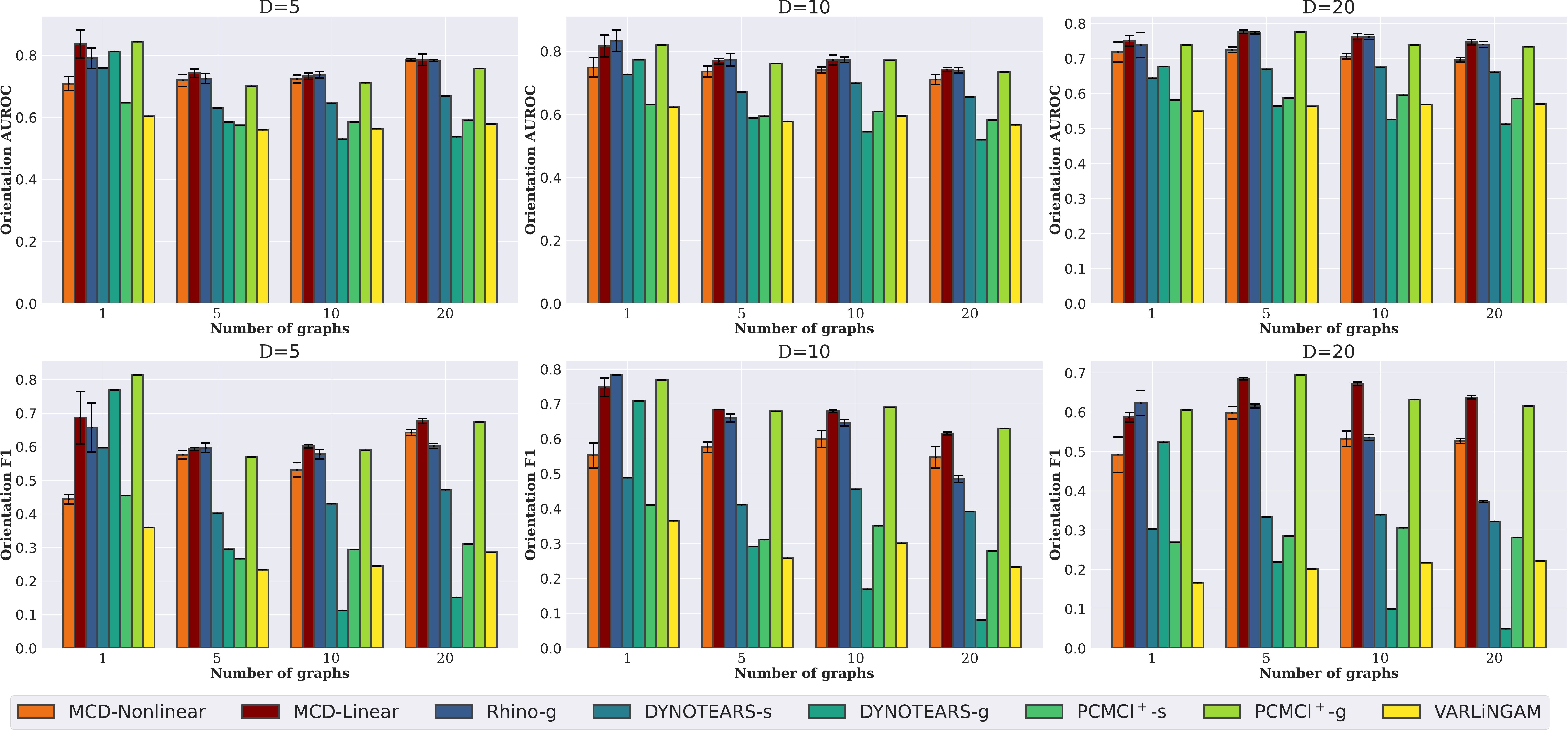}
    \caption{Results on the linear synthetic dataset for $D=5, 10, 20$. We report both the orientation F1 and AUROC scores. Average of 5 runs reported.}
    \label{fig:all_syn_linear}
\end{figure}

\begin{figure}
    \centering
    \includegraphics[width=\textwidth]{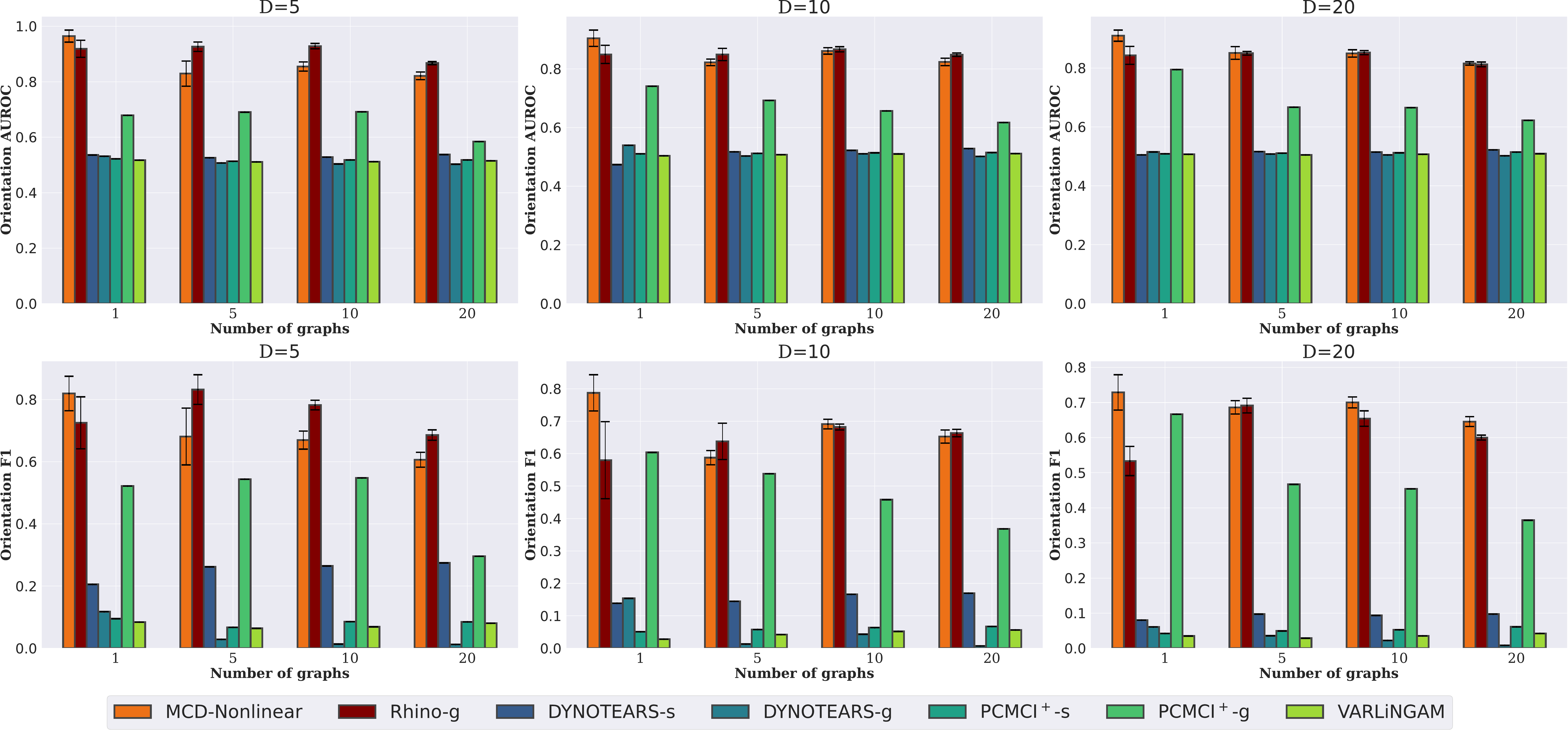}
    \caption{Results on the nonlinear synthetic dataset for $D=5, 10, 20$. We report both the orientation F1 and AUROC scores. Average of 5 runs reported.}
    \label{fig:all_syn_nonlinear}
\end{figure}

Figure \ref{fig:all_syn_linear} shows the results of all methods on the linear synthetic datasets, and Figure \ref{fig:all_syn_nonlinear} shows the results on the nonlinear synthetic datasets. We observe that the difference in performance between \ours{} and Rhino-g is much lower in terms of AUROC compared to orientation F1. \ours{} is able to achieve similar performance to the `gold-standard' baseline Rhino-g despite not having ground-truth membership information.


\subsection{S\&P100}
\label{sec:snp100_more_results}

\textbf{Setup.}  We provide more details on the setup for the experiment with the S\&P 100 dataset. We used grid search to iterate over multiple values for the sparsity term $\lambda$ in equation \eqref{eqn:prior_term}, number of graphs $K$, and 5 random seeds. We picked the setting that yielded the lowest validation loss and reported results using the inferred causal graphs. We aggregate the temporal adjacency matrix following the procedure described in Appendix \ref{sec:netsim_dream3}. 

\begin{figure}
    \centering
    \includegraphics[width=0.48\textwidth]{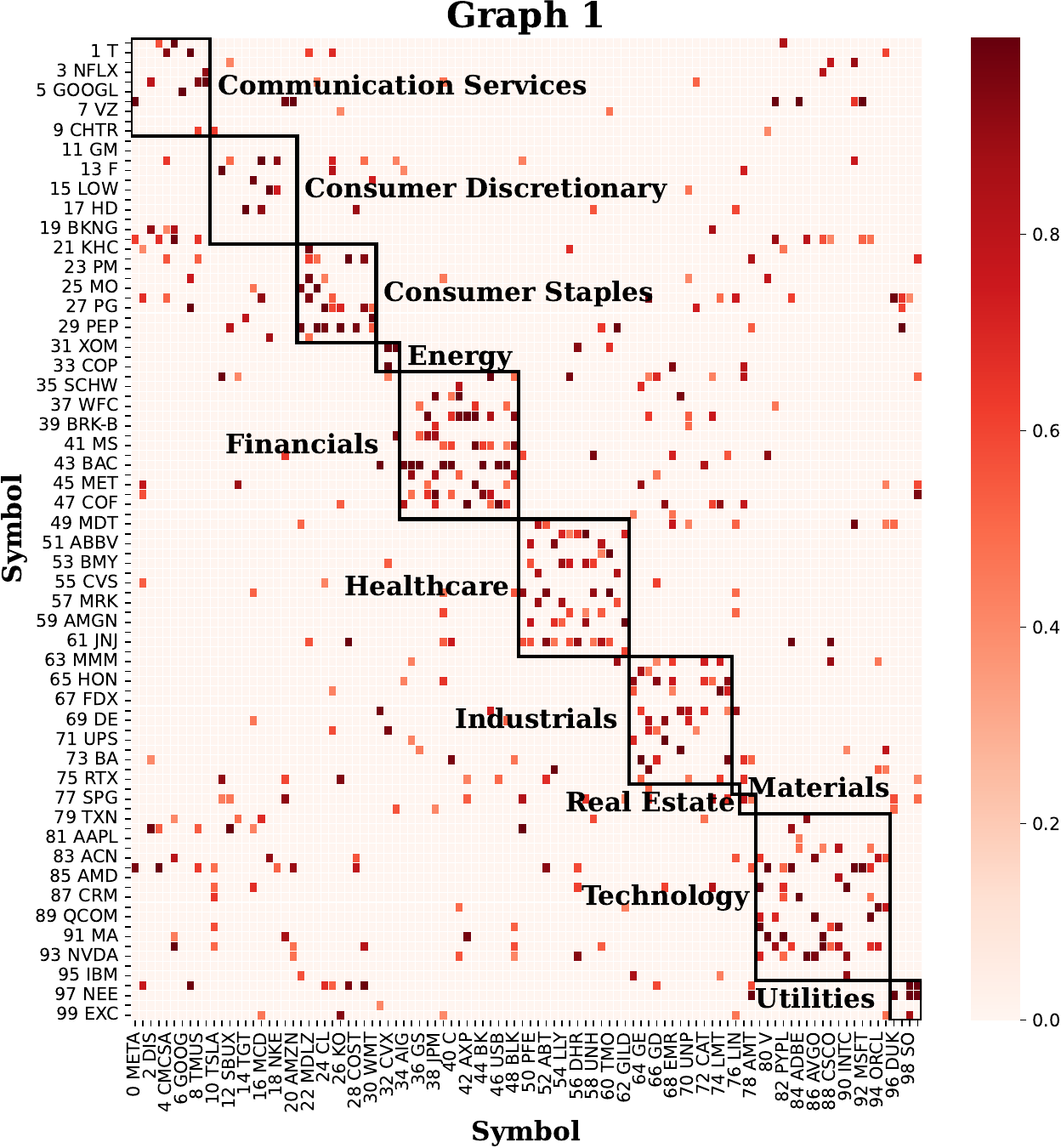}
    \includegraphics[width=0.48\textwidth]{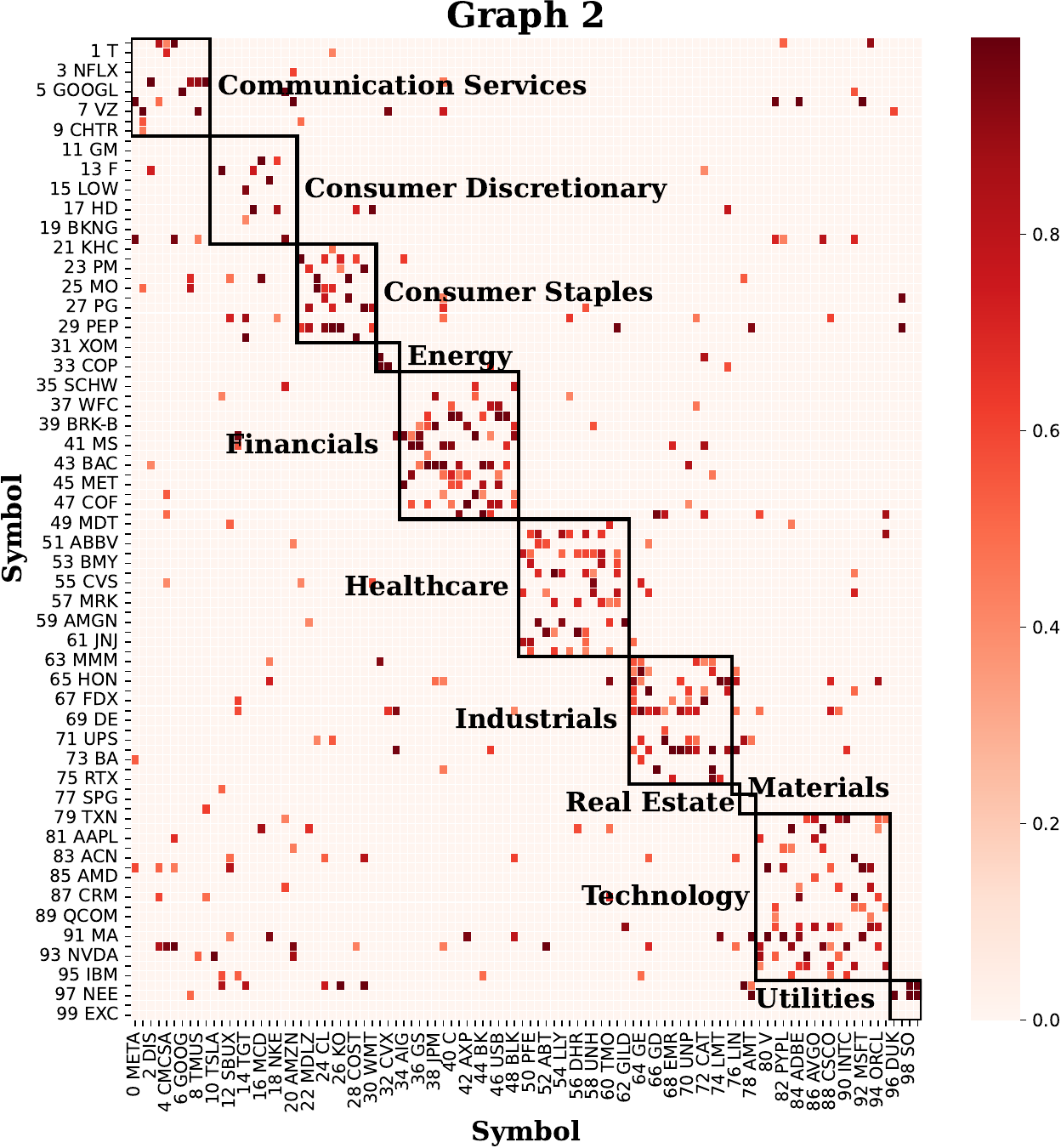}
    \caption{Discovered causal graphs from S\&P 100. \oursnonlinear{} discovers two distinct graphs from the dataset.  }
    \label{fig:snp100_graphs}
\end{figure}

\textbf{Additional results.} Figure \ref{fig:snp100_graphs} shows the heatmap of the two aggregated causal graphs inferred by \oursnonlinear{}. As noted in the main paper, we observe several interesting differences between the two graphs. Many sectors such as `Industrials', `Utilities' and `Technology' seem to have different patterns of intra-sector interactions in the two graphs. Further, Graph 1 shows more marked interactions for stocks in the `Real Estate' sector.

\begin{figure}
    \centering
    \includegraphics[width=0.45\textwidth]{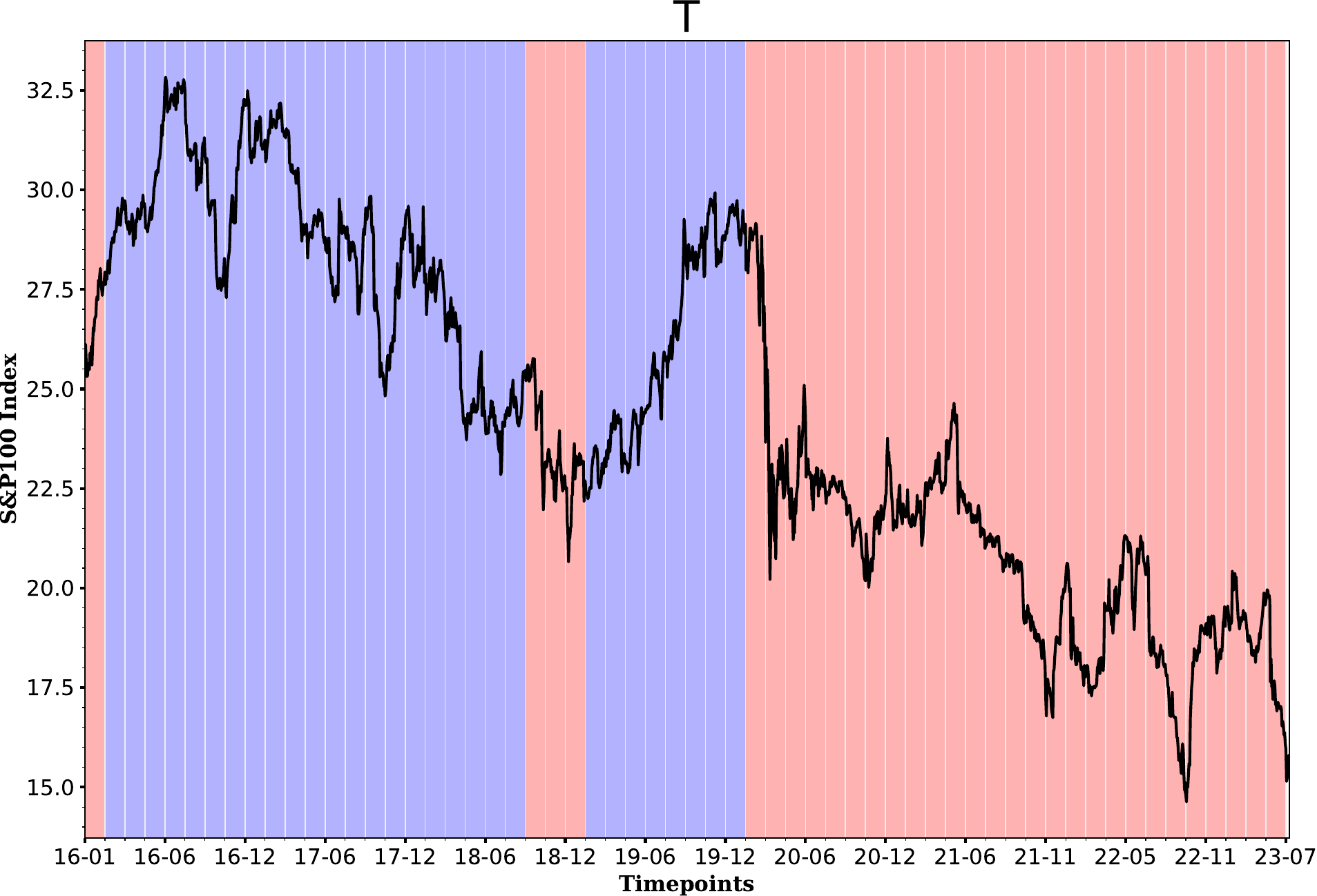}
    \includegraphics[width=0.45\textwidth]{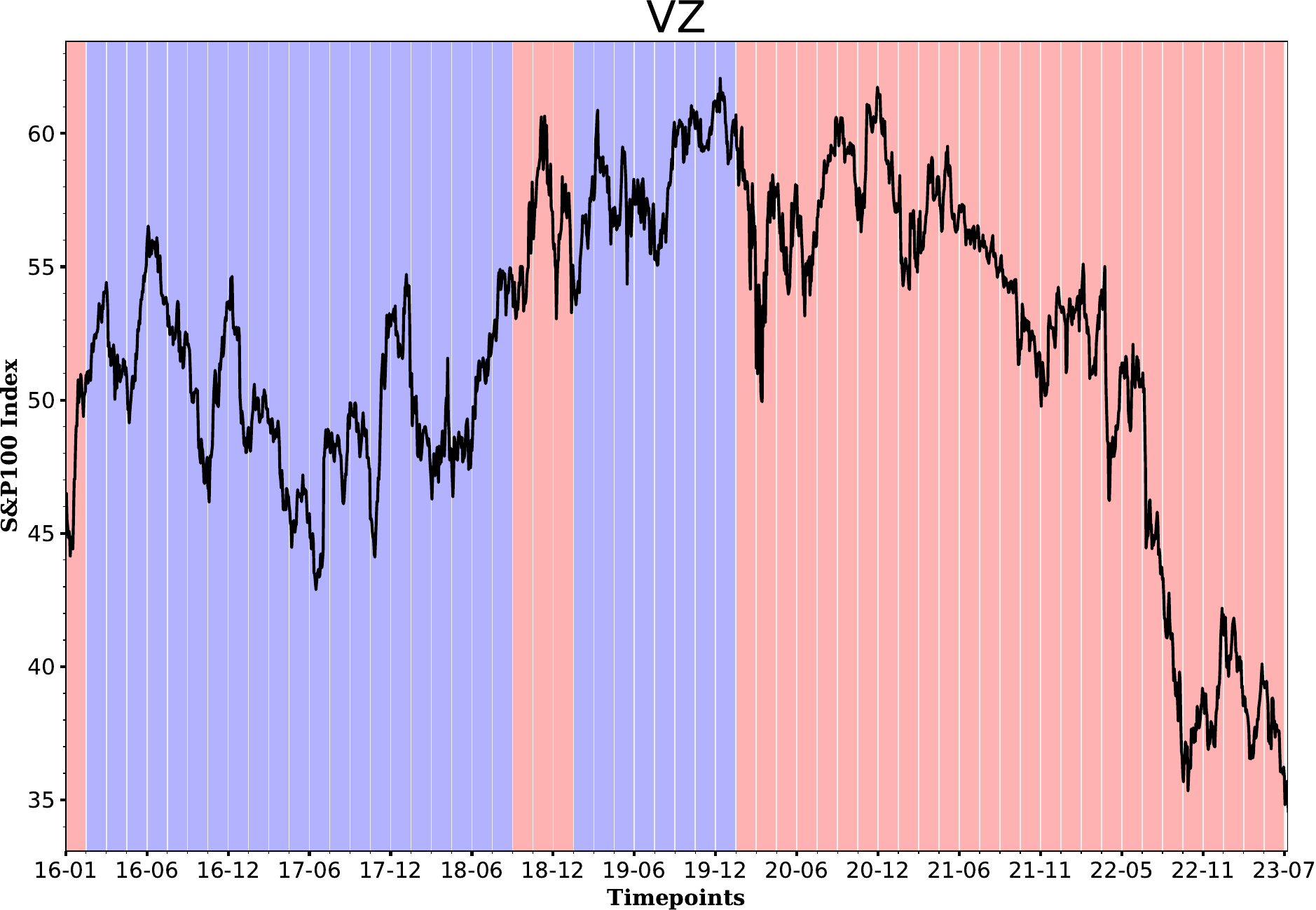}
    \includegraphics[width=0.45\textwidth]{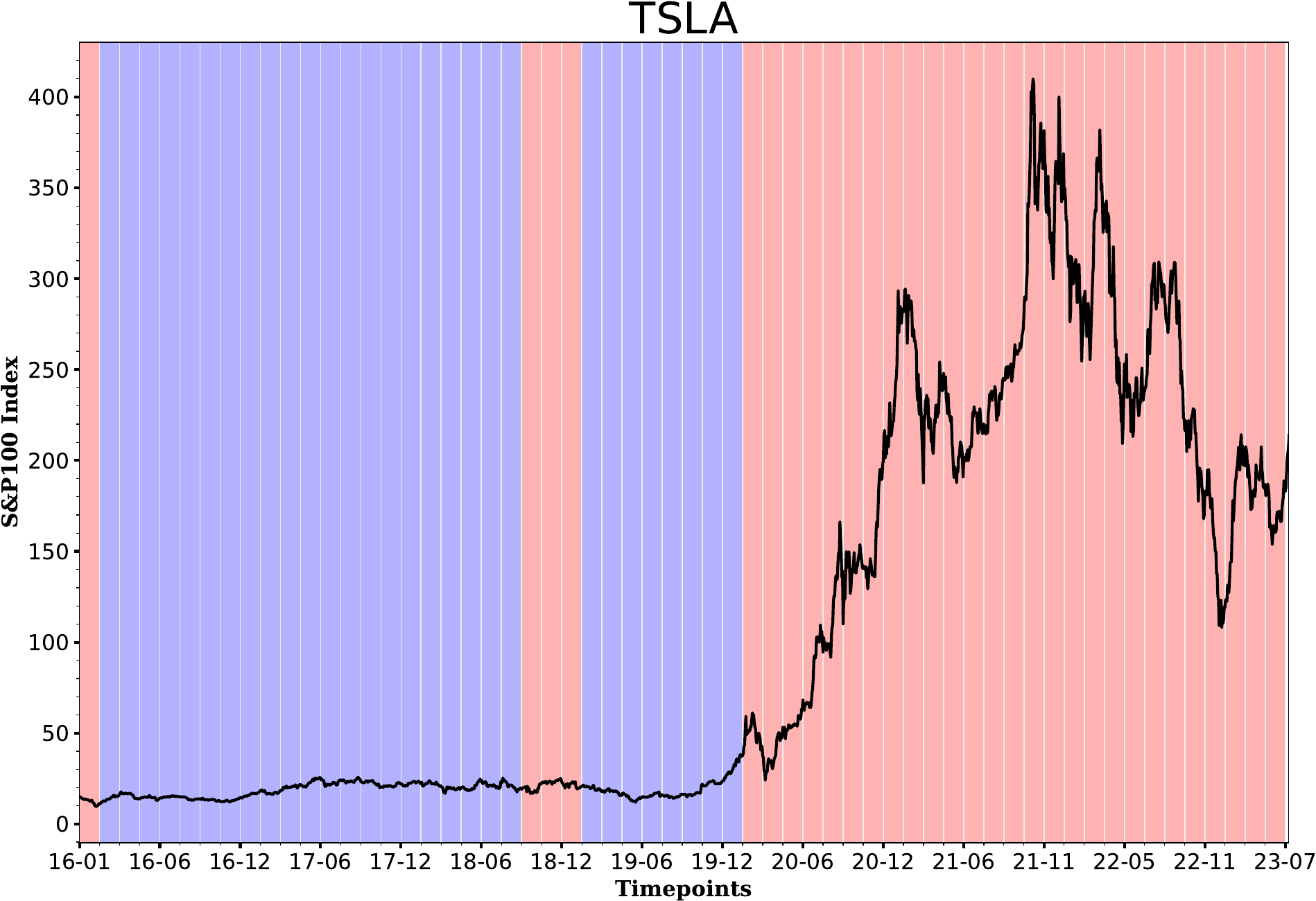}
    \includegraphics[width=0.45\textwidth]{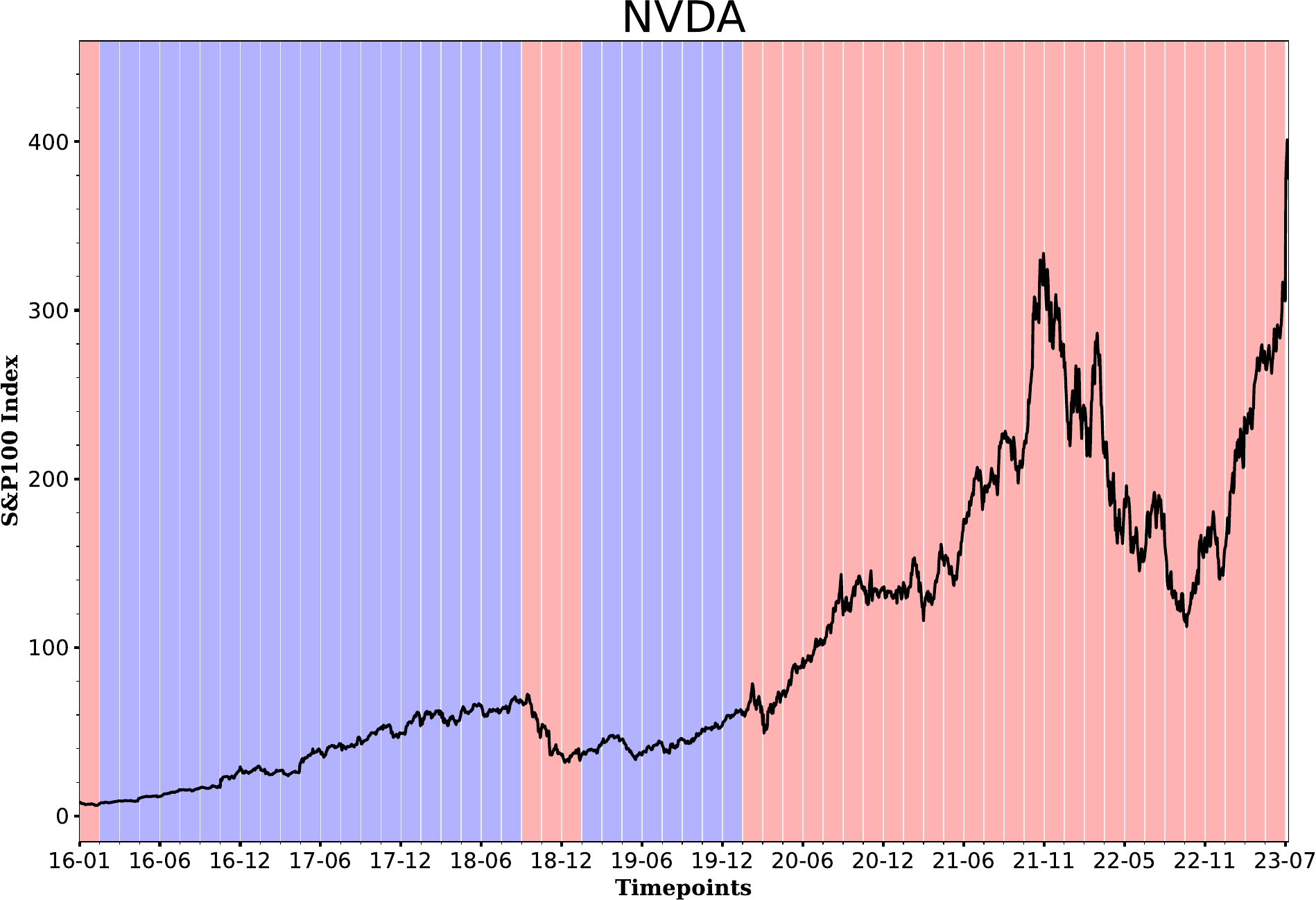}
    \includegraphics[width=0.45\textwidth]{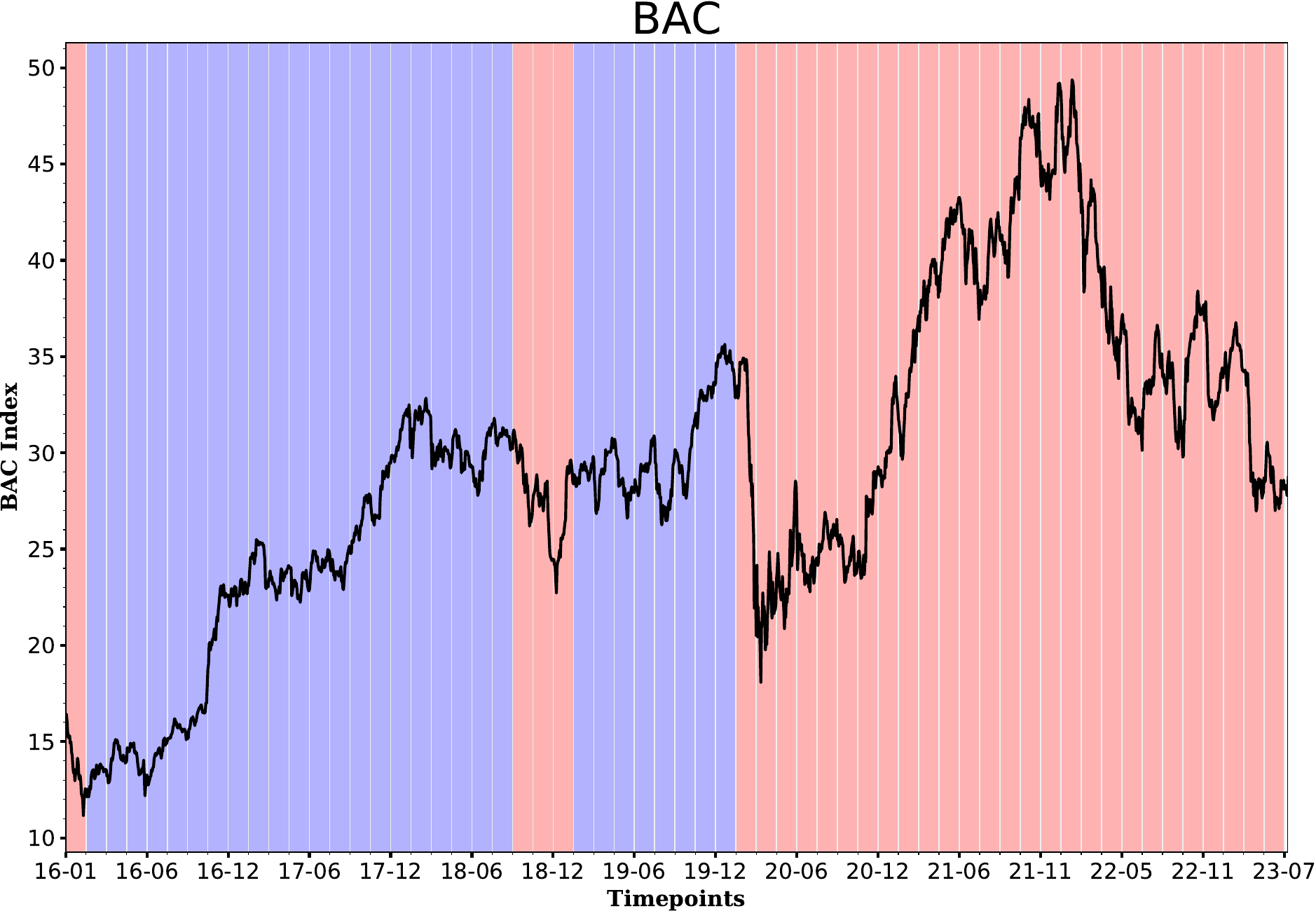}
    \includegraphics[width=0.45\textwidth]{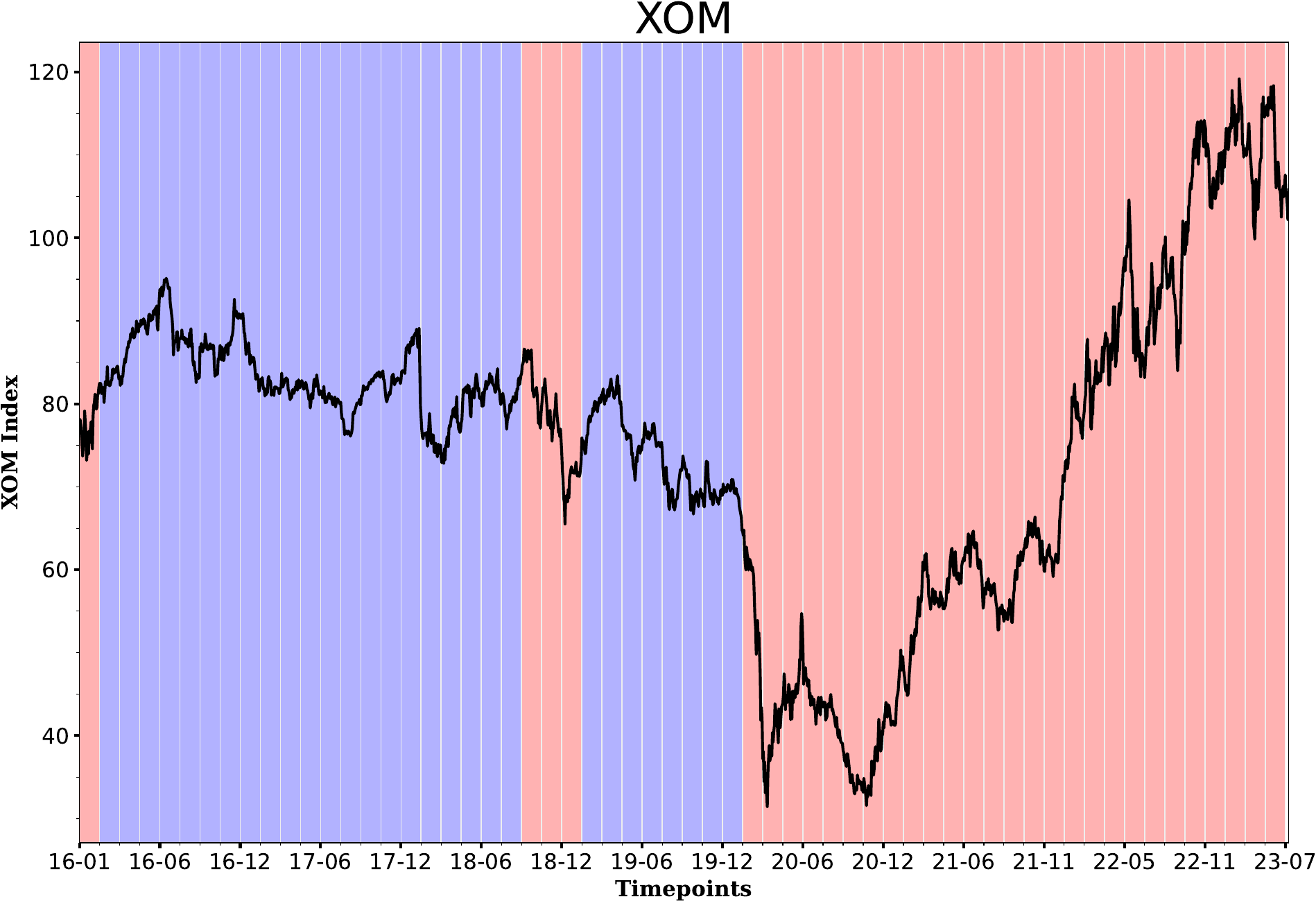}
    
    \caption{Segmentation of the stock prices of  AT\&T,  Verizon, Tesla, NVIDIA, Bank of America, and Exxon Mobil stock prices with respect to the inferred causal graphs from \oursnonlinear{}. }
    \label{fig:snp100_more_results}
\end{figure}

We also visualize the stock prices of the companies whose interactions changed between the two graphs and overlaid the membership information over their indices, as we did with Figure \ref{fig:snp100-results}. Figure \ref{fig:snp100_more_results} provides examples of 6 such stocks. We observe that in most cases, the `red' periods, i.e., periods in which Graph 1 is active, show more pronounced trends and marked movements of the stock prices compared to the `blue' periods.  

\begin{figure}
    \centering
    \includegraphics[width=0.5\textwidth]{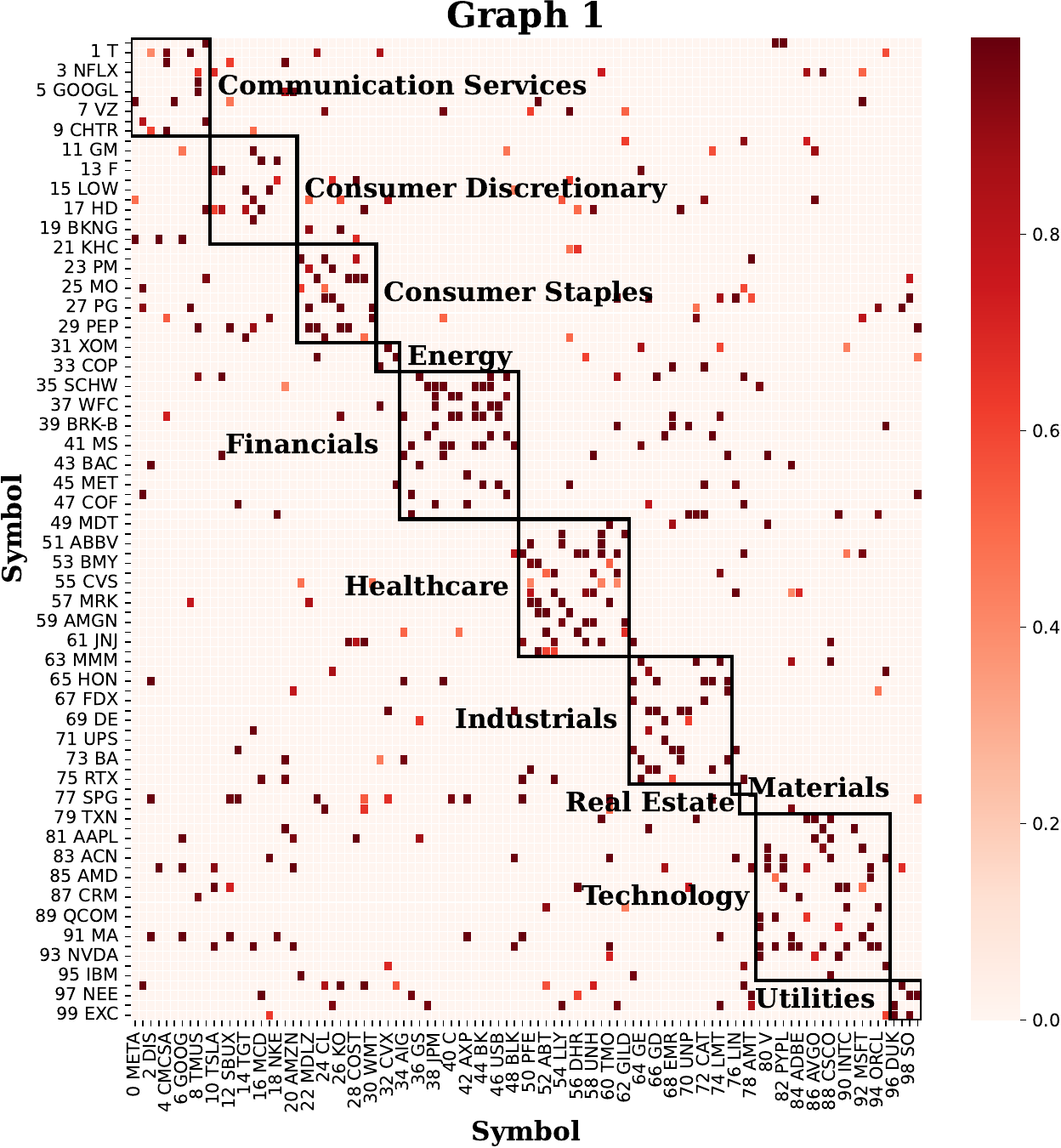}
    \caption{Discovered causal graph from S\&P100 using \ourslinear{}. \ourslinear{} only discovers a single mode in the dataset}
    \label{fig:snp100_linear}
\end{figure}

Additionally, we run \ourslinear{} on this dataset and observe that it only discovers a single mode in the dataset. Figure \ref{fig:snp100_linear} shows the discovered causal graph from \ourslinear{}.

\subsection{Netsim}

\begin{table}
\centering

\begin{tabular}{|c|cc|}
\hline
                                & \multicolumn{2}{c|}{\textbf{Netsim}}                   \\ \hline
\textbf{Method}                 & \multicolumn{1}{c}{\textbf{AUROC}$\left(\big \uparrow \right)$} & \textbf{F1} $\left(\big \uparrow \right)$ \\ \hline
PCMCI$^+$-s          & \multicolumn{1}{c}{$0.702$}          & $0.648$                 \\ 
PCMCI$^+$-o             & \multicolumn{1}{c}{$0.827$}          & $\mathbf{0.803}$                 \\ 
PCMCI$^+$-g             & \multicolumn{1}{c}{$0.810$}          & $0.785$                 \\ 
VARLiNGAM                       & \multicolumn{1}{c}{$0.638$}          & $0.598$                 \\ 
DYNOTEARS-s          & \multicolumn{1}{c}{$0.706$}          & $0.588$                 \\ 
DYNOTEARS-o             & \multicolumn{1}{c}{$0.674$}          & $0.626$                 \\ 
DYNOTEARS-g             & \multicolumn{1}{c}{$0.629$}          & $0.584$                 \\ 
Rhino                           & \multicolumn{1}{c}{$\mathbf{0.873 \pm 0.026}$}          & $0.707 \pm 0.033$                 \\
\oursnonlinear{} (this paper)                          & \multicolumn{1}{c}{${ 0.733 \pm 0.060}$}          & $ 0.607\pm 0.052$                 \\
\ourslinear{} (this paper) & \multicolumn{1}{c}{$0.728 \pm 0.018$}          & $ 0.623 \pm 0.022$                 \\ 
\hline
\end{tabular}

\caption{Results on the Netsim dataset. -s indicates that the baseline predicts one graph per sample. -o indicates that the baseline predicts one graph for the whole dataset. -g signifies that the baseline is run on samples grouped according to the ground truth causal graph. VARLiNGAM does not run on the DREAM3 dataset. \oursnonlinear{} achieves a clustering accuracy of $ 35.2 \pm 6.6\%$ on Netsim, while \ourslinear{} has a clustering accuracy of $35.4 \pm 5.2 \%$ .}
\label{tab:netsim}
\end{table}
\textbf{Setup.} We additionally experiment with a different setup on the Netsim Brain connectivity dataset.
We combine the time series with length $T=200$ and number of nodes $D=5$ from simulations 1, 8, 10, 13, 14, 15, 16, 18, 21, 22, 23, and 24. This dataset comprises $N=600$ samples, with $K^\ast=14$ distinct underlying causal graphs. We refer to this setup as \textbf{Netsim}. This dataset exhibits significant graph membership imbalance, with the top 3 causal graphs accounting for 500 out of the 600 samples. Hence, we consider an exponentially weighted prior for the membership indicators, i.e. $p(Z=k) \propto \exp\left( -\lambda_p k \right)\, \forall k \in \left\{1, \ldots, K \right\}$. We set $\lambda_p = 5$ and $K=20$.

\textbf{Results.} In this setup, we observe that \oursnonlinear{} and \ourslinear{} are outperformed by the baselines PCMCI$^+$ and Rhino, even though they only predict one graph for the entire dataset. This is attributed to the similarity among the various underlying graphs in the Netsim dataset and the strong imbalance in the data. Our model faces sample complexity issues because it learns multiple causal graphs, whereas other methods perform reasonably well by predicting only one. This highlights the idea that learning a mixture model is only beneficial when the underlying SCMs differ from one another significantly. In such a scenario, the benefits of learning multiple graphs outweigh the drawbacks of limited samples per model. This explanation is also supported by the observation that PCMCI$^+$ (grouped) achieves lower performance than its single graph counterpart. Further, \oursnonlinear{} and \ourslinear{} achieve relatively low clustering accuracy of $35.2 \pm 6.6\%$ and $35.4 \pm 5.2 \%$, due to the inherent similarities in the underlying SCMs.

\subsection{Ablation studies} \label{sec:ablation}
\textbf{Effect of number of samples per component.} We investigate the effect of a decreasing number of samples per mixture component on the performance of \ours{}, as the number of ground truth SCMs $K^\ast$ increases. We consider synthetic nonlinear data of dimension $D=10$ with $K^\ast=40, 60, 80, 100$ ground truth graphs in addition to the settings discussed in the main paper. We run \oursnonlinear{}, Rhino and PCMCI$^+$ on $N=1000$ samples generated from $K^\ast$ SCMs for increasing values of $K^\ast$, with $K=2K^\ast$. The results are presented in Figure \ref{fig:ablation_truenum_graphs}. \oursnonlinear{} suffers a gradual decrease in model performance, with roughly a $40\%$ decrease in F1 and $22\%$ decrease in AUROC from $K^\ast=1$ to $K^\ast=100$. Meanwhile, the performance of Rhino falls off more drastically and becomes equivalent to random guessing for large $K^\ast$.  The performance of PCMCI$^+$ (grouped) also decreases quite rapidly with the increase in $K^\ast$.

 {
\begin{figure}
    \centering
    \includegraphics[width=\textwidth]{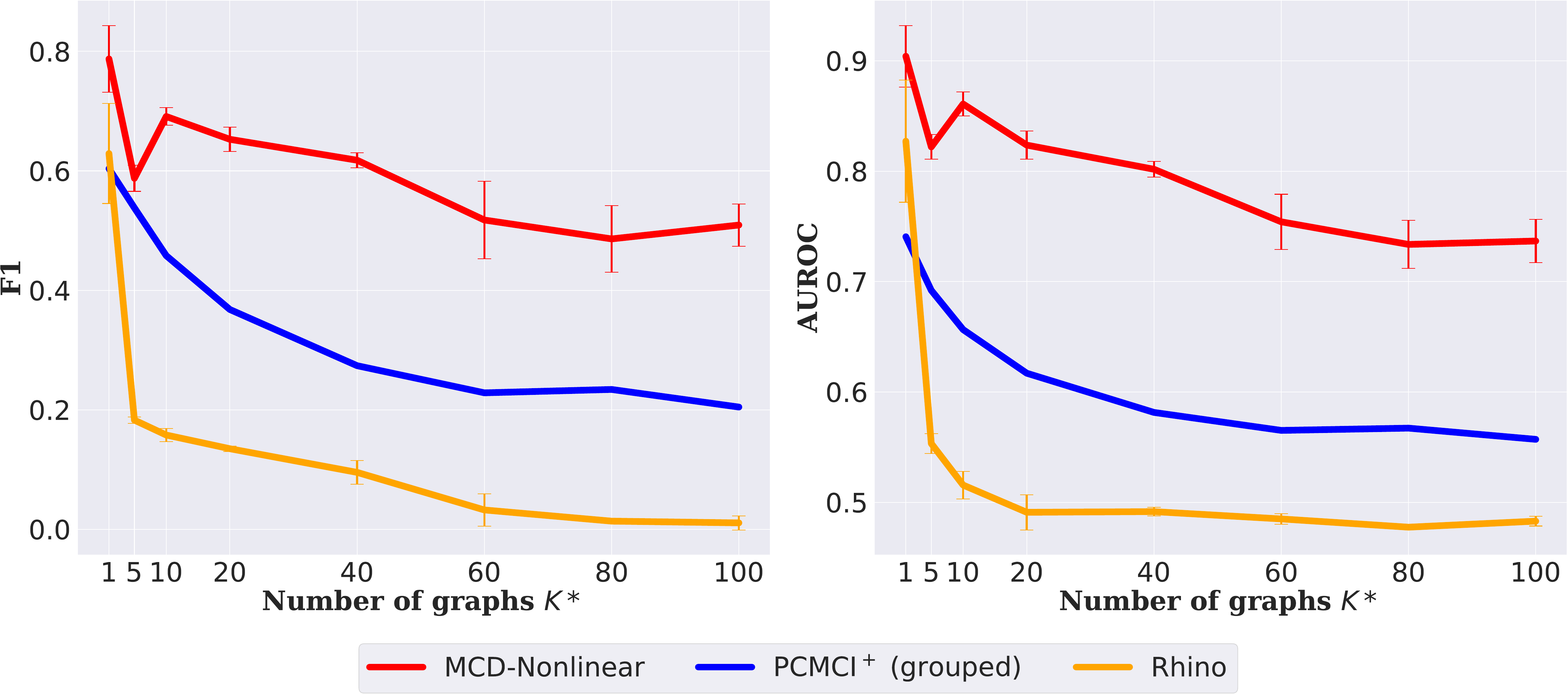}

    \caption{Effect of increasing number $K^\ast$ of underlying SCMs on F1 and AUROC on synthetic data with $D=10$. \ours{}'s performance declines gradually with decreased number of samples per mixture component, while Rhino's performance decays drastically.  {The performance of Rhino (grouped) decays slightly faster than \ours{}.}}
    \label{fig:ablation_truenum_graphs}
\end{figure}
}

\begin{figure}
    \centering
    \includegraphics[width=0.45\textwidth]{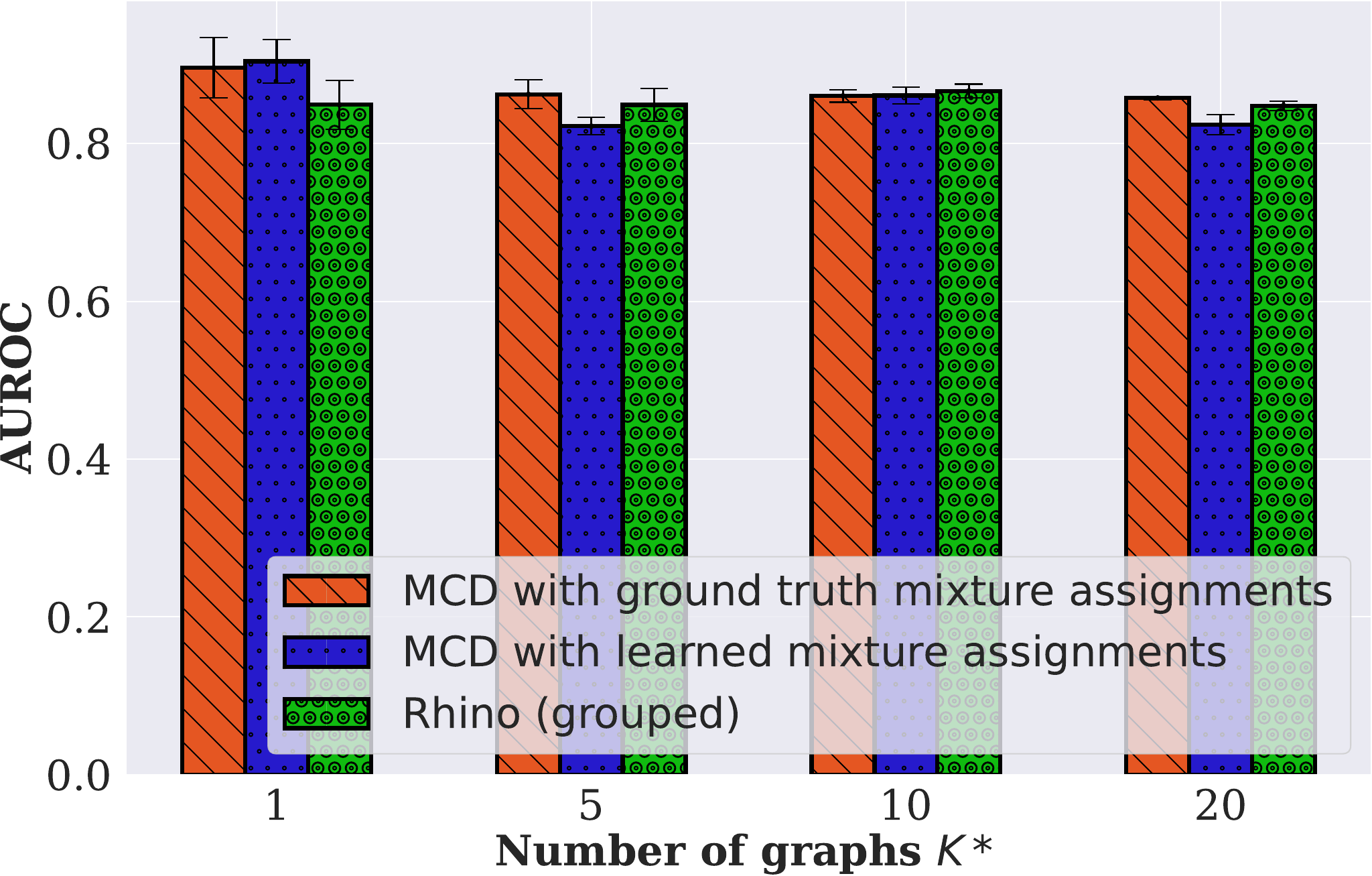}
    \includegraphics[width=0.45\textwidth]{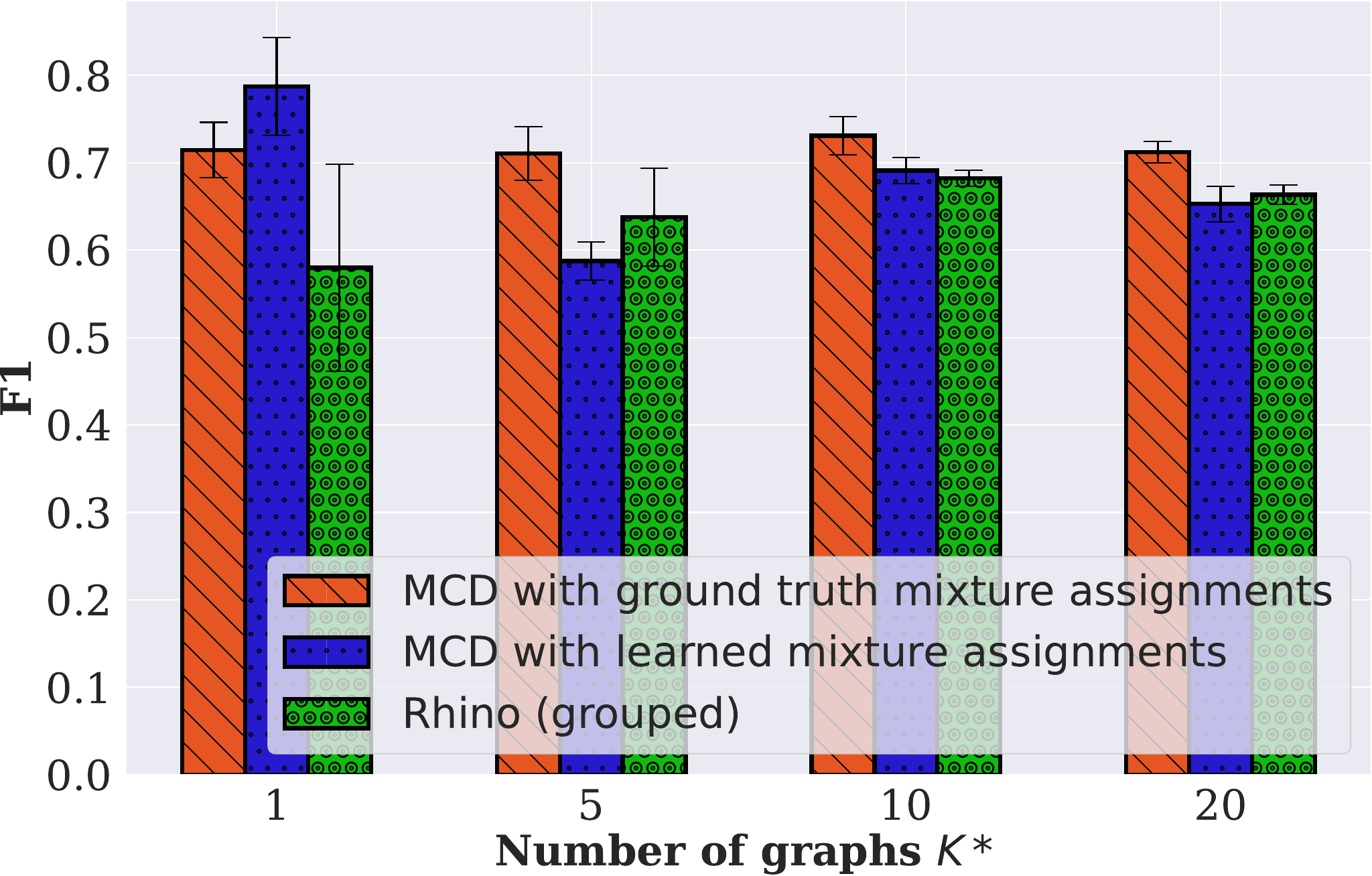}
    
    \caption{Comparison of model performance of \ours{} with ground-truth versus learned mixture assignments  {and Rhino (grouped)} on synthetic data with $D=10$. Expectedly, \ours{} performs better with explicit information about the cluster assignments, but it achieves comparable performance even with learned membership information.}
    \label{fig:ablation_truegraph}
\end{figure}

\textbf{Using ground truth membership assignments.} We assess \ours{} performance with learned versus ground-truth membership associations on synthetic data with $D=10$. As before, we set $K=2K^\ast$. Figure \ref{fig:ablation_truegraph} shows the results of this ablative experiment run with \oursnonlinear{} on the nonlinear synthetic dataset. The performance of \oursnonlinear{} with ground truth labels and Rhino-g is theoretically an upper bound on its performance. Encouragingly, we observe that our model performs close to this upper bound. 
% The largest difference in performance is observed for $K^\ast=1$, where \ours{} learns two separate causal models to explain a single mode.  {We also observe that for $K^\ast>1$, Rhino (grouped) performs a similar or slightly worse level of performance than \ours{}. This is due to the reduced number of samples per graph, which \ours{} is more robust to due to weight sharing (as implemented in Equation \eqref{eqn:mlp}).}

\paragraph{Effect of similarity of the graphs on performance}

\begin{figure}
    \centering
    
    \includegraphics[width=\textwidth]{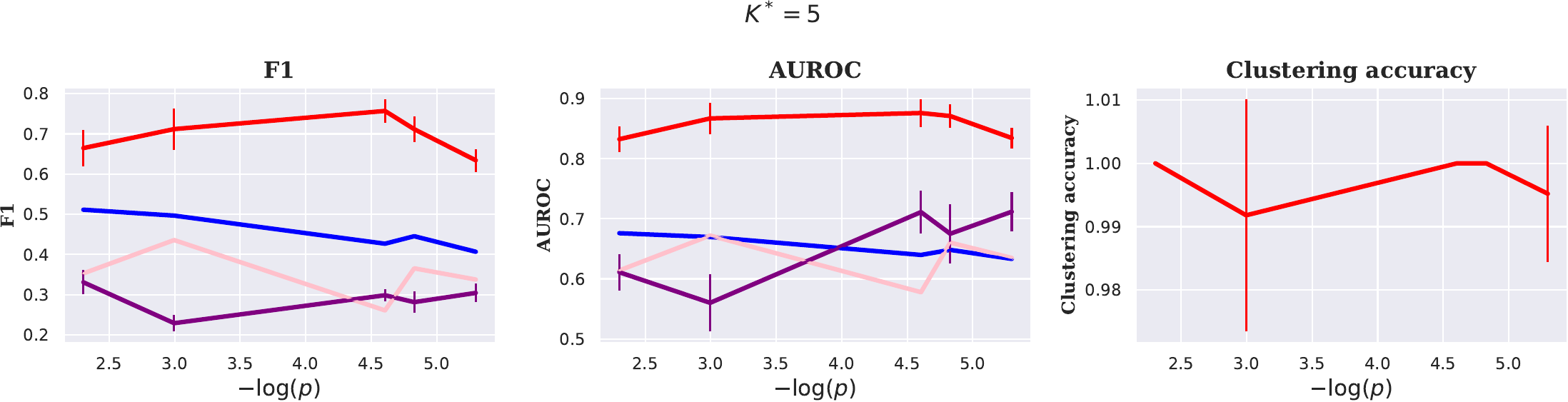}

    \includegraphics[width=\textwidth]{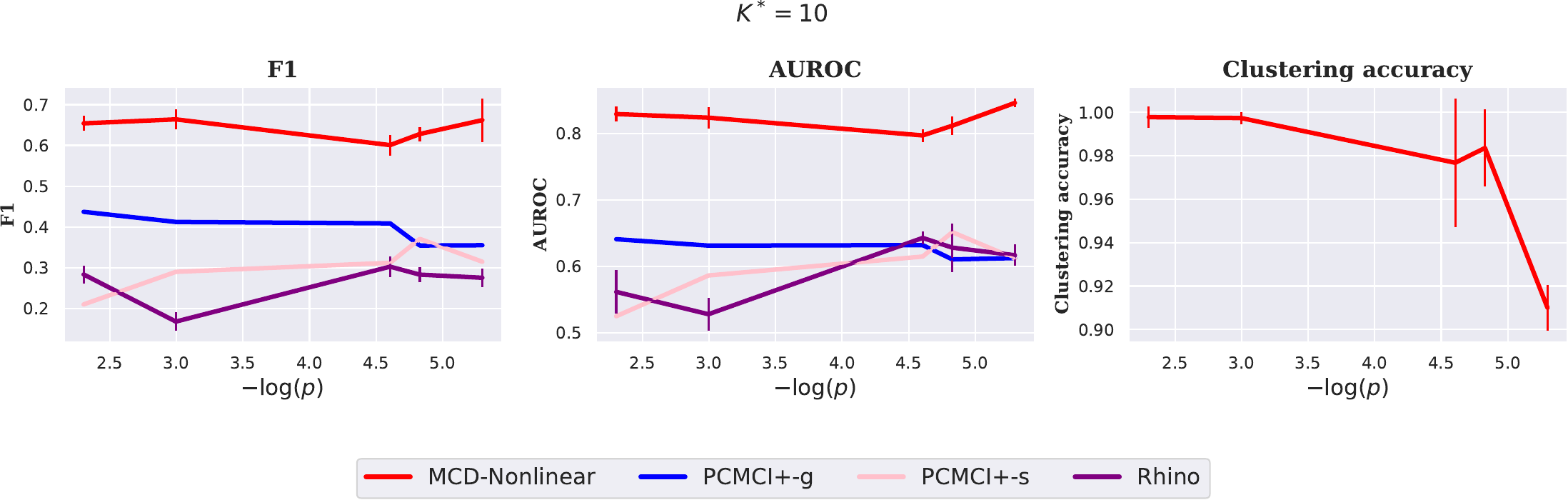}
    \caption{Causal discovery performance and clustering accuracy of \ours{} on synthetic datasets with similar graphs. \ours{} achieves good causal discovery performance and clustering accuracy even when $-\log p$ is high, i.e., the graphs are similar.}
    \label{fig:perturbed}
\end{figure}

We examine the performance of \ours{} when the causal graphs are similar to each other. The dataset setup is as follows: we first generate a random ER graph. We then perturb each edge with a probability $p$, i.e. flip the entry in the adjacency matrix from 0 to 1, and vice versa with probability $p$. We check if the resulting graph is a DAG. If yes, we add it to the pool of generating graphs. We repeat this procedure until we obtain $K^*$ DAGs. We then generate the synthetic dataset ($N=1000$ samples) with these $K^*$ DAGs using randomly generated MLPs and spline functions. We run \oursnonlinear{} with $K=2K^*$ on the resulting datasets with $D=10$ and $K=5, 10$. We set $p=0.005, 0.008, 0.01, 0.05, 0.1$, corresponding to varying levels of similarity of the underlying graphs. We also attach the pair-wise statistics for the resulting graphs in Table \ref{tab:pairwise_dataset}. The results are reported in Figure \ref{fig:perturbed}

The results indicate that \ours{} can achieve good clustering accuracy and causal discovery performance even when the constituent causal graphs are similar. In the $K^*=5$ case, we notice that the clustering accuracy remains high for all considered settings, and the F1 score decreases slightly for the higher values of $-\log p$ when the causal graphs are similar. 
For $K^*=10$, the clustering accuracy is considerably lower for $p=0.005$, but nevertheless the F1 score remains high. On the other hand, PCMCI$^+$-g suffers a gradual performance drop as the graphs become similar, while the single graph baselines Rhino and PCMCI$^+$-s have slightly better performance when the constituent causal graphs are more similar. 

 {
\subsection{Clustering accuracy for $D=5,10,20$ on synthetic datasets}
\begin{figure}
    \centering
    \includegraphics[width=0.3\textwidth]{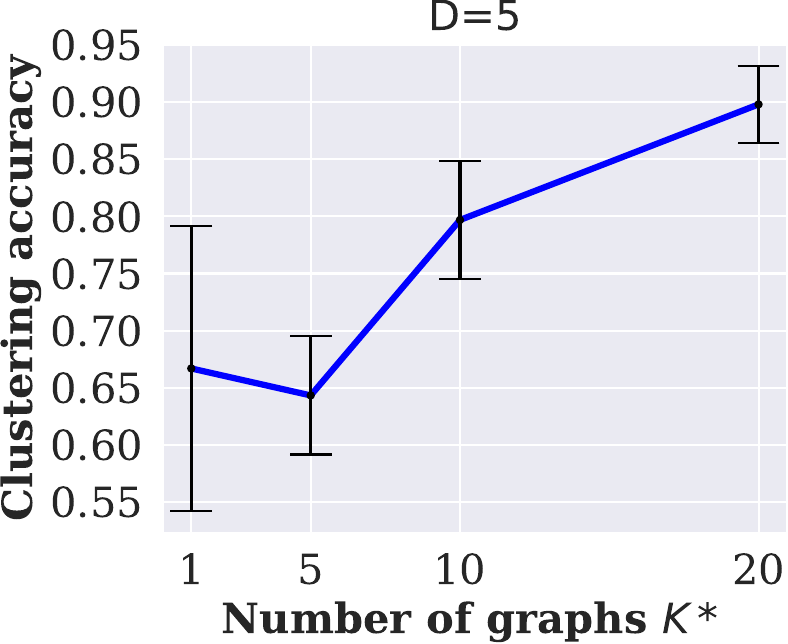}
    \includegraphics[width=0.3\textwidth]{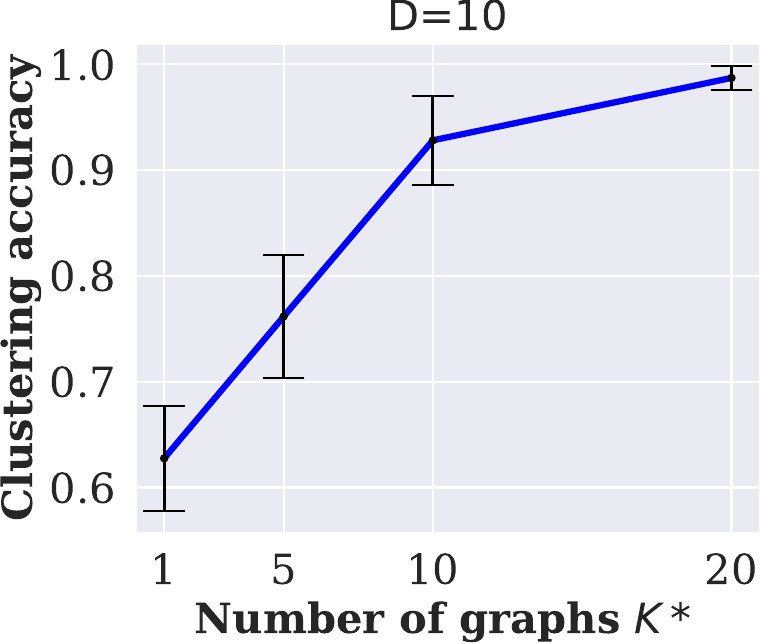}
    \includegraphics[width=0.3\textwidth]{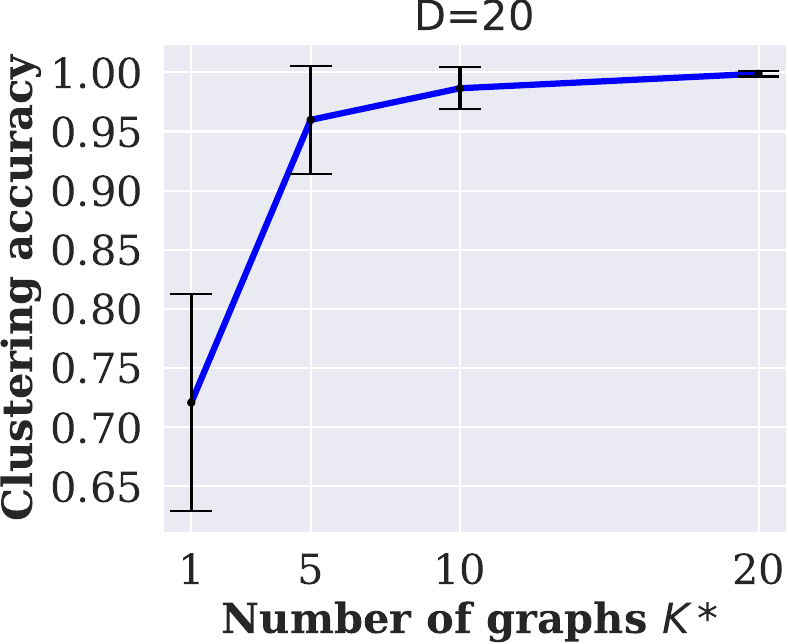}

    \includegraphics[width=0.3\textwidth]{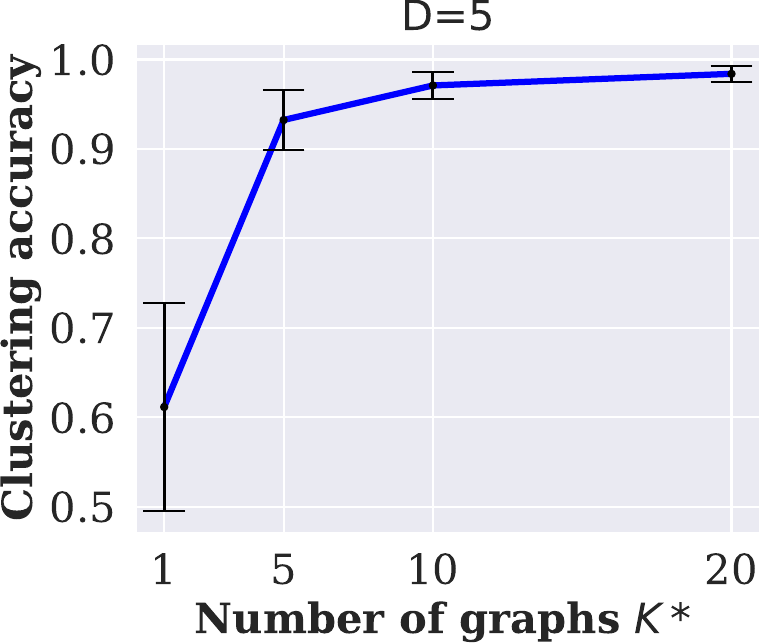}
    \includegraphics[width=0.3\textwidth]{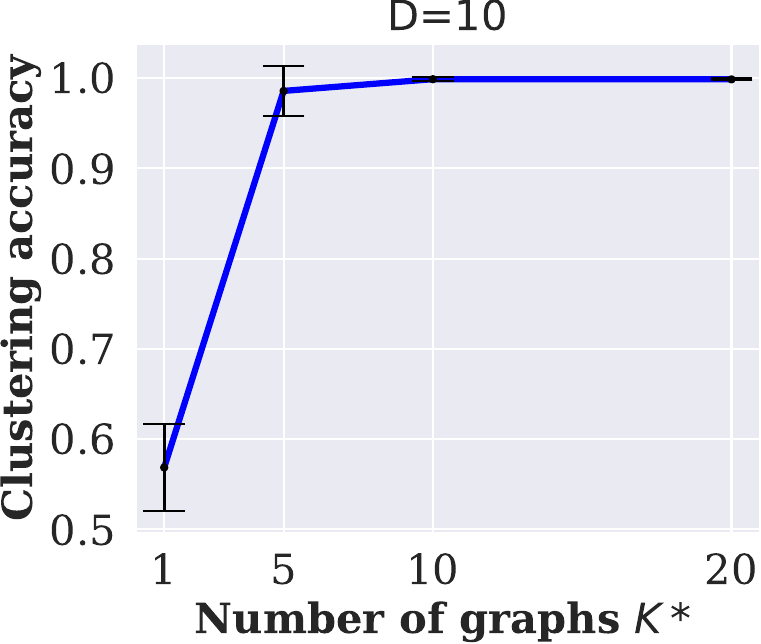}
    \includegraphics[width=0.3\textwidth]{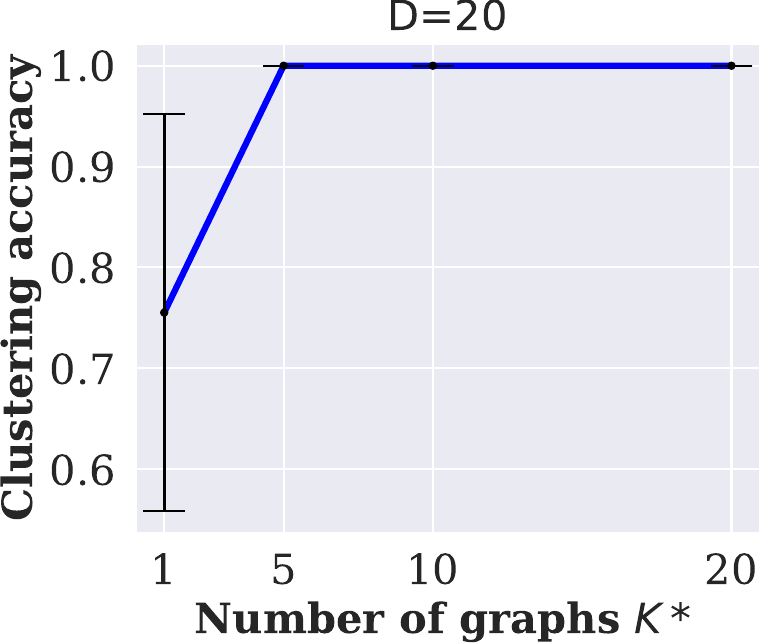}
    \caption{ {Plots showing the clustering accuracy of (top) \ourslinear{} and (bottom) \oursnonlinear{} vs $K^\ast$ on the linear and nonlinear synthetic datasets for $D=5,10,20$. }}
    \label{fig:cluster_acc_all_d}
\end{figure}

Figure \ref{fig:cluster_acc_all_d} shows the clustering accuracy of \ourslinear{} and \oursnonlinear{} for different values of $D$ on the linear and nonlinear synthetic datasets, respectively. For all settings, we set the hyperparameter $K=2K^\ast$. We observe that for most values of $K^\ast>1$, the clustering accuracy is quite high, while it remains low for $K^\ast=1$. Both \ourslinear{} and \oursnonlinear{} are particularly good at clustering for higher number of nodes $D$. As noted earlier, the low clustering accuracy for $K^\ast=1$ is expected since the single mode in the data distribution is `split' across two learned causal graphs.  
}

 {

\subsection{Clustering progression with training}
\label{sec:cluster_over_steps}

\begin{figure}
    \centering
    \includegraphics[width=0.48\textwidth]{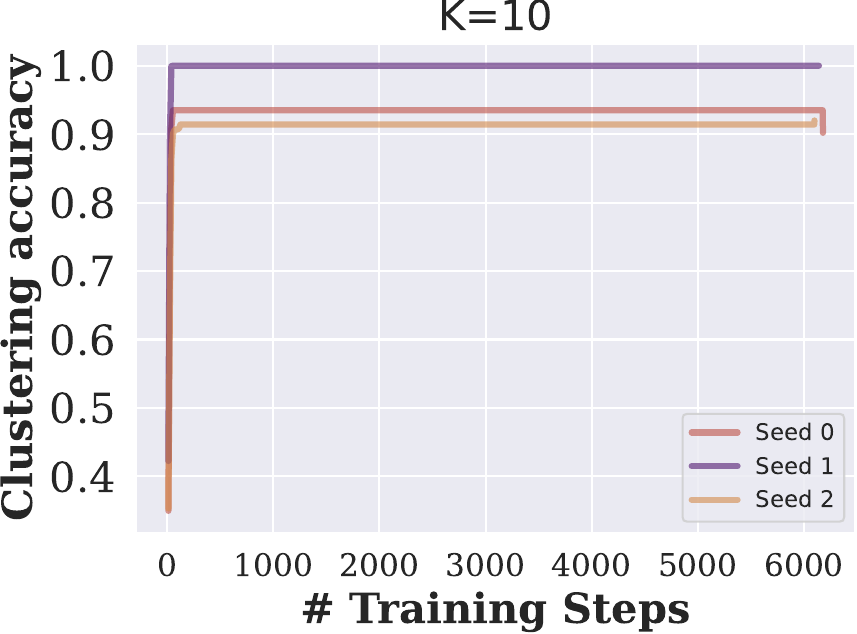}
    \includegraphics[width=0.48\textwidth]{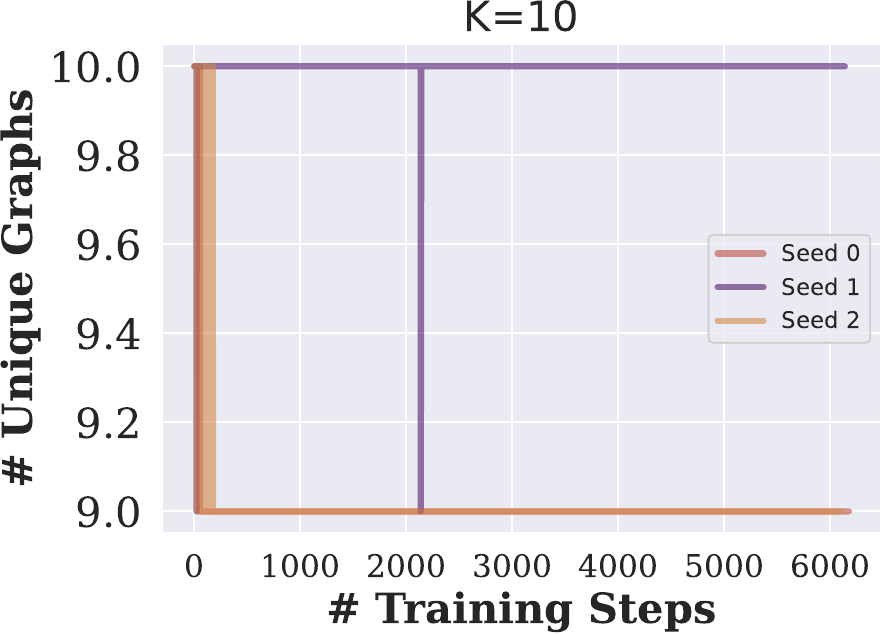}
    \includegraphics[width=0.48\textwidth]{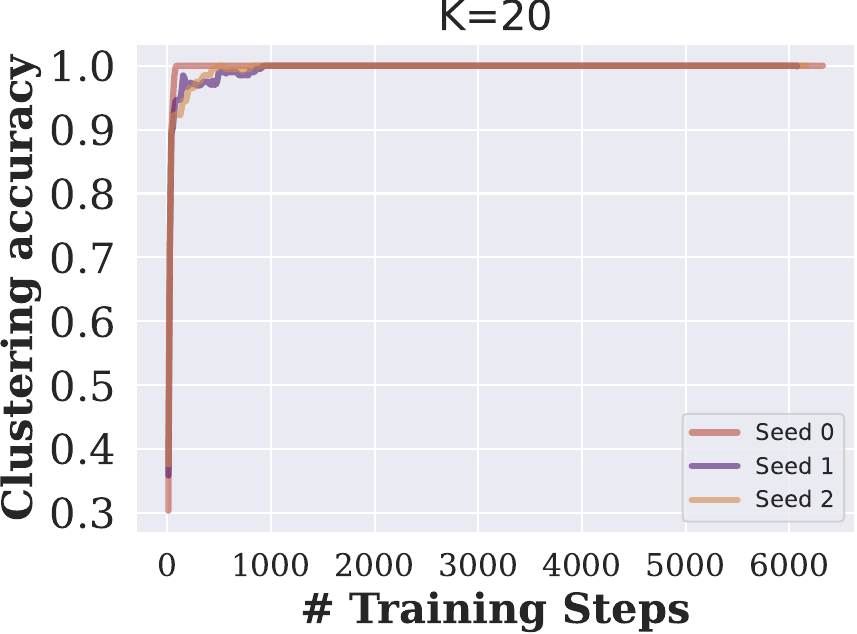}
    \includegraphics[width=0.48\textwidth]{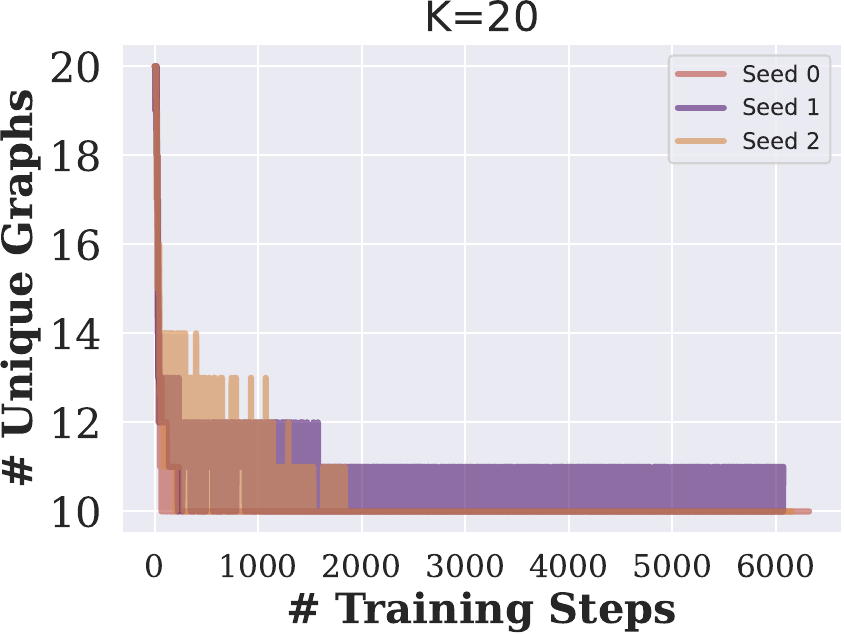}
    
    \caption{ {Plots showing the progression of (left) clustering accuracy (right) number of unique learned graphs for \oursnonlinear{} with the number of training steps on the nonlinear synthetic dataset with $D=10, K^\ast=10$. We observe that as training progresses, clustering accuracy increases for both the $K=10$ and $K=20$ runs; however, when $K=10$, some runs tend to learn a lower number of graphs, thus resulting in suboptimal clustering accuracy.}}
    \label{fig:graphs_over_time}
\end{figure}

We analyze the progression of clustering accuracy and the number of unique graphs learned with the number of training steps for \oursnonlinear{} on the nonlinear synthetic dataset with $D=10, K^\ast=10$. As training progresses, not all $K$ graphs are utilized. We count only those graphs for which at least one associated sample exists. Figure \ref{fig:graphs_over_time} shows the plots. We observe that when $K=20$, as training progresses, the algorithm groups together points from different causal graphs until they converge to the ``true" number of causal graphs $K^\ast=10$ and clustering accuracy converges to (approximately) 100\%; however, when $K=10$, we observe that the number of unique graphs can sometimes fall below $K^\ast=10$, resulting in suboptimal clustering accuracy.
}

\subsection{Netsim visualization}
\begin{figure}
    \centering
    \includegraphics[width=\textwidth]{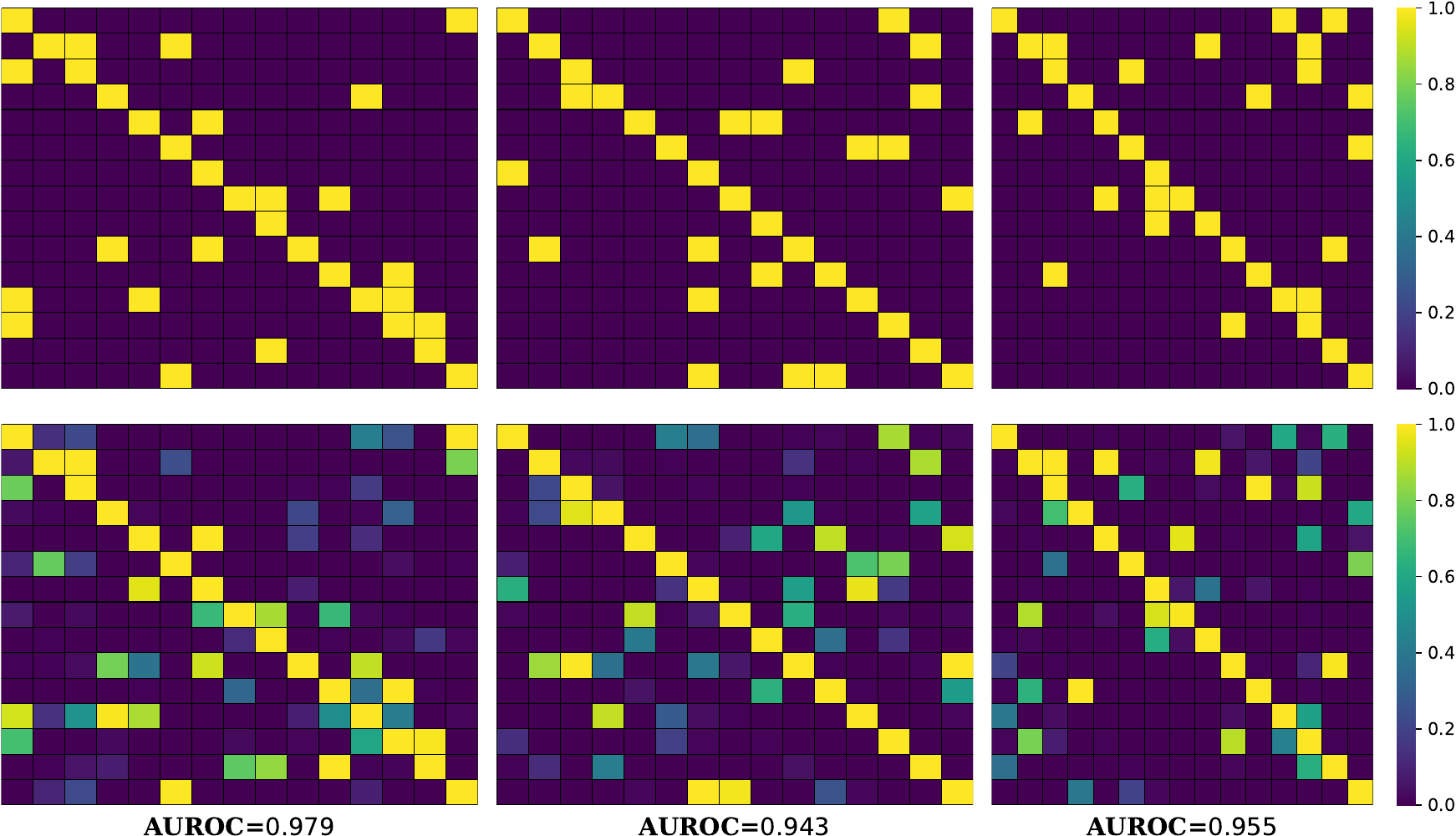}
    \caption{Heatmap for the Netsim-mixture dataset showing the (top) adjacency matrices of the ground-truth causal graphs, and the (bottom) edge probabilities for the top-3 discovered adjacency matrices (ranked by frequency of occurrence). We also report the graph-wise AUROC metrics.}
    \label{fig:netsim_heatmap}
\end{figure}

Figure \ref{fig:netsim_heatmap} shows a visualization of a heatmap of the predictions for the Netsim-mixture dataset. The 3 ground truth adjacency matrices and the top-3 discovered adjacency matrices, ranked by the prediction frequency, are shown. All 3 matrices achieve a high AUROC score, even though the poor calibration of scores results in the prediction of many spurious edges. 

\subsection{Timing analysis}
\label{sec:timing_analysis}

\begin{figure}
    \centering
    \includegraphics[width=0.6\textwidth]{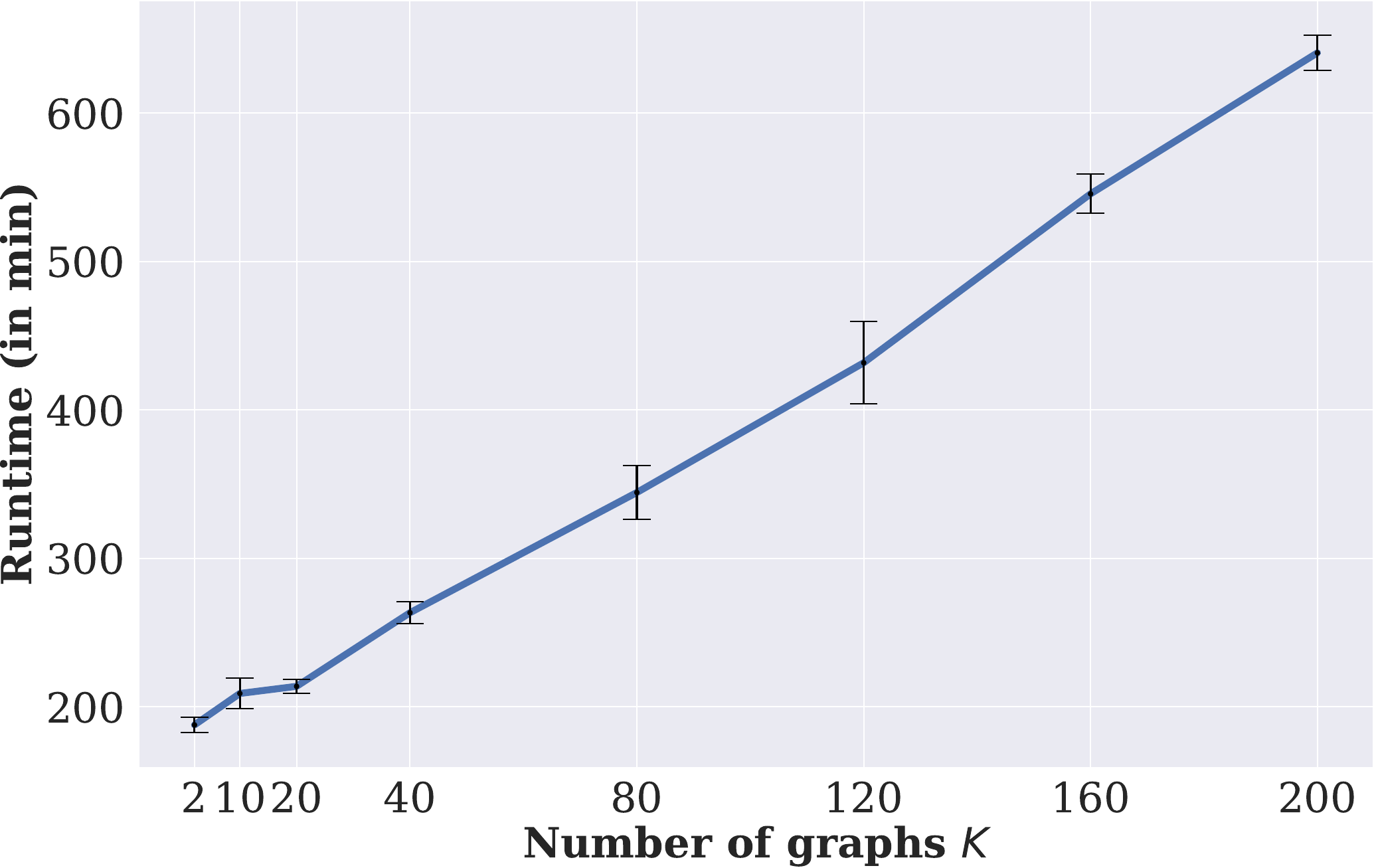}
    \caption{Run time plot of \ours{} as a function of $K$. A $100\times$ increase in $K$ from $2$ to $200$ results in a less than $4\times$ increase in run-time. }
    \label{fig:timing}
\end{figure}

In this section, we analyze the run-time of \oursnonlinear{} as a function of the hyperparameter $K$. As noted in Section \ref{sec:model_implementation}, \oursnonlinear{}, in theory, needs roughly $K$ times more operations than Rhino in each epoch due to the evaluation of the expectation over the variational distribution $r_\psi \left( Z^{(n)} \mid X^{(n)}\right)$ while calculating the ELBO. However, this does not translate to a $K$ times increase in model runtime.  We measure and plot the total runtime for training our model for the synthetic dataset with $D=10$ nodes as a function of $K$. Figure \ref{fig:timing} shows the plot.  

We observe that although the plot shows an approximately linear trend, the slope is much lesser than 1. In fact, a $100\times$ increase in $K$ from $2$ to $200$ results in a less than $4\times$ increase in run-time. Thus, \ours{} scales reasonably well with the number of mixture components $K$.

\section{Implementation details}  \label{sec:implementation_details}

In this section, we elaborate more on how we model the terms in equation \eqref{eqn:elbo} for \ourslinear{} and \oursnonlinear. 

We model the $K$ SCMs as additive noise models. For the $k^\text{th}$ causal model, we have:
\begin{equation*}
    \left. X^{ (n)}_t\right\rvert_k = f_k(\text{Pa}_{\mathcal{G}_k}(<t), \text{Pa}_{\mathcal{G}_k}(t)) + g_k(\text{Pa}_{\mathcal{G}_k}(<t), \epsilon_t),
\end{equation*}
where the function $f_k$ models the structural equation between the nodes and $g_k$ models the exogenous noise under causal model $\mathcal{M}_k$. 

\textbf{\ourslinear{}.} We implement the each of the $K$ models using a linear model:
\begin{equation}
    f_{k}^d\left(\text{Pa}_{\mathcal{G}_{k}}(\leq t)\right) = \sum_{\tau=0}^L \sum_{j=1}^D \left(\mathcal{G}_k \circ \mathcal{W}_k\right)_{\tau}^{j, d} \times
      X^{j, (n)}_{t-\tau},  
\end{equation}
where $\circ$ denotes the Hadamard product, and $\mathcal{W}_k \in \mathbb{R}^{(L+1)\times D \times D}$ is a learned weight tensor. We only model independent noise with this model, and set $g_k(\text{Pa}_{\mathcal{G}_{k}}(<t), \epsilon_t) = \epsilon_t$, i.e., the identity function.

Assuming that $\epsilon_t \sim \mathcal{N}(0, I)$, the marginal likelihood under each model $\mathcal{M}_k$ can be further simplified as follows, using the causal Markov assumption:
\begin{align}
    \log p_\theta \left(X_{1:T}^{(n)} \middle|  \mathcal{M}_{k} \right) &= \sum_{t=L}^T \sum_{d=1}^D \log p_\theta \left(X_t^{d, (n)} \middle|  \text{Pa}_{\mathcal{G}_{k}^d}(<t), \text{Pa}_{\mathcal{G}_{k}}^d(t)\right) \nonumber \\ &= \sum_{t=L}^T \sum_{d=1}^D  \left[X_t^{d, (n)} - f^{d}_{k} \left(\text{Pa}_{\mathcal{G}_{k}}^d(<t), \text{Pa}_{\mathcal{G}_{k}} \right) \right]^2 \nonumber \\
    &= \sum_{t=L}^T \sum_{d=1}^D  \left[X_t^{d, (n)} - \sum_{\tau=0}^L \sum_{j=1}^D \left(\mathcal{G}_k \circ \mathcal{W}_k\right)_{\tau}^{j, d} \times
      X^{j, (n)}_{t-\tau} \right]^2. 
\end{align}

\textbf{\oursnonlinear{}.} We use the Rhino model \citep{gong2022rhino} to model each of the $K$ SCMs. The functions $f_k$ are modeled as follows:
\begin{equation}
    f_{k}^d \left( \text{Pa}_{\mathcal{G}_{k}}(\leq t) \right) = \Xi_f \left( \left[ \sum_{\tau=0}^L \sum_{j=1}^D \left(\mathcal{G}_{k}\right)_{\tau}^{j, d} \times 
       \ell_f \left( \left[X^{j, (n)}_{t-\tau}, \left(\mathcal{E}_{k}\right)^{j, (n)}_{\tau} \right] \right),  \left(\mathcal{E}_{k}\right)^{d, (n)}_0 \right] \right),
\end{equation}
where $\Xi_f$ and $\ell_f$ are multi-layer perceptron networks that are shared across all $K$ causal models $\mathcal{M}_{1:K}$ and all $D$ nodes, and $\mathcal{E}_k \in \mathbb{R}^{(L+1)\times D \times e}$ are embeddings (with embedding dimension $e$) corresponding to model $k$. A similar architecture is used for the hypernetwork that predicts parameters for the conditional spline flow model, with embeddings $\mathcal{F}_k$, and hypernetworks $\Xi_\epsilon$ and $\ell_\epsilon$. The only difference is that the output dimension of $\Xi_\epsilon$ is different, being equal to the number of spline parameters. 
The noise variables $\epsilon_t^d$ are described using a conditional spline flow model, 
\begin{equation}
    p_{g^d_k}(g^d_k(\epsilon_t^d) \mid  \text{Pa}_{\mathcal{G}_k}^d(<t) ) = p_\epsilon(\epsilon_t^d) \left|  \frac{\partial (g^d_k)^{-1}}{\partial \epsilon_t^d}\right|,
\end{equation}

with $\epsilon_t^d$ modeled as independent Gaussian noise.  

Using the causal Markov assumption:
\begin{align}
    \log p_\theta \left(X_{1:T}^{(n)} \middle|  \mathcal{M}_{k} \right) &= \sum_{t=L}^T \sum_{d=1}^D \log p_\theta \left(X_t^{d, (n)} \middle|  \text{Pa}_{\mathcal{G}_{k}}^d(<t), \text{Pa}_{\mathcal{G}_{k}}^d(t) \right) \nonumber \\ &= \sum_{t=L}^T \sum_{d=1}^D \log p_{g^d_{k}}\left(u_t^{d, (n)} \middle| \text{Pa}_{\mathcal{G}_{k}}^d(<t) \right)
\end{align}
where $\displaystyle u_{t}^{d, (n)} = X_t^{d, (n)} - f^{d}_{k} \left(\text{Pa}_{\mathcal{G}_{k}}^d(<t), \text{Pa}_{\mathcal{G}_{k}}^d(t) \right)$. 

The prior distribution $p(\mathcal{M}_{1:K})$ is modeled as follows:
\begin{equation}
    p_\theta(\mathcal{M}_{1:K}) \propto \prod_{k=1}^K \exp \left(-\lambda \left\lVert \left({\mathcal{G}_k}\right)_{1:T}\right\rVert^2 - \sigma h\left(\left(\mathcal{G}_k\right)_0 \right) \right). \label{eqn:prior_term}
\end{equation}
The first term is a sparsity prior and $h\left(\left(\mathcal{G}_k\right)_0\right)$ is the acyclicity constraint from \citep{zheng2018dags}.

\subsection{Calculation of clustering accuracy}
\label{sec:cluster_acc}

We would like to evaluate the accuracy of our method in grouping samples based on the underlying SCMs. However, the assigned cluster indices by the model and the `ground-truth' cluster indices might not match nominally, even though they refer to the same grouping assignment. For example, the cluster assignment of $\left(1, 1, 1, 2, 2 \right)$ for $N=5$ points is equivalent to the assignment $\left(2, 2, 2, 1, 1 \right)$. In other words, we want a permutation invariant accuracy metric between the inferred cluster assignments $\tilde{Z}$ and true cluster assignments $Z$ with $\tilde{Z}, Z \in \mathbb{N}^N$. We define
\begin{equation*}
    \text{Cluster Acc.} \left( \tilde{Z}, Z\right) = \max_{\pi \in S_{K}} \frac{1}{N} \sum_{n=1}^N \mathrm{1}\left( \pi(\tilde{Z}_n) = Z_n\right)
\end{equation*}
with $S_K$ denoting the permutation group over $K$ elements. Evaluating the cluster accuracy naively would require $K!$ operations. However, we use the Hungarian algorithm to find the correct permutation in $O(K^3)$ time\footnote{This approach and implementation are adapted from \url{https://smorbieu.gitlab.io/accuracy-from-classification-to-clustering-evaluation/}}. 

\section{Experimental details}

\subsection{Synthetic datasets setup}
\label{sec:syn_data_setup}
This section provides more details about how we set up and run experiments using \ours{} on synthetic datasets.  We set the number of mixture components $K$ to twice that of true graphs (i.e., $K = 2K^\ast$) to showcase its robustness against over-specification of the number of components. We set a uniform prior for the mixing probabilities $p\left(Z^{(n)}\right)$, i.e. $p(Z^{(n)}=k)=\frac{1}{K}\, \forall k \in \left\{1, \ldots, K \right\}$. Our implementation of the likelihood function for Rhino-g on the synthetic datasets matches the type of causal relationships modeled, i.e., we use the linear model described in Equation \eqref{eqn:linear_model} on the linear dataset and the nonlinear variant described in Equation \eqref{eqn:mlp} for the nonlinear datasets.  

\paragraph{Dataset generation.} We generate two separate sets of synthetic datasets: a linear dataset with independent Gaussian noise and a nonlinear dataset with history-dependent noise modeled using conditional splines \citep{durkan2019neural}. We generate a pool of $K^\ast$ random graphs (specifically, Erdős-Rényi graphs) and treat them as ground-truth causal graphs. To generate a sample $X^{(n)}$, we assign it to a graph by drawing $Z^{(n)} \sim \text{Categorical}(K^\ast)$, and use the corresponding graph $\mathcal{G}_{Z^{(n)}}$ from this pool to model relationships between the variables.

\textbf{Linear dataset.} We model the data as:
\begin{equation*}
    X^{d, (n)}_t = \sum_{\tau=0}^L \sum_{j=1}^D \left(\mathcal{G}_{Z^{(n)}} \circ \mathcal{W}_{Z^{(n)}}\right)_{\tau}^{j, d} \times X^{j, (n)}_{t-\tau} +  \epsilon_t^d,
\end{equation*}
with $\epsilon_t^d \sim \mathcal{N}(0, 0.25)$. 
Each entry of the matrices $\mathcal{W}_k, \quad k=1,\ldots, K$ is drawn independently from $\mathcal{U}[0.1, 0.5] \cup \mathcal{U}[-0.5, -0.1]$.

\textbf{Nonlinear dataset.} We model the data as:
\begin{equation*}
    X^{d, (n)}_t = f^{d}_k \left(\text{Pa}_{\mathcal{G}_k}^d(<t), \text{Pa}^i_{\mathcal{G}_k}(t) \right) +  \epsilon_t^d,
\end{equation*}
where $f^{d}_k$ are randomly initialized multi-layer perceptrons (MLPs), and the random noise $\epsilon_t^d$ is generated using history-conditioned quadratic spline flow functions \citep{durkan2019neural}.

\subsection{Implementation of validation step for \ours{}}
\label{sec:validation}
\ours{} learns a sample-wise membership variable $Z^{(n)}$ for every sample in the dataset by optimizing the ELBO. In order to evaluate the log-likelihood of the samples in the validation set, we still need to infer their corresponding membership information $Z^{(n)}$. Hence, during each validation step, we fix the weights of the other parameters and perform one step of gradient descent for the membership weights $Z^{(n)}$ with respect to the ELBO for the samples in the validation set. This way, we ensure that all samples in the input dataset are assigned to a mixture component, and the validation likelihood can be evaluated. 
 
\subsection{Hyperparameter details}
Since \ours{} uses the acyclicity constraint from \citet{zheng2018dags}, we use an augmented Lagrangian training procedure to ensure that our model produces DAGs. We closely follow the implementation of the procedure from \citet{geffner2022deep, gong2022rhino} with one exception: we modify the convergence criteria by increasing the number of outer steps for which the DAG penalty needs to be lower than a threshold (set to $10^{-8}$). This modification enables the model to train for longer and prevents premature stopping. In the interest of fairness, we make this change to both \ours{} and Rhino.

We used the rational spline flow model described in \citep{durkan2019neural}. We use the quadratic or linear rational spline flow model in all our experiments, both with 8 bins. The MLPs $\ell$ and $\Xi$ have 2 hidden layers each, with hidden dimensions set to $\max\left(4D, e, 64 \right)$ with LeakyReLU activation functions, where $e$ is the embedding dimension. We also use layer normalization and skip connections. The temperature for sampling the adjacency matrix from $q_\phi \left( \mathcal{M}_{1:K}\right)$ using the Gumbel Softmax distribution was set to 0.25, and the temperature $\tau_r$ for sampling from the mixing rates variational distribution was set to 1. We used the same hyperparameters for both \oursnonlinear{} and \ourslinear{}.
Table \ref{tab:hyperparameters} summarizes the hyperparameters used for training.

\begin{table}
\centering
\begin{tabular}{|c|ccccc|}
\hline
\textbf{Dataset}                   & Synthetic $(D=5, 10, 20)$    & Netsim-mixture             & DREAM3                     & S\&P100                    & Netsim                     \\ \hline
\textbf{Hyperparameter}            &                            &                            &                            &                            &                            \\
\hline
Matrix LR                 & $10^{-2}$ & $10^{-2}$ & $10^{-3}$ & $10^{-2}$ & $10^{-2}$ \\
Likelihood LR             & $10^{-3}$ & $10^{-3}$ & $10^{-3}$ & $10^{-3}$ & $10^{-3}$ \\
Batch Size                & 128                        & 64                         & 64                         & 64                         & 64                         \\
\# Outer auglag steps     & 100                        & 60                         & 60                         & 60                         & 60                         \\
\# Max inner auglag steps & 6000                       & 2000                       & 6000                       & 2000                       & 2000                       \\
Embedding dim $e$           & $=D$                         & 15                         & 32                         & 100                        & 15                         \\
Sparsity factor $\lambda$ & 5                          & 25                         & 10                         & 20& 25                         \\
Spline type               & Quadratic                  & Linear                     & Linear                     & Linear                     & Linear                     \\ \hline
\end{tabular}
\caption{Table showing the hyperparameters used with \ours{}.}
\label{tab:hyperparameters}
\end{table}

\paragraph{Baselines.} Rhino was trained with similar hyperparameters as \ours{} on all datasets. For all other baselines, the default hyperparameter values are used. 
 {For Rhino and \ours{}, which parameterize the causal graphs as Bernoulli distributions over each edge, we use the inferred edge probability matrix as the ``score", and evaluate the AUROC metric between the score matrix and the true adjacency matrix. For DYNOTEARS, we use the absolute value of the output scores and evaluate the AUROC. Since PCMCI+ and VARLiNGAM only output adjacency matrices, we directly evaluate the AUROC between the predicted and true adjacency matrices.}

\subsection{Post-processing the output of PCMCI$^+$} \label{sec:pcmci_process}
 PCMCI$^+$ produces Markov equivalence classes rather than fully oriented causal graphs for the instantaneous adjacency matrix. To make its outputs comparable, we post-process the resultant edges. We follow the setup in \citet{gong2022rhino} and enumerate up to 3000 DAGs for the instantaneous matrix. We ignore the edges (i.e., set the corresponding entries in the adjacency matrix to 0) whose orientations are undecided. We compare all outputs against the ground truth during the evaluation and return the average metric across all enumerations.

\subsection{Aggregating the temporal adjacency matrix across time} \label{sec:netsim_dream3}
The Netsim and DREAM3 datasets used in the evaluation provide ground-truth time-aggregated causal graphs. In order to make our model output comparable, we follow the procedure outlined in \citep{gong2022rhino} to convert the time-lag adjacency matrix to an aggregated matrix. The $(i,j)^\text{th}$ entry of the aggregated matrix $\mathcal{G}_\text{agg}$ is 1 iff $\mathcal{G}_{\ell}^{ij}=1$ for some lag value $\ell$ in the time-lag matrix $\mathcal{G}$. Both Rhino and \ours{} represent the edges as Bernoulli random variables, hence output a probability score for each edge. For evaluating the F1 score, we threshold the probability values at 0.5, i.e., edges with a probability $\geq$ 0.5 are considered as predicted edges. 

 {

\subsection{Pair-wise graph distance in the mixture distributions}
Table \ref{tab:pairwise_dataset} shows the pairwise graph distances between the ground-truth graphs of the mixture distributions used in the paper. We calculate the Structural Hamming Distance (SHD) between every pair of graphs in the mixture, and report the mean, standard deviation, minimum and maximum values. 
}

\begin{table}
\centering
\begin{tabular}{|c|c|c|c|c|c|c|}
\hline
      \textbf{Dataset}                & $D$    &  $K^\ast$  &  \textbf{Avg. SHD} &  \textbf{Std. dev. SHD} & \textbf{Min. SHD} & \textbf{Max. SHD}     \\ \hline
Synthetic (nonlinear) & 5   & 5  & 21.00  & 2.57   & 15  & 24  \\ 
Synthetic (nonlinear) & 5   & 10 & 22.09  & 2.60   & 16  & 26  \\ 
Synthetic (nonlinear) & 5   & 20 & 22.19  & 2.92   & 14  & 32  \\ 
Synthetic (nonlinear) & 10  & 5  & 49.80  & 2.64   & 43  & 54  \\ 
Synthetic (nonlinear) & 10  & 10 & 53.42  & 3.03   & 47  & 59  \\ 
Synthetic (nonlinear) & 10  & 20 & 52.73  & 3.90   & 43  & 64  \\ 
Synthetic (nonlinear) & 20  & 5  & 120.20 & 2.14   & 117 & 124 \\ 
Synthetic (nonlinear) & 20  & 10 & 114.89 & 4.42   & 104 & 123 \\ 
Synthetic (nonlinear) & 20  & 20 & 114.28 & 5.13   & 101 & 127 \\ 
Synthetic (linear)    & 5   & 5  & 21.40  & 3.14   & 16  & 27  \\ 
Synthetic (linear)    & 5   & 10 & 22.18  & 2.81   & 16  & 28  \\ 
Synthetic (linear)    & 5   & 20 & 22.07  & 3.16   & 12  & 30  \\ 
Synthetic (linear)    & 10  & 5  & 48.60  & 3.75   & 41  & 54  \\ 
Synthetic (linear)    & 10  & 10 & 54.16  & 3.53   & 45  & 60  \\ 
Synthetic (linear)    & 10  & 20 & 54.52  & 4.14   & 42  & 67  \\ 
Synthetic (linear)    & 20  & 5  & 114.20 & 2.23   & 111 & 118 \\ 
Synthetic (linear)    & 20  & 10 & 111.58 & 5.17   & 103 & 124 \\ 
Synthetic (linear)    & 20  & 20 & 113.42 & 4.82   & 101 & 126 \\ 
Synthetic-Perturbed $(p=0.005)$& 10& 5& 2.40& 0.92& 1& 4
\\ 
Synthetic-Perturbed $(p=0.008)$& 10& 5& 5.20& 0.98& 4& 7
\\ 
Synthetic-Perturbed $(p=0.01)$& 10& 5& 7.00& 0.89& 5& 8
\\ 
Synthetic-Perturbed $(p=0.05)$& 10& 5& 21.40& 5.37& 15& 30
\\ 
Synthetic-Perturbed $(p=0.1)$& 10& 5& 48.20& 2.64& 45& 53
\\ 
Synthetic-Perturbed $(p=0.005)$& 10& 10& 2.44& 1.04& 1& 5
\\ 
Synthetic-Perturbed $(p=0.008)$& 10& 10& 3.56& 1.36& 1& 6
\\ 
Synthetic-Perturbed $(p=0.01)$& 10& 10& 6.11& 2.51& 1& 12
\\ 
Synthetic-Perturbed $(p=0.05)$& 10& 10& 23.04& 4.17& 15& 32
\\ 
Synthetic-Perturbed $(p=0.1)$& 10& 10& 47.56& 8.56& 29& 65
\\
DREAM3                & 100 & 5  & 517.60 & 202.13 & 234 &
896 \\
Netsim-mixture        & 15  & 3  & 34.00  & 1.63   & 32  &
36  \\
Netsim                & 5   & 14 & 2.59   & 1.17   & 1   &
5  \\
\hline
\end{tabular}
\caption{ {Pair-wise graph statistics for experimental datasets used in the paper.}}
\label{tab:pairwise_dataset}
\end{table}

 {

\section{Toy example}
We provide a toy example to illustrate the importance of modeling the heterogeneity of a multi-modal dataset. Consider a dataset where each sample $X^{(n)}$ from the dataset $\displaystyle \left\{ X_{1:T}^{1:D, (n)} \right\}_{n=1}^N$ is generated from one out of the two following SCMs, chosen with equal probability:
\begin{align*}
    X_t^{1, (n)} &=  0.4 X_{t-1}^{2, (n)} + 0.6 X_{t}^{3, (n)} + \epsilon_1^{(n)} \\
    X_t^{2, (n)} &=  0.3 X_{t-1}^{3, (n)} + 0.3 X_{t}^{3, (n)} + \epsilon_2^{(n)} \\
    X_t^{3, (n)} &=  0.5 X_{t-1}^{1, (n)} + \epsilon_3^{(n)}
\end{align*}
(or)
\begin{align*}
    X_t^{1, (n)} &=  0.7 X_{t-1}^{3, (n)} - 0.2 X_{t}^{2, (n)} + \epsilon_1^{(n)} \\
    X_t^{2, (n)} &=  0.2 X_{t-1}^{1, (n)} + 0.4 X_{t}^{3, (n)} + \epsilon_2^{(n)} \\
    X_t^{3, (n)} &=  -0.3 X_{t-1}^{1, (n)} + \epsilon_3^{(n)}.
\end{align*}
These SCMs can be represented through the temporal causal graphs given in Figure \ref{fig:ex-causal-graphs}.
\begin{figure}
    \centering
    \includegraphics[width=0.6\textwidth]{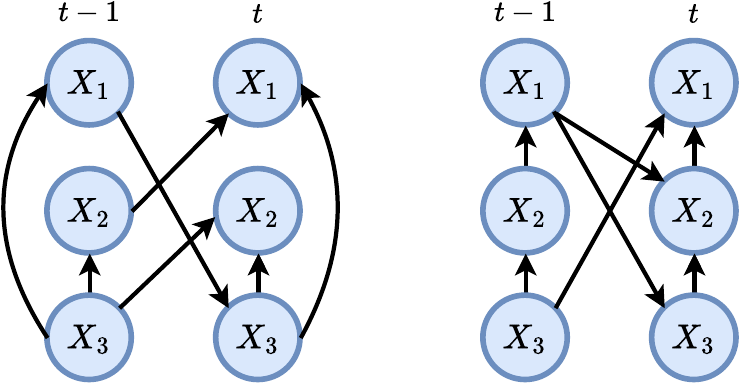}
    \caption{Temporal causal graphs which represent the causal relationships encoded by the SCMs.}
    \label{fig:ex-causal-graphs}
\end{figure}

However, if the graph membership of the samples is unknown, inferring a single causal graph to explain the causal relationships from the dataset would result in spurious causal relationships. For example, going by conditional independence tests, note that none of the nodes would be conditionally independent of each other for any conditioning set. This is also shown in the output of the PCMCI$^+$ algorithm, where a dense graph is inferred as shown in Figure \ref{fig:pcmci-output}. Thus, it is crucial to use a mixture distribution to model observational data from such heterogeneous distributions.

\begin{figure}
    \centering
    \includegraphics[width=0.3\textwidth]{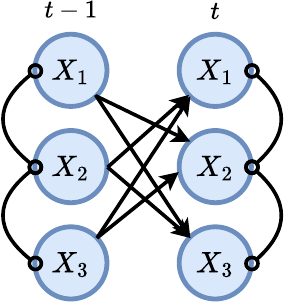}
    \caption{ {PCMCI$^+$ output on the toy-example. The algorithm infers a dense graph with many spurious causal relationships.}}
    \label{fig:pcmci-output}
\end{figure}
}

\end{document}